\pdfoutput=1
\documentclass[11pt, letterpaper, shortlabels, inline]{berkeley}

\usepackage{color-edits}

\newcommand{\eg}{\textit{e.g.}}
\newcommand{\etc}{\textit{e.t.c.}}
\newcommand{\ie}{\textit{i.e.}}

\ifx\assumption\undefined
\newtheorem{assumption}{Assumption}
\fi

\usepackage[capitalize,noabbrev]{cleveref}
\makeatletter
\def\adl@drawiv#1#2#3{%
        \hskip.5\tabcolsep
        \xleaders#3{#2.5\@tempdimb #1{1}#2.5\@tempdimb}%
                #2\z@ plus1fil minus1fil\relax
        \hskip.5\tabcolsep}
\newcommand{\cdashlinelr}[1]{%
  \noalign{\vskip\aboverulesep
           \global\let\@dashdrawstore\adl@draw
           \global\let\adl@draw\adl@drawiv}
  \cdashline{#1}
  \noalign{\global\let\adl@draw\@dashdrawstore
           \vskip\belowrulesep}}
\makeatother
\newcommand{\method}{{Agentopia}\xspace}

\usepackage{wrapfig}
\title{\method: Long-Term Life Simulation and Learning in Agent Societies}

\usepackage[all]{hypcap}

\usepackage[authoryear, round]{natbib}

\usepackage{hyperref}[citecolor=magenta,linkcolor=magenta]

\hypersetup{
    colorlinks = true,
    citecolor = {magenta},
}

\usepackage{multirow, makecell, caption}\usepackage{microtype}
\usepackage{graphicx}
\usepackage{booktabs} %

\usepackage[utf8]{inputenc}
\usepackage[T1]{fontenc}
\usepackage{url}
\usepackage{amsfonts}
\usepackage{xcolor}
\usepackage{colortbl}
\usepackage{pifont}

\usepackage{mathrsfs}
\usepackage{nicefrac}
\usepackage{dsfont}
\newenvironment{inparaenum}[1][]
  {\begin{enumerate*}[#1]}
  {\end{enumerate*}}
\usepackage{ulem}
\usepackage{arydshln}
\usepackage{xspace}
\usepackage[capitalize,noabbrev]{cleveref}
\usepackage{subcaption}
\usepackage{wrapfig}
\usepackage{lipsum}
\usepackage{listings}

\usepackage{amsmath}
\usepackage{amssymb}
\usepackage{mathtools}
\usepackage{amsthm}
\usepackage{bbm}

\usepackage{algpseudocode}
\usepackage{setspace}

\usepackage{tabularx}
\newcolumntype{C}{>{\centering\arraybackslash}X}

\definecolor{mygray}{RGB}{226, 226, 226}
\definecolor{myred}{RGB}{252, 142, 142}
\definecolor{mygreen}{RGB}{147, 255, 143}
\definecolor{myblue}{RGB}{144, 155, 255}
\definecolor{myyellow}{RGB}{253, 253, 143}
\definecolor{mypurple}{RGB}{255, 142, 250}

\newcommand{\dq}[1]{``#1''}

\newcommand{\apartment}{\textbf{The Apartment}\xspace}
\newcommand{\campus}{\textbf{The Campus}\xspace}
\newcommand{\academy}{\textbf{Arcane Academy}\xspace}
\newcommand{\cmark}{\ding{51}}
\newcommand{\xmark}{\ding{55}}

\usepackage[export]{adjustbox}

\newcommand\pythonstyle{\lstset{
basicstyle=\ttfamily\footnotesize,
language=Python,
morekeywords={self, clip, exp, mse_loss, uniform_sample, concatenate, logsumexp},              %
keywordstyle=\color{deepblue},
emph={MyClass,__init__},          %
emphstyle=\color{deepred},    %
stringstyle=\color{deepgreen},
frame=single,                         %
showstringspaces=false
}}

\lstnewenvironment{python}[1][]
{
\pythonstyle
\lstset{#1}
}
{}

\newcommand\pythoninline[1]{{\pythonstyle\lstinline!#1!}}

\makeatletter
\def\mathcolor#1#{\@mathcolor{#1}}
\def\@mathcolor#1#2#3{%
  \protect\leavevmode
  \begingroup
    \color#1{#2}#3%
  \endgroup
}
\makeatother

\Crefformat{equation}{#2Eq.\;(#1)#3}

\Crefformat{figure}{#2Figure #1#3}
\Crefformat{assumption}{#2Assumption #1#3}
\Crefname{assumption}{Assumption}{Assumptions}

\usepackage{crossreftools}
\title{Agentopia: Long-Term Life Simulation and Learning in Agent Societies}

\author[1]{Xintao Wang}
\author[2]{Sirui Zheng}
\author[2]{Hongqiu Wu}
\author[1]{Weiyuan Li}
\author[3]{Jen-tse Huang}
\author[2]{Minghao Zhu}
\author[2]{Can Zu}
\author[2]{Qi Deng}
\author[4]{Jiawei Wang}
\author[1]{Qianyu He}
\author[2]{Heng Wang}
\author[2]{Xiaojian Wu}
\author[2]{Yunzhe Tao}

\affil[1]{Fudan University}
\affil[2]{Independent Researcher}
\affil[3]{Johns Hopkins University}
\affil[4]{University of Science and Technology of China}

\correspondingauthor{Corresponding author: xtwang21@m.fudan.edu.cn}

\begin{abstract}
\textbf{Abstract:} Humans learn from social life.
Simulating this process with LLM-powered agents represents a promising research direction, raising a natural question: whether LLMs can learn from such simulated social experience to better understand and replicate human behavior.
However, prior agent society simulations typically operate at the scale of days, limiting the depth of social interactions and long-term growth.
In this paper, we study long-term life simulation and LLM learning in agent societies, with two goals:
\textit{(1)} investigating social behaviors that emerge from life-long simulation, and
\textit{(2)} developing anthropomorphic capabilities in LLMs, particularly intelligence in social life, through years of simulated social experience.
Specifically, we present \method, a comprehensive framework for long-term life simulation in multi-agent societies, where 100 agents autonomously pursue personal growth, develop social relationships, and fulfill their needs and goals over 10 simulated years.
We define \textit{life reward} to mirror human well-being, and leverage this reward to train LLMs via rejection sampling. 
Extensive experiments show that agents exhibit rich emergent social behaviors.
Furthermore, life reward training effectively enhances the underlying LLM,
which leads to improved agent well-being in simulation, 
and generalizes to downstream role-playing benchmarks with +15.6\% improvement.
Our code is available at \href{https://github.com/Neph0s/Agentopia}{https://github.com/Neph0s/Agentopia}.

\end{abstract}

\let\origunderscore\_
\renewcommand{\_}{\origunderscore\allowbreak}

\begin{document}

\maketitle

\begin{figure*}[h]
  \centering
  \includegraphics[width=0.9\textwidth]{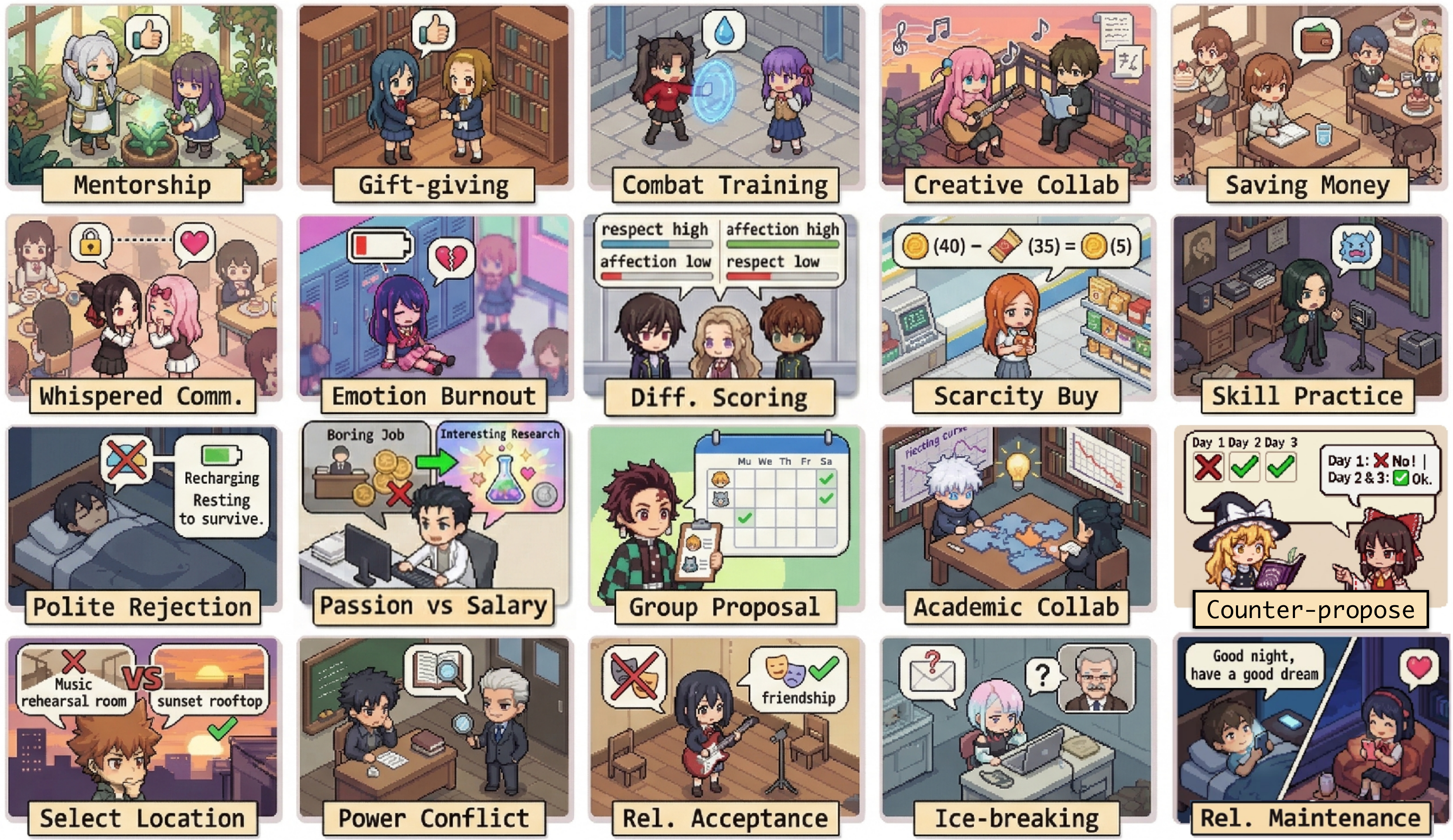}
  \caption{Illustration of emergent behaviors observed in \method simulations.
  Each scene depicts a real behavioral pattern documented in our case studies (Tables~\ref{tab:planning}--\ref{tab:emergent}).
  Without explicit scripting, agents autonomously develop diverse behavioral patterns reflecting agents' intelligence in social life.
  }
  \label{fig:emergent-behaviors}
\end{figure*}

\vspace*{-10pt}

\section{Introduction}
\label{sec:intro}

Humans learn from social environments~\citep{vygotsky1978mind}, and large language models (LLMs) learn from humans.
As LLMs advance, how to better understand and simulate human thought, emotion, and behavioral patterns has become an important research problem, which is known as LLM-based persona simulation, or role-playing~\citep{chen2024from}.
LLM role-playing is widely adopted in AI companions, digital games, and content creation~\citep{shao2023character, zhou2023characterglm}, such as \textit{Whispers from the Star}.~\footnote{\url{https://wfts.anuttacon.com/}}
However, faithfully aligning LLMs with human mind remains a challenge.
As human data approaches exhaustion,
future AI agents need to learn primarily through experience~\citep{silver2025era}.
This motivates a straightforward idea:
since humans learn and grow through social lives in human societies~\citep{maslow1943theory},
\textit{can agents do the same --- learning, growing, and becoming more human-like through lives in agent societies?}
This requires building an agent society
that supports long-term life simulation,
encourages social behaviors such as growth, competition, intimate relationships, and resource allocation,
and provides rewards that mirror human well-being for optimization.

Prior work has studied LLM-based social simulation and persona simulation in short-term settings.
For social simulation,
previous efforts have built prototypes of agent societies,
\eg, Generative Agents~\citep{park2023generative} and Aivilization~\citep{fan2026aivilization}.
However, these works primarily focus on low-level operations
such as \dq{collecting 2 wheat to craft 1 flour},
instead of long-term social dynamics, keeping simulations at the scale of days.
For persona simulation,
existing efforts focuses on role-playing within individual conversations~\citep{chen2024from, shao2023character},
emphasizing anthropomorphism, character fidelity, and user engagement, 
rather than life simulation of agents themselves.
These methods rely heavily on human data for optimization~\citep{wang2025coser, liu-etal-2025-cogdual, du2026her}, which is costly to collect and difficult to scale up, and thus remain insufficient to deeply align LLMs with human cognition.
Overall, existing research only study simulation at the scale of days or single conversations,
leaving long-term life simulation unexplored.

\begin{figure*}[t]
    \centering
    \includegraphics[width=\textwidth]{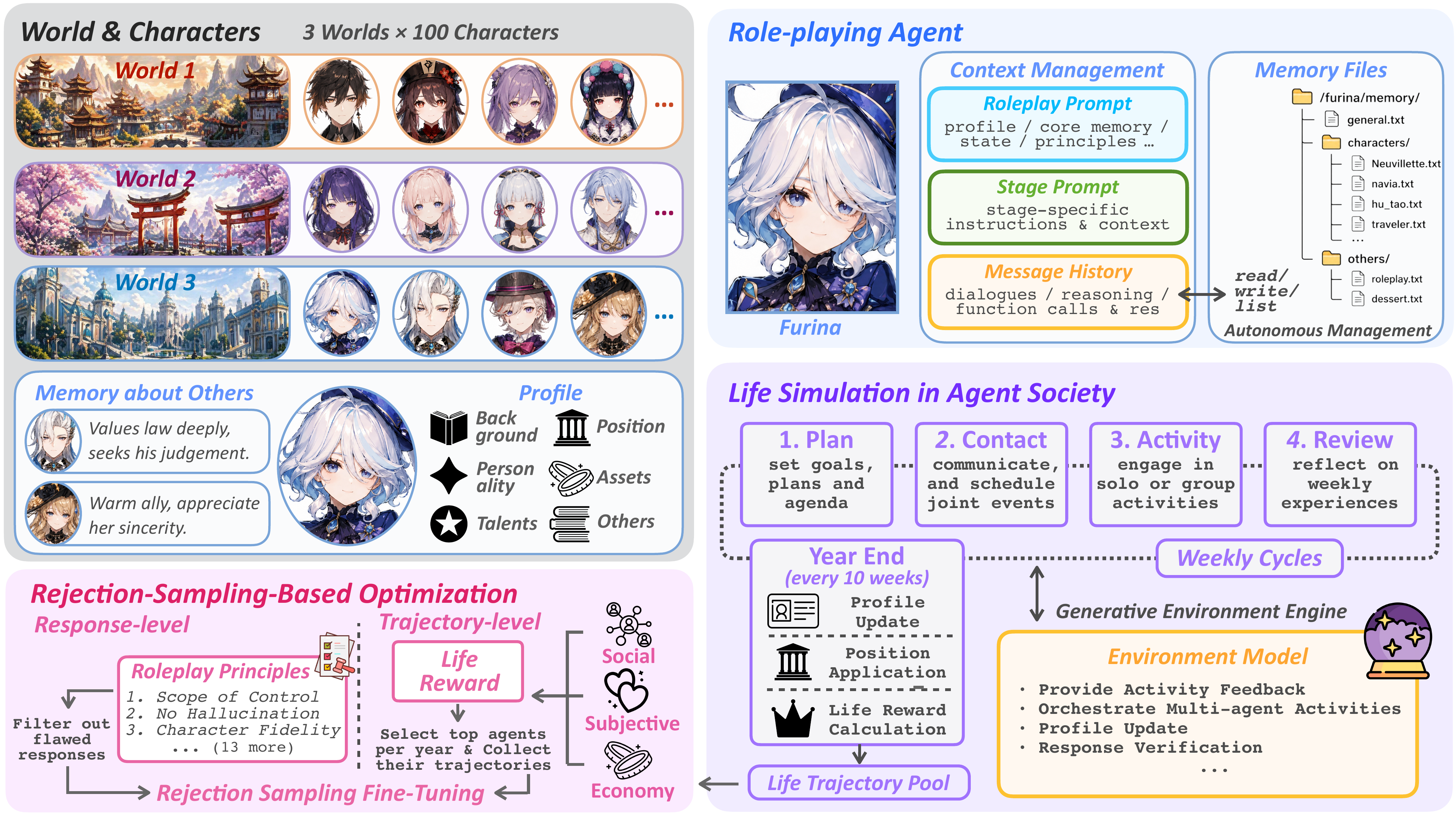}
    \caption{Overview of \method: world and character construction (\S~\ref{app:creation}), role-playing agent (\S~\ref{sec:agent_design}), life simulation in agent society (\S~\ref{sec:simulation_procedure}), and life reward training (\S~\ref{sec:training_data}).}
    \label{fig:framework}
\end{figure*}

In this paper, we present \method,
a framework for long-term life simulation in multi-agent societies,
where agents autonomously simulate years of human social lives, as illustrated in Figure~\ref{fig:framework}.

In \method, agents build social relationships,
develop personal skills,
set and accomplish goals,
develop and fulfill needs,
and engage in economic activities.
\method focuses on social interactions rather than low-level operations.
The framework is characterized by several key features:
\textit{(1)} \method defines \textit{life reward} to model human well-being,
representing social standing, subjective fulfillment, and economic status.
\textit{(2)} \method employs an \textit{environment model} to serve as a generative environment engine that orchestrates the simulation,
avoiding the need to hard-code massive rules.
It serves multiple purposes including
verifying agent responses against role-playing principles,
providing feedback on agent behaviors, and creating and scheduling events.
\textit{(3)} \method implements a comprehensive \textit{context management} mechanism to provide agents with sufficient context, and equips agents with \textit{memory files}, a file-system-based long-term memory that agents autonomously manage what to remember, update, or discard.
Furthermore, leveraging simulation and reward in \method, we propose \textit{life reward training} optimize LLMs via rejection sampling,
improving LLMs in anthropomorphism, role-playing ability, and intelligence in social life.

Towards long-term life simulation,
\method introduces a carefully designed simulation procedure.
It aims to model as many real-world social interactions as possible within a unified framework.
The core challenge is that LLMs generate by turns, 
while humans can perceive and act at any time.
Hence, we structure time into discrete units, allowing concurrent yet ordered interactions.
\method uses the \textit{week} as its basic time unit.
Each week consists of four stages:
\textit{(1)} \textit{Plan},
\textit{(2)} \textit{Contact} others and arrange schedules,
\textit{(3)} carry out solo or group \textit{Activity} over subsequent days,
and \textit{(4)} \textit{Review} the week's experiences.
The \textit{year} serves as a larger cycle.
At the end of each year, 
\method updates agents' profiles , 
allows agents to apply for new careers, 
and calculates life rewards for them.

To validate this framework,
we create three diverse fictional worlds and conduct simulations.
Each world contains 100 agents running for 10 simulated years.
This scale is orders of magnitude larger than prior studies.
We conduct multi-faceted analyses of these simulations,
such as reward and behavior analysis,
social relationship evolution,
and social mobility.
We also conduct extensive case studies to observe emergent behaviors in agent societies, as shown in Figure~\ref{fig:emergent-behaviors}.
Besides, we use \method as an arena to compare different models' performance.
Subsequently, we demonstrate the effectiveness of life reward training.
The trained model demonstrates improved overall well-being in simulation,
including improved social relationships, higher subjective fulfillment, and better economic gains.
Furthermore, the improvements generalize to enhanced anthropomorphism and role-playing ability,
achieving +15.6\% performance gain on CoSER Test~\citep{wang2025coser}, a downstream role-playing benchmark.

Our contributions are summarized as follows:
\begin{enumerate}
  \item We introduce \method, a system for long-term life simulation in agent societies.
        Compared to prior work,
        \method extends the scale of life simulation from days to years for the first time,
        enabling long-term social dynamics 
        such as personal growth, relationship building, life planning, and, \etc
  \item Based on \method, we define \textit{life reward} to mirror human well-being, and propose \textit{life reward training} that fine-tunes LLMs on high-advantage agent experiences, without relying on human data.
       
  \item We conduct extensive experiments on \method,
        including comprehensive analyses and case studies on agents' social behaviors.
        We also validate the effectiveness of life reward training,
        which improves in-simulation well-being
        as well as anthropomorphism and role-playing ability in downstream evaluation.
\end{enumerate}

\section{Related Work}
\label{sec:related}

\paragraph{Agent-based Social Simulation}
Recent work has explored LLM-powered agents for social simulation.
Generative Agents~\citep{park2023generative} pioneers this direction by simulating a society of 25 agents for two days,
observing emergent social behaviors such as organizing a party.
Humanoid Agents~\citep{wang2023humanoid} extends this by incorporating human needs inspired by Maslow's hierarchy of needs~\citep{maslow1943theory}.
Project Sid~\citep{altera2024projectsid} grounds agents in an established digital game environment, \textit{Minecraft},
showing emergent specialization, collective rule formation, and cultural transmission.
BookWorld~\citep{ran-etal-2025-bookworld} constructs multi-agent systems from fictional works,
enabling story generation through character interactions.
Aivilization~\citep{fan2026aivilization} designs a game environment to simulate agents' production and economic behaviors.
CAMEL~\citep{li2024camel} investigates role-playing multi-agent systems for collaborative problem-solving in mathematics and coding tasks.
However, existing systems such as Aivilization face limitations in enabling long-term life simulation, where most LLM calls are spent on \textit{low-level operations}, \ie, \dq{physical} interactions within the virtual environment (\eg, picking up a bottle and moving between locations), rather than long-term social interactions between agents.
Table~\ref{tab:simulation-comparison} provides a systematic comparison of these systems across key dimensions.

\begin{table*}[h]
\centering
\small
\setlength{\tabcolsep}{4pt}
\caption{Comparison of \method and existing agent society systems.
``Time Scale'' denotes the temporal scope of a simulation run.
``Env.\ FB'' indicates whether environmental feedback is rule-based or LLM-generated.
``Econ.\ Sys.'' denotes whether the system includes items and economic mechanisms.
``Career Sys.'' refers to whether agents can choose and change occupations over time.
``Group Interact'' means the system supports interactions among more than two agents.
Project Sid's numbers are from Section 4.2 of its paper.
}
\label{tab:simulation-comparison}
\resizebox{\textwidth}{!}{%
\begin{tabular}{lcccccc}
\toprule
& \textbf{Gen.\ Agents} & \textbf{Humanoid} & \textbf{Project Sid} & \textbf{BookWorld} & \textbf{Aivilization} & \textbf{Agentopia} \\
& \citep{park2023generative} & \citep{wang2023humanoid} & \citep{altera2024projectsid} & \citep{ran-etal-2025-bookworld} & \citep{fan2026aivilization} & (ours) \\
\midrule
Time Scale      & Days   & Days   & Days  & --- & Weeks  & Years  \\
Num.\ Agents       & 25     & ---    & 50 & ---   & ${\sim}$10,000     & 100    \\
Action Space    & Predefined & Predefined & Predefined & Free-form & Predefined & Free-form \\
Env.\ FB        & Rule & Rule & Rule & LLM & Rule & LLM \\
Memory          & Retrieval     & Retrieval & Retrieval & Retrieval  & Fixed Scratchpad & File-based \\
Skill Growth    & \xmark & \xmark & \xmark & \xmark & \cmark & \cmark \\
Econ.\ Sys.     & \xmark & \xmark & \cmark & \xmark & \cmark & \cmark \\
Career Sys.     & \xmark & \xmark & \cmark & \xmark & \cmark & \cmark \\
Group Interact  & \cmark & \xmark & \cmark & \cmark & \xmark & \cmark \\
Reward  & \xmark & \xmark & \xmark & \xmark & \cmark & \cmark \\
Training       & \xmark & \xmark & \xmark & \xmark & \xmark & \cmark \\
\bottomrule
\end{tabular}%
}
\end{table*}

\paragraph{Persona Simulation and LLM Role-Playing}
This line of research aims to improve LLMs' anthropomorphism and role-playing ability~\citep{chen2024from}.
\citet{shanahan2023role} argue that LLMs are inherently role-playing,
whether acting as general assistants or portraying specific characters.
\citet{wang2025coser} propose that training on diverse characters enables LLMs to generalize anthropomorphic abilities,
bridging persona simulation and role-playing research.
Prior studies focus on data, evaluation, and optimization:
For data,
Character-LLM~\citep{shao2023character} and RoleLLM~\citep{wang2023rolellm} apply LLMs to synthesize character-specific question-answer data.
CoSER~\citep{wang2025coser} extracts authentic character data from literary corpora,
obtaining large-scale, high-quality role-playing datasets.
For evaluation,
previous studies primarily employ LLM judges~\citep{wang2023rolellm, tu2024charactereval}
and introduce rubrics aligned with human preferences~\citep{wang2025coser, du2026her}.
Additionally, RoleEval~\citep{shen2023roleeval}, InCharacter~\citep{wang2024incharacter},
and LifeChoice~\citep{xu2024character}
evaluate LLMs objectively from the perspectives of knowledge, personality, and decision-making, respectively.
For optimization, 
prior research applies supervised fine-tuning with human data~\citep{wang2025coser, zhou2023characterglm}
or synthetic data~\citep{chan2024personahub, lu2024large}
CogDual~\citep{liu-etal-2025-cogdual}, CPO~\citep{ye-etal-2025-cpo}, and HER~\citep{du2026her}
explore reinforcement learning for role-playing,
where obtaining high-quality, human-aligned reward signals remains a key challenge.
Besides, Persona Vector~\citep{chen2025persona}, Assistant Axis~\citep{lu2026assistant},
and Emotion Concepts~\citep{sofroniew2026emotion}
explore the internal mechanisms of persona and emotion in LLMs,
proposing methods for detection and manipulation.

\section{\method}
\label{sec:framework}

This section presents the design of \method,
a unified framework for long-term life simulation in agent societies.
The framework is organized around three core concepts: the simulation procedure, the agent, and the environment model:
\textit{(1)} For the simulation procedure, agents live through weekly cycles, each consisting of four stages: Plan, Contact, Activity, and Review, which jointly support diverse social behaviors.
\textit{(2)} For the agent, the core challenge is context management, \ie, providing each LLM with sufficient context including the character's persona, states, relationships, and memories. 
Agents' long-term memory is implemented as a file system that agents autonomously manage via function calls. 
\textit{(3)} To orchestrate the simulation, \method introduces the environment model, a separate LLM that organizes events, provides environmental feedback, and drives the simulation forward, replacing hard-coded rules.
We describe agent design in \S~\ref{sec:agent_design},
the simulation procedure in \S~\ref{sec:simulation_procedure},
contact and scheduling in \S~\ref{sec:contact},
activities in \S~\ref{sec:activity},
and environment design in \S~\ref{sec:environment}.

\subsection{Agent Design}
\label{sec:agent_design}

In \method, each agent role-plays as a specific persona, which comprises a \textit{profile}, \textit{social relationships}, and \textit{dynamic states}, comprehensively representing the character.

\begin{enumerate}
    \item \textbf{Profile}:
    The profile stores an agent's identity information, including a background story, personality traits, talents, as well as initial values for positions and assets. 
    These attributes remain relatively stable during simulation and are only updated once per simulated year.
    The full profile schema is provided in \S~\ref{app:schema}, and the character creation process is described in \S~\ref{app:creation}.
    
    \item \textbf{Social Relationships}:
    In \method, relationships are modeled entirely via inter-character memory.
    We do not explicitly define or store character relationships.
    Instead, relationships are represented through each character's memory of the other, stored as free text and autonomously managed by agents.
    This design uniformly represents friends, lovers, strangers, rivals, and other relationship types without requiring separate mechanisms.
    Relationships can be unidirectional and expanded through multi-person activities.
    The relationship network is initialized during world creation (\S~\ref{app:creation}) and subsequently managed autonomously by agents.
    
    \item \textbf{Dynamic States}:
    Dynamic states represent an agent's current life conditions and evolve with activities.
    They include vitality, fulfillment, skills, position, and assets.
    \textit{Vitality} represents energy level. 
    \textit{Fulfillment} is based on Maslow's hierarchy of needs~\citep{maslow1943theory}, which captures agents' subjective satisfaction from four dimensions: mood, material, social, and esteem.
    It serves as the basis for subjective rewards.
    Following hedonic adaptation theory~\citep{Headey2014}, fulfillment decays naturally each week.
    Fulfillment and vitality changes are determined by the environment model based on activity content, rather than hard-coded rules.
    The details are provided in \S~\ref{sec:appendix_states}.
    \textit{Skills} represent an agent's acquired abilities.
    \textit{Position} represents an agent's current occupation, which provides weekly income and skill growth (\S~\ref{sec:appendix_enrollment}).
    \textit{Assets} consist of a deposit and a list of possessions (\S~\ref{app:economy}).
    
\end{enumerate}

Besides, \method carefully designs a context management mechanism to provide agents with sufficient context across stages, and equips agents with memory files as long-term memory that agents autonomously manage.
Together, \method provides rich context for LLMs to simulate human social lives.

\begin{enumerate}
    
    \item \textbf{Context Management}:
    \method carefully designs a comprehensive context management mechanism to provide agents' underlying LLMs with sufficient information to simulate human social lives.
    This is achieved through three context layers:
    \textit{(1)} \textbf{Roleplay prompt}: provides the foundational role-playing information shared across all stages.
    It contains the agent's full persona, recent weekly diaries as short-term memory, summaries of key memory files as long-term memory, worldview rules and role-play principles, \etc,
    enabling agents to correctly accumulate memory across stages and weeks throughout the simulation.
    \textit{(2)} \textbf{Stage prompt}: provides stage-specific instructions and context, \eg, scheduling rules and communication history  for the Contact stage.
    \textit{(3)} \textbf{Message history}: records LLMs' multi-turn messages accumulated within the current stage, including prior dialogues, calls and results of memory functions, and compacted reasoning process.
    Together, these three layers combine automatic and agent-driven context management, providing comprehensive context for agents to simulate human social lives.
    The details are elaborated in \S~\ref{sec:appendix_context}.

    \item \textbf{Memory Files}:
    While weekly diaries serve as short-term memory by recording recent experiences, memory files provide agents with \textit{long-term memory}, a file-system-based persistent store that agents autonomously manage via function calls.
    Agents decide what to remember, update, or discard, giving them full control over what persists across weeks and years.
    Each agent maintains memory files in three categories:
    \texttt{general.txt} for personal notes and plans,
    \texttt{characters/<who>.txt} for the agent's knowledge of and relationship with specific people,
    and \texttt{others/<name>.txt} for any other topics.
    Memory files provide context to LLMs in two ways: (1)
    summaries of recently accessed files are \textit{automatically} included in the roleplay prompt, 
    and (2) agents \textit{autonomously} retrieve full content on demand, included in the message history.
    Agents manage these files through three functions: \texttt{read\_file} to retrieve content, \texttt{update\_file} to write new content, and \texttt{list\_files} to list all files.
    A read-before-write constraint is enforced: an update is only permitted after the agent has read the target file in the same invocation, ensuring updates build upon existing content rather than overwriting blindly.
    Agents can also create new files for topics they wish to track.

    \end{enumerate}
\subsection{Simulation Procedure}
\label{sec:simulation_procedure}

A fundamental constraint of LLM-based agents is that they generate responses turn by turn:
they receive a complete context and produce a single reply, unlike humans who continuously perceive and react in real time (\eg, interrupting a conversation mid-sentence).
The simulation procedure must therefore be designed around turn-based interaction.
Meanwhile, \method focuses on abstract social interactions (\eg, planning, socializing, decision-making) rather than low-level operations (\eg, movement, object manipulation).
This abstraction enables the system to model richer social behaviors with fewer LLM calls, achieving higher token efficiency for long-term simulation.

With these considerations, we design a comprehensive simulation procedure.
\method uses the \textit{week} as its basic time unit.
A simulated year spans $n_w$ weeks; year-end triggers profile updates, position application, and life reward calculation.~\footnote{%
The \textit{day}, \textit{week}, and \textit{year} in \method serve as three hierarchical layers of abstract simulation time.
They are designed to simulate human life cycles at comparable scales, without strictly corresponding to real-world calendar units.}
Each week consists of four stages: Plan, Contact, Activity, and Review.

\begin{enumerate}

\item \textbf{Plan}: 
Based on memory and current state, each agent makes a weekly plan.
Agents are able to review previous plans and store or update new ones through function calls (\eg, \texttt{read\_file}, \texttt{update\_file}).
Agents also select an abstract consumption level (\ie, living standard) for the week, which abstracts their monetary spending and influences material fulfillment.

\item \textbf{Contact (\& Scheduling)}: 
In this stage, agents communicate with each other over multiple rounds.
One primary purpose of communication is to arrange shared schedules: agents can propose invitations, which others may accept or decline, thereby creating joint activity schedules.
The detailed mechanism is introduced in \S~\ref{sec:contact}.

\item \textbf{Activity}: 
In this stage, agents carry out diverse activities.
Each weekly cycle contains $n_d$ active days, and on each day, every agent carries out exactly one activity.
Activities fall into four types: joint, solo, encounter, and public.
Through these activities, agents can pursue personal growth, work, or leisure on their own, or engage with others in shared experiences.
Activity types and execution are detailed in \S~\ref{sec:activity}.

\item \textbf{Review}: 
At the end of each week, agents reflect on their weekly experiences and summarize them into a weekly diary.
Based on this reflection, agents can update their memory files via function calls.
The weekly diaries are included in each agent's context in subsequent weeks, allowing agents to recall their past experiences.

\end{enumerate}

At the end of each simulated year, three processes take place.
\textit{(1) Profile Update:} the environment model makes incremental updates to each agent's profile (\eg, personality traits, talents) based on their accumulated experiences over the year.
\textit{(2) Position Application:} agents may apply for new positions based on their preferences and compete for limited vacancy based on their abilities and skills.
\textit{(3) Life Reward Calculation:} life rewards are calculated, measuring each agent's social standing, subjective fulfillment, and economic status over the past year.

\subsection{Contact \& Scheduling}
\label{sec:contact}

The Contact stage serves two purposes: communication between agents and scheduling of activities.
Communication is restricted to pairwise exchanges, while multi-person conversations are deferred to joint activities.
Constrained by the turn-based nature of LLM generation, communication in \method proceeds in rounds.
Each week consists of $n_c$ contact rounds.
In each round, an agent receives newly arrived messages along with the contact history from the past three weeks, then decides whom to contact and what actions to take.
After all rounds conclude, the system resolves the contact history to determine which joint activities have been successfully created.

\paragraph{Actions}
In \method, agents must generate \textit{actions} to trigger real communication and scheduling.
Actions are wrapped with \texttt{<role\_action>} tags in agents' responses, which the system then parses and executes.
They are different from function calls.
Four action types are available: \texttt{contact}, \texttt{propose\_joint\_activity}, \texttt{respond\_invitation} and \texttt{cancel\_joint\_activity} (\S~\ref{sec:appendix_contact_details}).

\paragraph{Schedule Resolution} After all contact rounds conclude, the system resolves which joint activities are successfully created based on all collected actions.
The process handles cancellations, deduplicates responses, resolves time conflicts, and checks creation condition. 
The full rules and procedure are described in \S~\ref{sec:appendix_procedure}.

\paragraph{Scheduling Public and Encounter Activities}
Public and encounter activities are scheduled in dedicated stages before and after Contact, respectively.
Before Contact, the environment model creates public events for the upcoming weeks, and agents sign up for those that match their interests.
After Contact, the environment model arranges encounter activities to create chance meetings for idle agents, \ie, those without any scheduled activity on a given day.

\subsection{Activities}
\label{sec:activity}

Activities are the core stage of \method, where agents carry out activities and develop abilities, fulfillment, wealth, and social relationships.
\method supports four activity types: joint, solo, encounter, and public.
Agents carry out activities according to their schedules, defaulting to solo activities when unscheduled.
After each activity, the environment model provides feedback based on the agent's experience (including state changes), and the agent reflects on the outcome and may update its memory files accordingly.
The four activity types are described below:

\begin{enumerate}
\item \textbf{Joint Activity}:
Joint activities model multi-agent, multi-turn interactions.
They are created through invitation and negotiation during the Contact stage.
Participants take turns speaking; following~\citet{wang2025coser}, we employ the environment model to provide environmental descriptions and select the next speaker each turn.
Agents can choose the visibility of their generated content, using special tags to mark each segment as public, private, or selective (visible only to specified persons).
Agents are also permitted two actions: \texttt{gift} to transfer items to another participant, and \texttt{exit} to leave the activity early.
Additionally, we introduce a response filtering mechanism, where the environment model evaluates each agent response based on a set of roleplay principles covering anthropomorphism, character fidelity, and feasibility, filtering out responses that violate any principle.
Details about joint activity execution and response filtering are provided in \S~\ref{sec:appendix_activity} and \S~\ref{sec:appendix_principles}, respectively.

\item \textbf{Solo Activity}:
Solo activity serves as the default when an agent has no other schedule, allowing activities such as studying, working, or leisure and consumption.
It follows a single-turn format: the agent describes its intended action, and the environment model evaluates the feasibility of the action based on the character's background and provides an outcome.
If the agent's intention does not match its abilities, the environment model provides negative feedback.
Solo activities also support spending, where agents may purchase goods or services to gain material fulfillment (\S~\ref{app:economy}).
Details about solo activity execution are described in \S~\ref{sec:appendix_solo_activity}.

\item \textbf{Encounter Activity}:
Encounters are chance meetings arranged by the environment model for idle agents, and serve as a special case of joint activity.
They follow the same multi-turn dialogue format as joint activities, but do not appear in agent schedules.
Encounters are designed to let the environment model create meaningful chance meetings: for example, introducing strangers to expand social networks, or bringing together agents with significant plot connections or special relationships.

\item \textbf{Public Activity}:
Public activities represent open community events centered around shared interests (\eg, workshops, interest groups).
They are designed by the environment model, and agents may sign up independently based on their own interests.
The execution generally follows a similar flow to solo activity.
The key difference is that, at the end of the activity, agents see what other participants did, and may get to know them.
Details for public activities are described in \S~\ref{sec:appendix_public_activity}.

\end{enumerate}

\subsection{Environment Design}
\label{sec:environment}

This section introduces the environment design in \method, including the environment model, the economy system, the position application mechanism, and the location system.
These components serve as essential parts that support the comprehensive simulation of social lives in agent societies.

\paragraph{Environment Model}
Building a comprehensive social simulation environment to appropriately orchestrate the simulation and provide environmental feedback for role-playing agents is a complex yet important task.
In \method, we introduce the environment model, a stateless LLM, to handle these functions instead of hard-coding extensive rules.
During activities, it provides feedback to agent behavior, judging feasibility and evaluating outcomes.
Besides, it predicts next speakers in joint activities, generates public and encounter activities, ranks candidates for position applications, updates agent profiles at year-end, and performs response filtering to exclude low-quality outputs.
These functions collectively support the smooth progression of life simulation in the multi-agent society.
Hence, the environment model essentially serves as a generative environment engine, providing intelligent, generative responses to agent behaviors.
Our prompts for the environment model are displayed in Tables~\ref{tab:prompts_god_joint}, \ref{tab:prompts_god_eval}, and~\ref{tab:prompts_god_world}.

\paragraph{Economy System}
\method includes a basic economy system where agents earn and spend money.
Income comes from three sources: a position-based weekly salary, character-specific weekly income, and additional work during the activity stage.
Expenses take two forms.
First, agents select a living standard each week (from frugal to luxurious), representing the richness of their material life for that week.
Second, agents may choose to engage in consumption activities during solo activities (\eg, shopping, entertainment).
Through spending, agents gain material fulfillment, which contributes to improvement in subjective reward.
Meanwhile, accumulating more wealth leads to higher economy reward (\S~\ref{app:economy}).

\paragraph{Position System}
Each agent holds a position that represents its job or social role.
A position provides two weekly benefits: a fixed income and skill growth in relevant abilities.
During character construction, the environment model assigns each agent an initial position based on the world setting and character profiles.
Once per year, a position application process allows agents to decide whether to change their position, and the environment model determines outcomes based on agents' abilities and position requirements.
The details about position and position application are described in \S~\ref{sec:appendix_enrollment}.

\paragraph{Location System}
\method implements a simple location mechanism.
Its primary purpose is to provide agents with grounded environmental perception.
In our early experiments, we find that agents tend to hallucinate environmental details and objects without location information explicitly provided.
Hence, we employ the environment model to create a set of locations for each world, and require joint activities and encounter activities to specify a location.
Details are provided in \S~\ref{sec:appendix_location}.

\section{Reward and Optimization}
\label{sec:reward}

This section presents life reward and the training method built upon it.
The design of life reward draws from human well-being as a prior to define agent rewards, guiding agents toward human-like goals.
As LLMs optimize toward these goals, they are expected to naturally develop emotional intelligence, social intelligence, and life wisdom, similar to how humans grow through social lives.

\subsection{Life Reward}
\label{sec:life_reward}

At the end of each simulated year, life reward is calculated for every agent.
Grounded in Maslow's hierarchy of needs~\citep{maslow1943theory}, life reward in \method is defined with three dimensions: social, subjective, and economy.
Social reward measures an agent's social standing based on perception from other agents.
Subjective reward measures an agent's fulfillment over the past year.
Economy reward measures an agent's yearly financial gain.
The three dimensions are determined or estimated by the external environment, rather than self-reported by agents.

\paragraph{Social Reward}
Social reward is computed from how others in an agent's social circle perceive them.
Based on the Warmth-Competence model~\citep{fiske2007universal}, we consider two dimensions of perception: \textit{affection} and \textit{respect}, corresponding to warmth and competence respectively.
Each agent is asked to independently score every person in their social circle on both dimensions, using a 0 to 100 scale.
Agents are told that these ratings are private and will not be revealed to others, preventing agents from giving dishonest ratings due to social pressure.
Then, these scores are sorted and rescaled to 0 to 100 based on the ranking (first place receives 100, last place receives 0), eliminating differences in scoring scales across agents.
Based on the scores, we construct two weighted directed graphs for affection and respect respectively, where the scores serve as edge weights.

We apply Weighted PageRank~\citep{page1999pagerank} to compute each agent $i$'s social standing score $S_i$ on each graph.
Inspired by Sociometer Theory~\citep{leary2012sociometer}, we add a Mutual Affection Bonus after PageRank convergence to obtain the final score:
\begin{equation}
S_i' = \sum_{j \in \mathcal{N}_{in}(i)} w_{ji} \cdot (1 + \alpha \cdot w_{ij}) \cdot S_j
\end{equation}
where $S_j$ is agent $j$'s raw PageRank score, $\mathcal{N}_{in}(i)$ means the set of agents who know $i$, $w_{ji}$ denotes the normalized edge weight from $j$ to $i$, and $\alpha$ is the mutual affection coefficient.
This mechanism models the psychological effect that \dq{being valued by those I value} matters more, amplifying the contribution from reciprocated relationships.

We denote the scores for affection and respect as $S_{\text{aff}}'$ and $S_{\text{resp}}'$ respectively, and take their average as the social reward:
$r_{\text{social}} = \frac{1}{2}S_{\text{aff}}' + \frac{1}{2}S_{\text{resp}}'$.

\paragraph{Subjective Reward}
Subjective reward is computed from an agent's fulfillment history over the past year.
It is primarily measured across four fulfillment dimensions: mood, material, social, and esteem.
Additionally, a penalty mechanism is introduced to penalize agents with excessively low fulfillment or vitality.
Specifically, for each fulfillment dimension, the 25th percentile across all agents in each week serves as the threshold.
Agents falling below this threshold receive a penalty.
Similarly, vitality below its threshold is penalized independently.
The final subjective reward is:
\begin{equation}
r_{\text{subj}} = \frac{\sum_{w=1}^{n_w} \sum_{d=1}^{D} f_{w,d} - n_p \cdot \lambda_p}{n_w \cdot D}
\end{equation}
where $n_w$ is the number of weeks in each year, $D$ is the number of fulfillment dimensions,
$f_{w,d}$ is the fulfillment value for week $w$ and dimension $d$,
$n_p$ is the total number of penalty instances over the year,
and $\lambda_p$ is the penalty weight.

\paragraph{Economy Reward}
Economy reward measures an agent's objective financial gain over the past year,
computed as $r_{\text{econ}} = \text{deposit}_{\text{end}} - \text{deposit}_{\text{start}}$, capturing both earning ability and spending wisdom.

\paragraph{Total Reward}
Total reward combines the three reward dimensions.
Since the three dimensions have different scales, we apply z-score normalization to each independently, and compute total reward $r$ as a weighted sum of the normalized scores:
\begin{equation}
r = \lambda_{\text{social}} \cdot z_{\text{social}} + \lambda_{\text{subj}} \cdot z_{\text{subj}} + \lambda_{\text{econ}} \cdot z_{\text{econ}}
\end{equation}
where $\lambda_{\text{social}}$, $\lambda_{\text{subj}}$, and $\lambda_{\text{econ}}$ are the weights for the three rewards.

\subsection{Life Reward Training}
\label{sec:training_data}

To optimize LLMs towards improved well-being in social life simulation, we propose life reward training, a rejection-sampling-based optimization method.
The challenge lies in the extremely long-horizon nature of social simulation, which makes end-to-end algorithms like PPO~\citep{schulman2017proximal} infeasible in this setting.
Each agent's trajectory involves hundreds of LLM calls per simulated year, and a full simulation spans tens of such years.
Therefore, we adopt a rejection-sampling approach that selects high-advantage trajectories based on life reward (\S~\ref{sec:life_reward}).

\paragraph{Estimating Returns and Advantages}
Given the life reward defined in \S~\ref{sec:life_reward}, we estimate returns and advantages for each agent at each time step \(t\) (\ie, a simulated year). For agent \(i\), the return is defined as the discounted sum of future rewards:
\[
G_{i,t} = \sum_{k=0}^{T-t} \gamma^k r_{i,t+k},
\]
where \(\gamma\) is the discount factor and \(T\) is the final time step.

Estimating advantages requires a baseline. In our life simulation setting, it is difficult to learn a critic as in PPO~\citep{schulman2017proximal} or to estimate a baseline by averaging over multiple rollouts as in GRPO~\citep{guo2025deepseek}, since each agent produces a single life trajectory. We therefore use each agent's own previous return as a self-referential baseline.

Before computing advantages, we normalize returns across time steps to make them more comparable, as raw returns may have different effective horizons and distributions; details are provided in \S~\ref{sec:appendix_sft}. Let \(G^{norm}_{i,t}\) denote the normalized return. We define the advantage as:
\[
A_{i,t} = G^{norm}_{i,t} - G^{norm}_{i,t-1}.
\]
This formulation measures whether an agent's expected future life reward improves relative to its own past, rather than comparing absolute performance across agents.
The advantage measures each agent's improvement relative to its own past, rather than cross-agent comparison.
It therefore reduces the tendency to favor agents with inherently better initial conditions, whose high absolute returns may reflect design-time advantages rather than behavioral quality.

\paragraph{Trajectory Selection}
Within each reward period, we select the top 25\% of agents by advantage, \ie, those who improved the most over the past year.
Once an agent is selected, all of its trajectories within that period are included as training data.
Since advantage measures improvement relative to one's own past rather than absolute standing, the selected trajectories represent beneficial behaviors for diverse personas and backgrounds, rather than converging toward homogeneous behavioral patterns.

\paragraph{Response Filtering}
Response filtering is applied as an additional quality filter to exclude low-quality responses from training data.
Agent responses with malformed actions or invalid parameters (\eg, referencing a non-existent character) are filtered out.
Also, agent responses in joint and encounter activities are checked against a set of 16 roleplay principles (Table~\ref{tab:principles}) covering anthropomorphism, character fidelity and reasonableness, with violating responses filtered out. %

To prevent catastrophic forgetting, we employ self-distillation~\citep{lu2025onpolicy}.
We sample the model's own responses to general-purpose instructions from the Tulu 3 dataset~\citep{lambert2024tulu}, and mix them with \method trajectories at a 50:50 ratio, which are measured by output tokens.
Detailed training settings are provided in \S\ref{sec:appendix_sft}.

\section{Experiments}
\label{sec:experiments}

\subsection{Experimental Setup}
\label{sec:exp_setup}

\paragraph{World Data}
We design three worlds with distinct social settings to study agents' behaviors in different social environments.
Each world contains 100 uniquely designed agents with diverse backgrounds,
personalities, and initial social relationships.
The details of the world design and character creation are introduced in \S~\ref{app:creation}.
The three worlds are summarized as follows:
\begin{itemize}
  \item \textbf{The Apartment}: a shared-living residential building in New York City,
        inhabited by young professionals, students, and artists.
        It highlights how strangers organically form community bonds
        and relationships within a shared living space.
  \item \textbf{Arcane Academy}: a magical academy setting
        where students and faculty navigate both academic and interpersonal challenges.
        It highlights how complex relationships develop
        within a structured academic institution.
  \item \textbf{The Campus}: a Chinese high school setting,
        centering on student and teacher interactions in a contemporary educational environment.
        It highlights school social network formation
        and personal growth trajectories.
        Its simulation runs in Chinese.
\end{itemize}

\paragraph{Simulation}
We conduct simulations across the three worlds, running for 10 simulated years.
Each simulated year consists of $n_w$ weeks, and each week consists of $n_c$ contact rounds and $n_d$ activity days.

\paragraph{Models}
We use Qwen3.5-397B-A17B~\citep{qwen35} as the primary model
for both the agents and the environment model.
When a model fails to produce valid outputs,
we fall back to Gemini 3 Flash\footnote{\url{https://ai.google.dev/gemini-api/docs/models/gemini-3-flash-preview}} for re-generation.
For more details on generation parameters, see \S~\ref{app:config}.

\paragraph{Metrics}
We track a set of analytical metrics to describe agent behaviors and social dynamics,
covering rewards, fulfillment, activity behaviors, contact behaviors,
personal growth, and computational cost.
Full metric definitions are provided in \S~\ref{sec:appendix_metrics}.

\subsection{Analysis of Long-Term Social Simulation}
\label{sec:long_term_analysis}

\begin{figure*}[htbp]
  \centering
  \includegraphics[width=\textwidth]{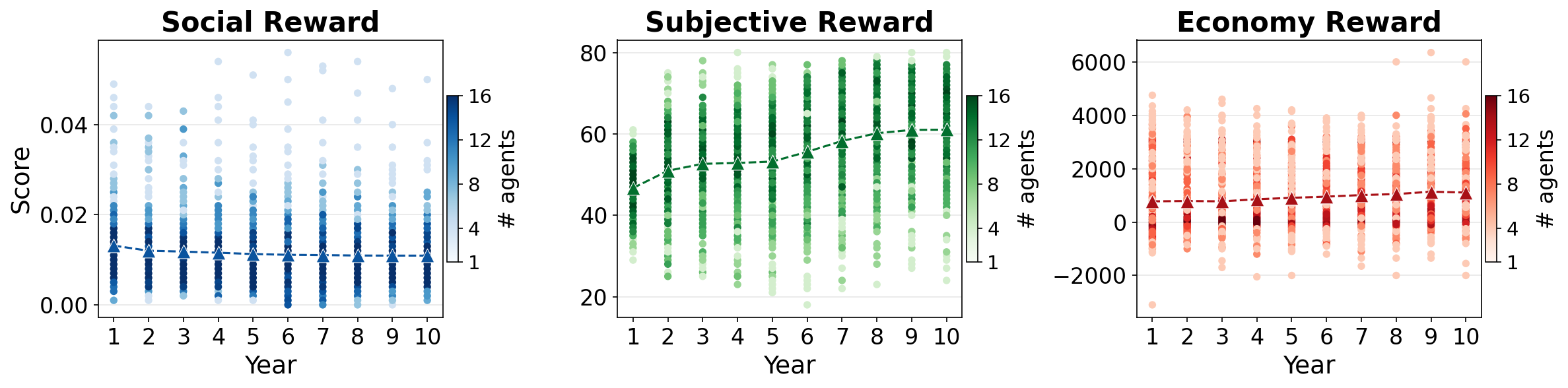}
  \caption{%
    Distribution of three reward dimensions over 10 simulated years, pooled across all three worlds.
    Triangles mark annual means.
  }
  \label{fig:reward_distribution}
\end{figure*}

\paragraph{Reward Analysis}
We analyze the distribution and temporal trends of the three rewards across
all three 10-year simulations (100 agents $\times$ 10 years $\times$ 3 worlds = 3{,}000 observations).
Figure~\ref{fig:reward_distribution} presents the pooled distributions,
where color intensity indicates agent density
and triangles mark annual means.
Per-world breakdowns are provided in \S~\ref{sec:appendix_reward_per_world}.

Both subjective and economy rewards show upward trends on average,
while social reward remains broadly stable,
as it is based on agents' relative rankings rather than absolute performance.

\begin{figure}[htbp]
  \centering
  \includegraphics[width=\textwidth]{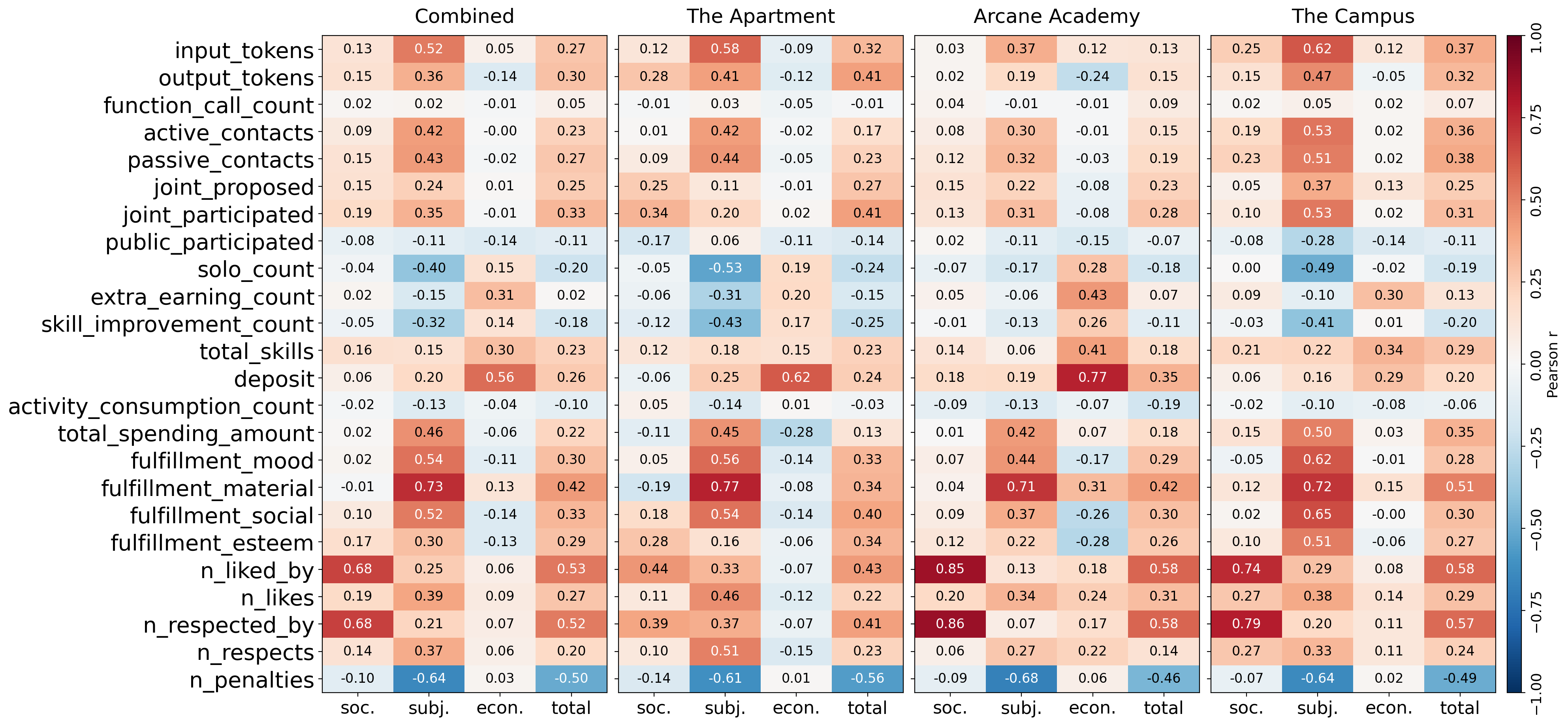}
  \caption{%
    Pearson correlation heatmap between 24 metrics and reward dimensions.
    The \textit{Combined} column averages per-world values.
  }
  \label{fig:reward_behavior_correlation}
\end{figure}

\paragraph{Reward--Behavior Correlation}
To understand what drives each reward dimension,
we compute Pearson correlations between behavioral metrics and the three rewards
for each world separately, then average the per-world coefficients to obtain combined values.
Figure~\ref{fig:reward_behavior_correlation} presents the full correlation matrix.

We highlight four findings:
(1) Total reward correlates most strongly with fulfillment dimensions,
social evaluation metrics, and token consumption,
while penalties show the strongest negative signal ($r = -0.50$);
(2) social reward correlates almost exclusively with reputation metrics
(respected by and liked by, $r = 0.68$), which is natural by design; 
it also correlates with numbers of active and passive contacts ($r = 0.09$ and $r = 0.15$)
as well as numbers of proposed and participated joint activities ($r = 0.15$ and $r = 0.19$),
suggesting that agents who engage more in social interactions gain higher social reward;
(3) subjective reward is driven by four fulfillment dimensions by design---mood
($r = 0.54$), material ($r = 0.73$), social ($r = 0.52$), and esteem ($r = 0.30$)---and
unmet fulfillment needs trigger penalties ($r = -0.64$), which substantially reduce subjective reward;
it also correlates with metrics about contacts, joint activities and spending;
(4) economy reward is primarily determined by deposit accumulation ($r = 0.56$)
and extra earning counts, with skills as an indirect positive factor.
Detailed analysis is provided in \S~\ref{sec:appendix_reward_correlation}.
Additional analyses are provided in the appendix, including
friend distribution (\S~\ref{sec:appendix_friend_distribution}),
social network visualization (\S~\ref{sec:appendix_social_network}),
inter-reward correlations (\S~\ref{sec:appendix_reward_correlation_inter}),
wealth inequality (\S~\ref{sec:appendix_matthew_effect}),
reward quartile profiles (\S~\ref{sec:appendix_reward_success_profile}),
striving vs.\ leisurely agent profiles (\S~\ref{sec:appendix_ambition_profile}),
cross-world divergence (\S~\ref{sec:appendix_cross_world}),
and model comparison (\S~\ref{sec:appendix_model_comparison}).

\subsection{Training via Life Rewards}
\label{sec:sft}

\paragraph{Training}
We apply life reward training (\S~\ref{sec:training_data}) to fine-tune Qwen3.5-397B-A17B on simulation data from the first four years across all three worlds.
Training is conducted for 1 epoch on 30 nodes of 8$\times$H100 80GB GPUs with a learning rate of $1 \times 10^{-5}$.
Full training details are provided in \S~\ref{sec:appendix_sft}.
We refer to the fine-tuned model as Qwen3.5-397B-Agentopia.

Qwen3.5-397B-Agentopia is then evaluated via 4-year simulations
on \campus{} and \apartment{},
and compared against the original Qwen3.5-397B baseline
under identical world configurations.

\paragraph{Learning via life rewards improves agent performance in social simulation}
\begin{table}[t]
\caption{%
Qwen3.5-397B-Agentopia vs.\ Qwen3.5-397B: cross-world average metrics over four simulated years on \campus{} and \apartment{}.
Delta shows the percentage change of the four-year average from Qwen3.5-397B to Qwen3.5-397B-Agentopia.}
\label{tab:sft_cross_world}
\centering
\small
\setlength{\tabcolsep}{3pt}
\begin{tabular}{l rrrr rrrr rrr}
\toprule
& \multicolumn{2}{c}{Y1} & \multicolumn{2}{c}{Y2} & \multicolumn{2}{c}{Y3} & \multicolumn{2}{c}{Y4} & \multicolumn{3}{c}{Avg} \\
\cmidrule(lr){2-3} \cmidrule(lr){4-5} \cmidrule(lr){6-7} \cmidrule(lr){8-9} \cmidrule(lr){10-12}
Metric & Base & Tuned & Base & Tuned & Base & Tuned & Base & Tuned & Base & Tuned & Delta \\
\midrule
\multicolumn{12}{l}{\textit{Reward Components}} \\
Economy Reward & 1108 & 1271 & 1092 & 1130 & 1038 & 1006 & 1069 & 1008 & 1077 & 1104 & $\uparrow$\,2.5\% \\
Subjective Reward & 46.0 & 46.7 & 50.1 & 48.4 & 51.5 & 52.6 & 51.9 & 55.3 & 49.9 & 50.8 & $\uparrow$\,1.8\% \\
\midrule
\multicolumn{12}{l}{\textit{Social Evaluation}} \\
Respected By & 7.1 & 6.9 & 9.4 & 11.0 & 10.4 & 14.1 & 11.2 & 15.2 & 9.5 & 11.8 & $\uparrow$\,24.2\% \\
Liked By & 5.2 & 5.1 & 6.9 & 7.5 & 7.4 & 9.3 & 8.1 & 9.9 & 6.9 & 8.0 & $\uparrow$\,15.9\% \\
Mutual Respect & 5.2 & 5.1 & 7.3 & 7.9 & 8.1 & 9.9 & 8.3 & 10.1 & 7.2 & 8.3 & $\uparrow$\,15.3\% \\
Mutual Like & 4.1 & 3.9 & 5.2 & 5.3 & 5.5 & 6.4 & 5.6 & 6.0 & 5.1 & 5.4 & $\uparrow$\,5.9\% \\
\midrule
\multicolumn{12}{l}{\textit{Fulfillment}} \\
Material Fulfill. & 43.7 & 35.6 & 41.4 & 32.4 & 41.8 & 37.2 & 46.0 & 41.8 & 43.2 & 36.8 & $\downarrow$\,$-$14.8\% \\
Mood Fulfill. & 76.9 & 78.3 & 88.3 & 86.9 & 89.1 & 91.4 & 88.7 & 92.8 & 85.8 & 87.4 & $\uparrow$\,1.9\% \\
Social Fulfill. & 59.3 & 60.9 & 66.9 & 69.9 & 63.9 & 73.4 & 64.8 & 75.2 & 63.7 & 69.9 & $\uparrow$\,9.7\% \\
Esteem Fulfill. & 42.3 & 44.1 & 44.0 & 44.4 & 43.8 & 46.4 & 43.6 & 47.2 & 43.4 & 45.5 & $\uparrow$\,4.8\% \\
\midrule
\multicolumn{12}{l}{\textit{Activity Patterns}} \\
Joint Proposed & 6.1 & 6.0 & 6.6 & 5.8 & 6.3 & 6.1 & 6.8 & 6.9 & 6.5 & 6.2 & $\downarrow$\,$-$3.9\% \\
Joint Act. & 12.3 & 12.2 & 13.3 & 11.7 & 12.7 & 12.3 & 13.8 & 14.0 & 13.0 & 12.6 & $\downarrow$\,$-$3.6\% \\
Public Act. & 12.1 & 11.2 & 9.1 & 9.5 & 8.2 & 8.7 & 5.4 & 7.9 & 8.7 & 9.3 & $\uparrow$\,7.1\% \\
Solo Act. & 17.4 & 22.4 & 17.9 & 14.5 & 20.3 & 12.6 & 21.5 & 12.4 & 19.3 & 15.5 & $\downarrow$\,$-$19.8\% \\
Skill Advances & 11.9 & 11.3 & 11.3 & 9.2 & 14.7 & 8.2 & 15.1 & 8.6 & 13.3 & 9.3 & $\downarrow$\,$-$29.6\% \\
\bottomrule
\end{tabular}
\end{table}

Table~\ref{tab:sft_cross_world} summarizes the cross-world average metrics,
from which we observe that:
\begin{inparaenum}[\it (1)]
\item \textbf{Rewards.}
Both economy reward (+2.5\%) and subjective reward (+1.8\%) improve on average,
indicating that life reward training effectively enhances agents' overall well-being in social simulation;
\item \textbf{Social recognition.}
Social reward, being rank-based, remains unchanged when all agents improve uniformly.
We instead examine social evaluation metrics (\S~\ref{sec:appendix_metrics}), which measure absolute social recognition as the numbers of peers who like or respect an agent above a threshold.
The results show that Qwen3.5-397B-Agentopia agents are respected by 24.2\% more peers and liked by 15.9\% more on average,
earning broader recognition and more friendships;
\item \textbf{Fulfillment.}
Mood, social, and esteem fulfillment all improve,
suggesting that trained agents better simulate human needs and need-fulfilling behaviors.
On the other hand, material fulfillment declines ($-$14.8\%),
as the economy reward incentivizes saving over spending,
mirroring the real-life trade-off between consumption and savings;
\item \textbf{Activity patterns.}
Public activity participation rises (+7.1\%),
while solo activities decrease ($-$19.8\%)
and skill advances drop substantially ($-$29.6\%).
This shows that life reward shapes agent behavior, steering agents toward rewarded behavioral patterns, while unrewarded actions may be deprioritized.
\end{inparaenum}
Detailed per-year trends and per-world breakdowns are provided in \S~\ref{sec:appendix_sft_trends}.

\paragraph{Life reward training generalizes to role-playing ability}
Qwen3.5-397B-Agentopia is also evaluated on CoSER Test~\citep{wang2025coser},
a role-playing benchmark that assesses LLMs on given-circumstance acting
derived from classic literary scenarios,
across four dimensions:
Storyline Consistency, Anthropomorphism, Character Fidelity, and Storyline Quality.
It is compared against several state-of-the-art open-source and proprietary models,
using Qwen3-235B~\citep{qwen3} as the judge model
(see \S~\ref{sec:appendix_coser} for details).
As shown in Table~\ref{tab:coser_results},
Qwen3.5-397B-Agentopia achieves significant improvements over the baseline Qwen3.5-397B
and outperforms Claude-4.5-Sonnet,
with the most notable gains in Anthropomorphism (+23.7\%) and Character Fidelity (+16.4\%).
This indicates that training with life rewards does not merely optimize agents within \method,
but also improves general anthropomorphism and role-playing capabilities.

\begin{table}[t]
\caption{%
Performance of various LLMs on CoSER Test, judged by Qwen3-235B.
}
\label{tab:coser_results}
\centering
\small
\setlength{\tabcolsep}{3pt}
\begin{tabular}{l ccccccc}
\toprule
& \makecell{Claude\\4.5-Opus} & \makecell{Gemini\\3-Pro} & \makecell{Qwen3.5\\397B-Agentopia} & \makecell{Claude\\4.5-Sonnet} & \makecell{Qwen3.5\\397B} & \makecell{CoSER\\70B} & \makecell{GPT-5\\Mini} \\
\midrule
Storyline Consistency & 63.74 & 65.95 & 41.02 & 47.18 & 39.60 & 35.05 & 38.10 \\
Anthropomorphism      & 64.28 & 60.42 & 49.67 & 36.02 & 40.16 & 31.16 & 24.60 \\
Character Fidelity    & 58.45 & 58.34 & 46.93 & 47.55 & 40.32 & 32.28 & 27.20 \\
Storyline Quality     & 63.24 & 62.49 & 59.01 & 50.09 & 49.97 & 45.33 & 42.00 \\
\midrule
Average               & 62.43 & 61.80 & 49.16 & 45.21 & 42.51 & 35.95 & 32.97 \\
\bottomrule
\end{tabular}
\end{table}

\subsection{Computational Cost}
\label{sec:cost}

\begin{table}[t]
\centering
\caption{Overall cost for three simulation worlds.
Each world runs 100 agents for 10 simulated years.
M = million tokens, K = thousand calls, h = wall-clock hours.}
\label{tab:cost_overall}
\small
\begin{tabular}{l rrrrr}
\toprule
\textbf{World} & Input (M) & Output (M) & Tokens (M) & Calls (K) & Time (h) \\
\midrule
\campus{}       & 19,041 & 425 & 19,466 & 544 & 201.3 \\
\academy{}      & 11,302 & 315 & 11,617 & 572 & 174.2 \\
\apartment{}    &  9,699 & 317 & 10,016 & 584 & 183.2 \\
\midrule
\textbf{Average} & 13,347 & 352 & 13,700 & 567 & 186.2 \\
\bottomrule
\end{tabular}
\end{table}

\begin{figure}[h]
\centering
\includegraphics[width=\textwidth]{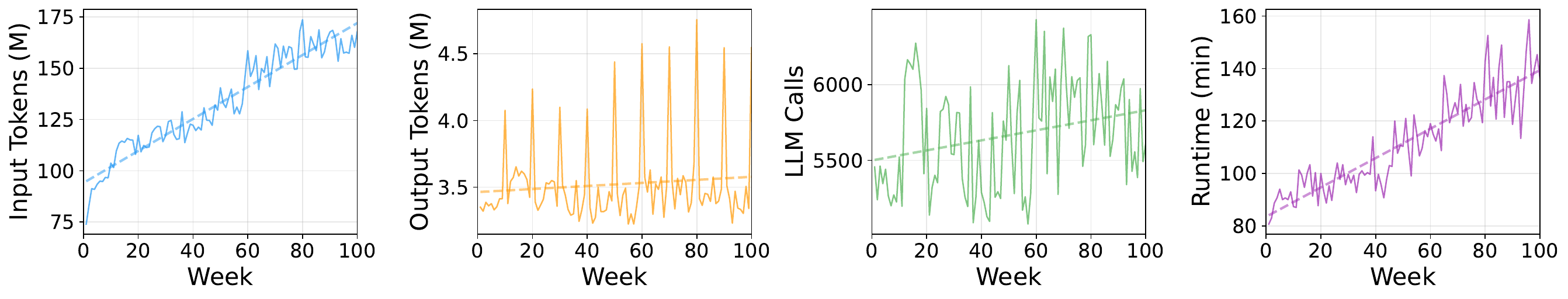}
\caption{Per-week cost trends averaged across three worlds over 10 simulated years.
Dashed lines indicate linear trend fits.}
\label{fig:cost_trend}
\end{figure}

We deploy three FP8 instances of Qwen3.5-397B for each simulation run.
Table~\ref{tab:cost_overall} summarizes the overall cost
for simulating 100 agents over 10 years in each world,
and Figure~\ref{fig:cost_trend} illustrates the per-week cost trends
averaged across three worlds.

On average, a single 10-year simulation consumes 13.7 billion tokens
across 567K LLM calls, completing in approximately 186 wall-clock hours.
Input tokens dominate the cost,
averaging 133M per week compared to only 3.5M for output tokens,
reflecting the heavy context required for each agent's persona and memory.
Both input tokens and runtime increase steadily over time
as agents accumulate memory and context grows,
with per-week runtime rising from approximately 80 to 140 minutes
over the 10-year span.
This indicates that memory growth serves as a major computational bottleneck for long-horizon social simulation, which makes effective context and memory management strategies essential.     
Detailed per-world breakdowns are provided in \S~\ref{sec:appendix_cost}.

\section{Conclusion}
\label{sec:conclusion}

This paper studies long-term life simulation and LLM learning in agent societies. 
We present \method, a multi-agent society where agents live their own lives.
We define life reward to model agents' well-being, and propose life reward training, a rejection-sampling-based method that selects high-advantage agent trajectories to optimize LLMs.

We conduct simulations across three diverse worlds, each with 100 agents running for 10 simulated years.
Agents exhibit diverse emergent behaviors, including personal growth, relationship formation, and life choices.
Life reward training leads to improved well-being in simulation, including broader social recognition, higher subjective fulfillment, and better economic gains.
Through evaluation on downstream role-playing benchmarks, we validate that life reward training improves LLMs' anthropomorphism and role-playing ability, yielding a 15.6\% overall improvement on CoSER Test, with the largest increases in anthropomorphism (+23.7\%) and character fidelity (+16.4\%).

\section*{Limitations}

In this paper, we present \method, a comprehensive framework for agent social life simulation that models a wide range of human social behaviors.
As an agent environment, its complexity goes beyond existing simulation systems.
However, building a agent system that comprehensively simulates human social life is extremely challenging, and our system has room for improvement in several aspects.

\paragraph{Turn-based design}
Our system is built on turn-based LLM generation, which fundamentally differs from real-time human perception and reaction.
As discussed in \S~\ref{sec:appendix_design_philosophy}, real-time perception would consume prohibitive computation on low-level operations, drastically reducing the density of social interactions and making long-term simulation infeasible.

\paragraph{Hallucination}
Human behaviors are naturally grounded by the physical world, but LLM-powered agents operate through text generation with few constraints, making them susceptible to hallucination, such as fabricating non-existent characters or locations.
\method mitigate this through the context management and the location system (\S~\ref{sec:appendix_location}), and employ principle verification to filter hallucinated outputs (\S~\ref{sec:appendix_principles}), but fully eliminating hallucination remains an open challenge.

\paragraph{Environment and numeric systems}
\method's environment includes predefined numeric systems and a generative environment engine, but fully aligning them with real-world societies is extremely difficult.
It is challenging for the environment model to produce responses that perfectly mirror real-world outcomes, and the numeric mechanisms governing agent states are difficult to align with human behavioral and physiological patterns.

\paragraph{Optimization and alignment}
While training with life reward yields notable improvements, two alignment gaps remain.
First, we cannot ensure that the life reward objective fully aligns with human well-being.
The simulated environment cannot fully reproduce the complexity of real-world societies.
Moreover, the reward signals themselves (\eg, subjective fulfillment) are difficult to fully align with human physiological and cognitive mechanisms.
Second, \method is fundamentally a society of agents rather than a human society: all feedback an agent receives comes from other AI models, not from humans, leaving it an open question whether the trained LLMs can align with human cognitive and psychological patterns.

\paragraph{Computational constraints}
Life simulation in \method is computationally expensive, and our limited resources prevent us from conducting more comprehensive experiments.
For example, we did not conduct simulations with more worlds, more agents, or longer time horizons, nor trained on different model families.
Also, for trajectory selection, we selected all generations from the chosen year and agent, without exploring fine-grained credit assignment strategies that attribute reward to specific responses.
We hope to address these with more computational resources in future work.

\section*{Impact Statement}

\method aims to advance research in agent society simulation and LLM-based role-playing.
All characters and worlds in our simulations are fictional and do not represent real individuals or groups.
Our methods could potentially be applied to simulate real-world individuals, and such applications must strictly respect privacy regulations and obtain necessary consent.

\bibliography{main}

\clearpage
\appendix
\section{World and Character Creation}
\label{app:creation}

\subsection{Pipeline Overview}
\label{app:creation_process}

Each simulation world is designed as a self-contained community resembling a small town,
where most characters know one another and the social network is intentionally dense.
This closed-community design serves a critical purpose:
it ensures that every person referenced in a character's backstory, memories, or relationships is probably an interactable agent within the simulation,
preventing agents from attempting to interact with non-existent entities.

Each world is constructed through a five-step pipeline:
(1) world setting design,
(2) reference story selection,
(3) incremental character generation,
(4) attribute assignment,
and (5) consistency verification.
Steps~(1)--(2) are manually crafted by the authors,
steps~(3)--(4) are driven by LLMs~\footnote{In this paper, we employ Claude Opus 4.6 for these steps.},
and step~(5) employs Claude Code~\footnote{https://claude.com/product/claude-code} for automated quality assurance~\footnote{In this paper, we employ Claude 4.6 Opus for Claude Code in this step.}.

\paragraph{Step 1: World Setting}
For each simulation world, we manually craft a \textit{worldview document} that defines the social structure, geographic environment, temporal setting, and thematic focus.
For example, \campus{} is set in a top-tier Chinese high school in 2020, emphasizing academic pressure, campus friendships, and adolescent growth;
\apartment{} depicts a shared-living scenario among young professionals;
and \academy{} adapts a fantasy school setting with magical elements.
Each worldview specifies the society structures, social norms, and everyday concerns that ground the agents' behavior.
Importantly, the setting is designed as a closed community so that all characters an agent might reference probably exist within the world.

\paragraph{Step 2: Reference Story Selection}
To encourage diverse and dramatically rich character dynamics, we curate a set of reference stories adapted from well-known literary works and films.
These references serve as creative seeds---the LLM is instructed to \textit{adapt} rather than replicate them, ensuring original character designs that exhibit realistic interpersonal tensions.
Each reference is a 100 to 200 word synopsis covering key relationships and conflicts (\eg, unrequited affection, friendship rivalry, mentor-student dynamics).

\paragraph{Step 3: Incremental Character Generation}
Characters are generated incrementally using Claude Opus 4.6 with temperature 1.0.
The process begins by creating two seed characters together with their mutual relationship descriptions.
Subsequently, characters are added one at a time: for each new character, the LLM receives the worldview, an example character template, summaries of all existing characters, and the current relationship matrix.
When the population exceeds 10, we randomly sample 6 relationship edges from the existing matrix to keep the prompt within a manageable length while preserving structural context.
The LLM outputs a structured YAML profile for each character---including demographic information, appearance, personality traits (qualitative), skills, values, conflicts, and core motivations---along with updated bidirectional relationship descriptions.
Each relationship entry describes one character's perception of another, covering appearance evaluation, ability assessment, personality impression, perceived relationship role, and known facts, within approximately 100 words.
The relationship descriptions are directly stored as each character's initial memory about the other characters it knows
(see \S\ref{app:schema} for storage details),
so that every agent begins the simulation with pre-existing social knowledge about its social circle.
To ensure name diversity, character names are drawn from a pre-curated name list tailored to each world's cultural setting (\eg, German-style names for the medieval world, Chinese names for the campus world), and the LLM is explicitly prohibited from reusing names.
Each character's values and motivations are derived by the LLM from the character's backstory, occupation, and personality, ensuring internal coherence.
We repeat this process until each world contains 100 characters.

\paragraph{Step 4: Attribute Assignment}
The character profiles generated in Step~3 contain qualitative attribute levels (\eg, personality ``High'', skill ``Proficient'').
In this step, we convert these qualitative descriptions into the quantitative values required by the simulation system.
Personality traits and talents are mapped to a 0 to 100 scale via range-based random sampling
(\eg, ``Low'' $\to$ [10, 25], ``Average'' $\to$ [55, 65], ``High'' $\to$ [90, 100]).
Initial skills use a five-tier mapping onto a 0 to 300 scale:
Untrained $\to$ [0, 10], Beginner $\to$ [15, 30], Some Experience $\to$ [35, 95], Proficient $\to$ [100, 295], and Master $\to$ 300.
The skill scale is substantially wider than that of personality and talents because skills grow continuously through practice during the simulation with no hard upper bound
(see \S\ref{app:numeric} for the full numeric system).
Additionally, economic attributes such as initial deposits are assigned based on each character's position and income level.
Initial deposits are capped at $20\times$ the fixed weekly income, approximating 20 weeks of savings.

\paragraph{Step 5: Consistency Verification}
After generation, we conduct a systematic quality review using Claude Code.
This automated verification identifies and resolves three categories of issues:
(1) \textit{Currency unit inconsistency}---during generation, some characters' economic data contain non-standard currency units
instead of pure numeric values; all monetary fields are normalized to a unified numeric format.
(2) \textit{Duplicate characters}---the LLM occasionally produces two characters with highly similar backstories, occupations, or personality profiles; such duplicates are identified and either merged or regenerated.
(3) \textit{Memory inconsistency across characters}---this includes
(a) contradictions between A's perception of B and B's perception of A (\eg, A considers B a close friend while B describes A as a stranger),
(b) conflicting accounts of a third party (\eg, A and B give incompatible descriptions of C's background), and
(c) references to characters who do not exist in the world.
For case~(c), when a character's backstory mentions a person outside the 100-agent population, we revise the description to indicate that this person has left the community and is no longer reachable, thereby preventing the agent from attempting to contact a non-existent entity.
All identified issues are flagged and corrected before the simulation begins.

\subsection{Details of Character Data and Memory System}
\label{app:schema}

Each character's data is organized as a directory tree under \texttt{persona/\{name\}/}, comprising a \textbf{profile} and a \textbf{memory system}.
Figure~\ref{fig:schema} shows an abridged example from the \academy{} world.

The \textbf{profile} (\texttt{profile/year=YYYY.json}) contains:
\textit{name}, \textit{gender}, \textit{birthday};
\textit{appearance\_and\_impression} ($\sim$70 words describing the first impression an outsider would form);
\textit{talents} (9 dimensions shared across all characters, each with a character-specific numeric value on a 0 to 100 scale: beauty, communication, creativity, health, honesty, integrity, intelligence, leadership, and trustworthiness);
\textit{personality\_traits} (10 dimensions shared across all characters, each with a character-specific numeric value on a 0 to 100 scale: confidence, control, curiosity, empathy, extraversion, feeling, intuition, judging, patience, and responsibility);
\textit{skills} (a dictionary mapping skill names to numeric proficiency on a 0 to 300 scale);
\textit{details} (detailed backstory), \textit{conflicts}, \textit{core\_motivation}, \textit{values}, and \textit{preferences};
\textit{position} (role, organization, weekly income);
and \textit{init\_assets} (initial deposit and possessions).

The \textbf{memory system} organizes each agent's memories across multiple files, which we call \textit{memory files}.
Each memory file is implemented as a JSONL file where new memory entries are continuously appended.
At initialization, the system pre-populates a \texttt{characters/} subdirectory containing one memory file per known character,
storing the agent's knowledge of that person (appearance, abilities, personality, relationship, etc.).
Beyond these initial files, agents can freely manage their own memory system during the simulation---they may create new memory files
to record various types of information (\eg, personal plans, general notes, working memory), even though these files do not exist at initialization.
All memory files are dynamically updated as agents interact and accumulate new experiences.

\begin{figure}[h]
\centering
\small
\begin{verbatim}
persona/Cecilia Fairmont/
+-- profile/year=2020.json
|   +-- name: "Cecilia Fairmont"
|   +-- gender: "Female", birthday: "Y2003-W8-activity-D4"
|   +-- appearance_and_impression: "Standing 167cm ... pale
|   |   ash-blonde ... projects the impression of a brilliant,
|   |   tightly wound young woman ..."
|   +-- talents: {intelligence: 90, creativity: 85,
|   |            leadership: 30, communication: 65, ...}
|   +-- personality_traits: {curiosity: 100, control: 90,
|   |                       introversion: 100, patience: 85, ...}
|   +-- skills: {Transfiguration: 170, Charms: 170,
|   |           Arithmancy: 170, Herbology: 15, ...}
|   +-- position: {role: "Student, Seventh Year",
|   |             organization: "The Academy / The Scholar House"}
|   \-- init_assets: {deposit: 85, possessions: [...]}
\-- memory/files/
    +-- characters/          (pre-initialized)
    |   +-- Adelaide Hawthorne.jsonl
    |   +-- Alistair Rowan.jsonl
    |   \-- ...
    +-- general.jsonl         (agent-created)
    \-- working_memory.jsonl  (agent-created)
\end{verbatim}
\caption{Abridged character data schema, illustrated with Cecilia Fairmont from \academy{}. Files under \texttt{characters/} are pre-initialized; \texttt{general.jsonl} and \texttt{working\_memory.jsonl} are autonomously created by agents during the simulation.}
\label{fig:schema}
\end{figure}

\section{System Design Details}
\label{app:system_design}

\subsection{Design Philosophy of the Simulation Procedure}
\label{sec:appendix_design_philosophy}

Our goal is to simulate the full range of human social behaviors as naturally and richly as possible.
However, the central challenge is that LLMs operate in a turn-based generation mode and cannot perceive or interact in real time as humans do.

A natural starting point would be a unified-timeline framework where agents act freely and perceive their surroundings in real time.
For example, when an agent moves from a classroom to a playground, it would observe every other agent present and what they are doing at each moment.
However, maintaining such real-time perception requires prohibitive computational cost, making long-term social simulation infeasible.
More fundamentally, such a framework would disproportionately focus on low-level perception and physical operations (\eg, movement, object manipulation), leaving less capacity for the social behaviors we actually care about (\eg, planning, socializing, decision-making).

We therefore design a simulation framework that embraces the turn-based nature of LLMs, starting from two foundational functions: \textit{communication} and \textit{activity}.

\paragraph{Activity Simulation}
Our primary focus is on multi-agent interactions.
We define multi-agent interactions as activities with explicit, specific themes, inspired by the given-circumstance acting paradigm in CoSER~\citep{wang2025coser}.
These activities are either created through invitations between agents (\ie, joint activities), or  arranged by the environment model (\ie, encounter activities).

\paragraph{Communication Simulation}
We design a multi-round communication framework where messages are exchanged in a text-message style.
In each round, every agent simultaneously sees the messages sent to it in the previous round, then decides whom to contact and what to say.
This avoids the need to simulate real-time perception, making computational cost significantly more manageable.
Within this framework, we design action types such as \texttt{propose\_joint\_activity}, \texttt{cancel\_joint\_activity}, and \texttt{respond\_invitation}, along with corresponding resolution logic, to model the complex dynamics of social scheduling in real life.

\paragraph{Synchronous vs.\ Asynchronous Design}
An alternative design is an asynchronous framework where different agents can be in different stages simultaneously, \eg, one agent communicating while another is carrying out an activity.
However, this introduces substantial design complexity.
For example, if an agent in the communication stage sends a message to another agent currently in the activity stage, should the recipient receive it immediately? When should it decide to reply?
After careful consideration, we adopt a synchronous framework: all agents proceed through the same stage together.
During the Contact stage, all agents communicate; during the Activity stage, all agents carry out activities.
This greatly simplifies the system design while preserving rich social behavior modeling.

Building on this foundation, we add the Plan and Review stages to give agents space for planning and reflecting on their social lives,
introduce solo and public activity types to diversify the range of experiences,
and create the position, location, and economy systems to enrich the simulation environment.
Together, these components form the complete simulation procedure described in \S~\ref{sec:simulation_procedure}.

\subsection{Details of Simulation Procedure}
\label{sec:appendix_procedure}

This section provides a complete description of the simulation procedure, expanding on the overview in \S~\ref{sec:simulation_procedure}.
Each simulated year spans $n_w$ weeks, and each week contains $n_d$ days.
The week serves as the scheduling unit, while the day serves as the activity execution unit.
In our experiments, we set $n_w = 10$, $n_d = 5$, and $n_c = 5$, chosen to balance simulation fidelity with computational cost.
Each week proceeds through the following stages in order.

\paragraph{Weekly Settlement (Before Plan)}
Before any agent action, two automatic settlements take place:
\begin{inparaenum}[\it (1)]
\item \textit{fulfillment decay}: mood, social, and esteem each decay by 15\% of their current value, while material fulfillment does not decay (it is controlled by the economy system, see \S~\ref{app:economy});
\item each agent receives its weekly income, including position-based salary and any extra fixed income.
\end{inparaenum}
The decay mechanism forces agents to continuously engage in activities to maintain fulfillment, creating an intrinsic motivation loop.

\paragraph{Plan}
Each agent reviews its goals, reflects on the previous week, makes a weekly plan, and selects a living standard (frugal, moderate, comfortable, or luxurious) that determines weekly expenses and material fulfillment changes (see \S~\ref{app:economy} for details).
Agents can access memory files via function calls to organize their knowledge and intentions.

\paragraph{Before Contact: Public Activity Generation and Sign-up}
Before the Contact stage, the environment model generates public events for the upcoming week(s).
Each event specifies a name, description, start date, duration in weeks (1 = one-time, $>$1 = repeats weekly on the same day for multiple weeks), and an eligibility list (\texttt{"all"} or a named subset of agents).
The number of events generated per week is controlled by configuration.
After generation, each agent is shown only the events they are eligible for and independently decides whether to sign up.

\paragraph{Contact}
The Contact stage runs for $n_c$ contact rounds.
In each round, agents may send messages or arrange joint activities via four action types: \texttt{contact},
\texttt{propose\_joint\_activity}, \texttt{respond\_invitation}, and \texttt{cancel\_joint\_activity}.
After all rounds complete, the system runs schedule resolution to determine which joint activities are
successfully created.
Full details of the communication mechanism, action formats, and schedule resolution are provided in
\S~\ref{sec:appendix_contact_details}.

\paragraph{After Contact: Encounter Activity Generation}
After schedule resolution, the environment model generates encounter activities for idle agents, \ie, those without any scheduled activity on a given day.
The number of encounters per day is $\lfloor n_{\text{idle}} / 5 \rfloor$, with probabilistic rounding of the fractional part.
Encounters occur only at public locations.
For each idle agent, the system retrieves their top 10 most recently interacted characters (via \texttt{get\_top\_related\_names}), and the environment model uses this relationship information to generate meaningful encounter scenarios, pairing agents with existing social connections into the same encounter.

\paragraph{Activity}
Each week contains $n_d$ activity days.
On each day, the system builds activities from agent schedules following the priority order: joint $>$ public $>$ encounter $>$ solo.
Agents without any scheduled activity are automatically assigned solo activities.
All activity types execute in parallel.
Detailed execution mechanisms, including concurrency control, are described in \S\ref{sec:appendix_concurrency}.

\paragraph{Review}
Each agent summarizes its weekly experiences into a diary entry and updates its memory files via function calls, consolidating new knowledge about relationships, plans, and self-reflection.

\paragraph{Weekly Cleanup}
At the end of each week, agents with possessions exceeding the maximum limit (default 50 items) are prompted to discard excess items.
The agent selects which items to discard, preventing unbounded item accumulation.

\paragraph{Year-End Settlement}
At the end of each simulated year, three processes take place in order:
\begin{inparaenum}[\it (1)]
\item the environment model updates each agent's profile (\S~\ref{sec:appendix_profile_update});
\item new positions are introduced and agents compete for positions through position application (\S~\ref{sec:appendix_enrollment});
\item life rewards are calculated (\S~\ref{sec:life_reward}).
\end{inparaenum}

\subsection{Details of Contact Stage}
\label{sec:appendix_contact_details}

\paragraph{Contact Rounds}
The Contact stage runs for $n_c$ contact rounds.
In each round, all agents act in parallel: they read newly arrived messages along with the contact history from the past three weeks, then send messages or arrange activities.
Each agent may issue at most 10 actions per round, encouraging agents to prioritize the most important interactions.

\paragraph{Error Feedback Mechanism}
If an agent's action has formatting errors or violates scheduling rules (\eg, invalid location, non-existent invitee, time outside the allowed window), the system captures the error and feeds it back in the next round's prompt.
The error is formatted as \texttt{"Action: [original action]\textbackslash nError: [error description]"}, allowing the agent to correct its behavior in subsequent rounds.
Actions with errors are treated as invalid and automatically discarded; they are not sent to other agents.

\paragraph{Action Types and Format}
Unlike function calls, actions are wrapped with \texttt{<role\_action>} tags in the agent's text output.
Four action types are available:

\begin{itemize}[nosep]
\item \texttt{contact}: send a text message to a specific person. Parameters: \texttt{to} (recipient name), \texttt{message} (message content).
\item \texttt{propose\_joint\_activity}: propose a joint activity. Parameters: \texttt{activity\_name} (unique name), \texttt{proposal} (description), \texttt{invited\_persons} (list of invitees), \texttt{time} (format: \texttt{Y[year]-\allowbreak W[week]\allowbreak -activity-D[day]}), \texttt{location}, \texttt{required\_participants} (optional list), \texttt{message} (accompanying message).
\item \texttt{respond\_invitation}: accept or decline an invitation. Parameters: \texttt{activity\_name} (matching the proposal), \texttt{to} (proposer name), \texttt{decision} (\texttt{"yes"} or \texttt{"no"}), \texttt{message} (response message).
\item \texttt{cancel\_joint\_activity}: cancel a previously proposed activity. Parameters: \texttt{activity\_name} (matching the proposal), \texttt{message} (cancellation reason, sent to all invitees).
\end{itemize}

Agents may propose activities up to 4 weeks in advance (configurable via \texttt{max\_weeks\_for\_future\_schedule}).
If \texttt{required\_participants} is specified, all listed persons must respond \texttt{"yes"} for the activity to be created; otherwise, at least one acceptance suffices.

\paragraph{Schedule Resolution}
After Contact ends, the system runs schedule resolution over all \texttt{propose\_joint\_activity}, \texttt{respond\_invitation}, and \texttt{cancel\_joint\_activity} actions collected during the week.
The resolution proceeds in the following steps:
\begin{enumerate}[nosep]
\item \textbf{Cancel processing}: if the proposer issued a \texttt{cancel\_joint\_activity}, the proposal is voided and all invitees are notified.
\item \textbf{Response deduplication}: for each invitee, only the last \texttt{respond\_invitation} to a given proposal is kept; earlier responses are discarded.
\item \textbf{Time conflict resolution}: for each agent, if multiple schedules fall on the same day, the system keeps the one with the highest priority and automatically sets the rest to \texttt{"no"} with a reason attached. The priority order is: existing joint activity (from earlier weeks) $>$ newly proposed joint activity $>$ newly accepted joint activity $>$ existing public/encounter activity. Concretely, a newly confirmed joint activity overrides a previously scheduled public or encounter activity, but cannot override an already-confirmed joint activity from an earlier week (first-come-first-served across weeks).
\item \textbf{Activity creation}: for each non-canceled proposal, the activity is created only if \textit{(a)} the proposer has not been removed by a time conflict, \textit{(b)} all \texttt{required\_participants} responded \texttt{"yes"}, and \textit{(c)} at least one invitee responded \texttt{"yes"}.
\end{enumerate}

\subsection{Details of Joint Activities}
\label{sec:appendix_activity}

This section describes the interaction details of joint activities.
Encounter activities are a special case of joint activities and follow the same procedure described below.

\paragraph{Enter Activity}
Before the multi-turn dialogue begins, each participant first performs an analysis of the situation.
The agent is asked to analyze its position, goals, action plan, potential risks, and strategies for the upcoming activity.
This analysis is then injected into the context for the subsequent multi-turn dialogue, helping the agent maintain a coherent strategy throughout the interaction.

\paragraph{Turn Control}
At the start of each turn, the environment model provides an environmental description and selects the next speaker, following CoSER~\citep{wang2025coser}.
Each activity has a minimum and maximum turn limit (default 5 and 20).
The activity ends if the maximum turn count is reached, or the environment model proactively decides to end it after the minimum turn count.
After the activity concludes, the environment model evaluates each participant's experience throughout the multi-turn dialogue and provides feedback, including changes to their states.

\paragraph{Visibility Tags}
In joint and encounter activities, agents can use visibility tags to control who sees their messages:
\begin{itemize}[nosep]
\item \texttt{<private>}: visible only to the speaker. Used for inner thoughts and private actions.
\item \texttt{<visible\_to="A,B">}: visible only to the specified participants. Others receive a notification that a private exchange is occurring, but cannot see the content.
\item Default (no tag): publicly visible to all participants.
\end{itemize}

\paragraph{Gift System}
During joint and encounter activities, agents can transfer items they own to other participants via a \texttt{gift} action.
The system validates that the sender owns the item and the receiver is a current participant, then immediately moves the item from the sender to the receiver.
Both parties receive a system notification confirming the transfer.

\paragraph{Early Exit}
Agents may choose to leave an activity before it concludes via an \texttt{exit\_activity} action.
Once an agent exits, it no longer participates in subsequent turns, and other participants receive a departure notification.
If fewer than 2 participants remain, the activity ends automatically.

\paragraph{Exit Activity (Reflection)}
After an activity concludes and the environment model provides feedback (including state changes), each agent performs an additional step to reflect on the experience.
This mechanism applies to all four activity types, with the reflection content varying by type.
For joint and encounter activities, the agent produces both a summary of the activity process and a personal reflection (mindset shifts, emotional arc) based on the full dialogue history and environment model feedback.
For solo activities, the agent reflects based on its action description and the environment model's outcome feedback.
For public activities, the agent additionally sees what other participants did, and may create or update memory files for participants of interest.
The reflection is recorded in the agent's activity history and serves as context in subsequent weeks.

\subsection{Details of Solo Activity}
\label{sec:appendix_solo_activity}

Solo activity is the fallback when an agent has no other schedule. The agent generates a description of its intended action.
The environment model evaluates the feasibility of the action, considering factors such as the agent's position, skill levels, and age, and determines the degree of success and gains accordingly.
When an agent with insufficient skills attempts a task beyond their capability, the environment model will probably provide a negative feedback with limited progress.
The environment model then determines whether the action involves spending. If so, it generates spending options for the agent to choose from, and the transaction is completed upon selection.
The environment model then evaluates the outcome and provides feedback including changes to vitality, fulfillment, and skills.

\subsection{Details of Encounter Activity}
\label{sec:appendix_encounter_activity}

Encounter activity is a special form of joint activity.
The difference is that, while joint activities are established through agents' proactive interactions during the Contact stage, encounter activities are system-arranged: the environment model pairs idle agents after the Contact stage concludes.
Each encounter involves exactly two agents.
The environment model receives as input the list of idle agents per day (including their known relationships), brief profiles of all idle agents, and the list of available locations.
For pairing, the environment model prioritizes agents who already have a relationship (to deepen existing connections), and otherwise pairs agents who have not yet met (for a first meeting scenario).
The environment model also generates a scene description for each encounter: it describes the objective circumstance of the meeting without assuming characters' thoughts and behaviors, and creates a natural conflict or interesting situation (\eg, ``At the convenience store, both reach for the last bottle of drink on the shelf at the same time.'').
Upon conclusion, mutual recognition is automatically established between participants, and the environment model provides feedback.
The detailed execution procedure is identical to that of joint activities (\S~\ref{sec:appendix_activity}).

\subsection{Details of Public Activity}
\label{sec:appendix_public_activity}

Public activities are generated by the environment model before the Contact stage. Agents may sign up for events that match their interests during Contact.
Each participant executes the activity independently, similar to solo activity, but multiple agents participate in the same event simultaneously.
After the activity concludes, each participant receives a summary of what other participants did during the event.
Based on this information, agents may choose to initiate contact with participants they find interesting.
For each pair that mutually choose to connect, the system creates an initial recognition entry in each agent's memory file for the other.
If only one agent chooses to connect, a unidirectional record is created only in that agent's memory.
The environment model then provides feedback for each participant, including changes to their states.

\subsection{Details of Context Management}
\label{sec:appendix_context}

Table~\ref{tab:context_management} presents the full context composition for each stage of agent LLM calls.
The context is organized into three layers.

The \textbf{roleplay prompt} serves as the system message and is shared across all stages.
It contains the agent's full persona (profile and dynamic state), worldview rules, summaries of key memory files, recent history, current location, commonsense guidelines, roleplay principles, and output requirements.
Memory file summaries are ranked by recency of access (tracked via a hidden access log).
When an agent knows more than 50 characters, only the 50 most recently accessed are shown, but current interaction partners are always included regardless of this limit.
Recent history comprises weekly diary summaries from previous weeks, recent activity records, and the agent's own responses from earlier stages of the current week.

The \textbf{stage prompt} is appended as a user message with stage-specific instructions and context, as detailed in Table~\ref{tab:context_management} and Table~\ref{tab:additional_stages}. 
Table~\ref{tab:additional_stages} lists the prompts for special stages, such as \dq{signup public activity} and \dq{position application}.

The \textbf{message history} records the LLM's input-output messages accumulated during the current multi-round interaction.
It contains three categories of information:
\begin{inparaenum}[\it (1)]
\item prior dialogue turns in multi-round interactions, such as conversation history during joint activities;
\item function call requests and results from the function calling loop;
\item and compacted reasoning summaries for open-source models.
\end{inparaenum}
Message history is cleared between stages.
Cross-stage continuity is instead provided by the recent history component of the roleplay prompt, which includes the agent's summarized outputs from earlier stages of the current week.

\paragraph{Function Calling Loop}
Each LLM invocation runs up to 8 rounds of function calling.
In rounds 1 through 7, \texttt{tool\_choice} is set to \texttt{auto}, allowing the agent to freely invoke functions (\eg, \texttt{read\_file}, \texttt{update\_file}) or produce a final answer.
In the last round, \texttt{tool\_choice} is forced to \texttt{none}, requiring a text response.

\paragraph{Compacted Reasoning}
For open-source models with explicit reasoning traces, multi-round function calling produces lengthy thinking content.
To preserve reasoning capability while reducing context length, an additional LLM call generates a compact summary ($\sim$200 words) of the reasoning process after the final answer is produced.
This summary captures background motivations, the core thinking process, and a concise function call history, and is prepended to the final answer for use as context in subsequent stages.
Closed-source models (\eg, GPT-4o, Claude) skip this step, as their reasoning is not exposed in the output.

\begin{table}[htbp]
  \centering
  \caption{Context composition for agent LLM calls at the main weekly stages. The roleplay prompt is shared across all stages and includes recent history for cross-stage continuity. Each stage appends its own stage prompt. Message history accumulates during multi-round interactions within a stage and is cleared between stages. Additional stages are listed in Table~\ref{tab:additional_stages}.}
  \label{tab:context_management}
  \small
  \begin{tabular}{p{2.0cm} p{3.0cm} p{8.0cm}}
    \toprule
    \textbf{Stage} & \textbf{Component} & \textbf{Information Provided} \\
    \midrule
    \multicolumn{3}{l}{\textbf{\textit{Roleplay Prompt}} {\scriptsize (system message, shared across all stages)}} \\
    \midrule
      & Persona          & Profile (background, personality, talents, position) + dynamic state (vitality, fulfillment, skills, assets) \\
      & Worldview        & Time system, weekly cycle, activity types, reward rules \\
      & Memory File Summary & Recent memory files ranked by access recency (limit 50; interaction partners guaranteed) \\
      & Recent History   & Weekly diaries (previous weeks) + activity records (recent weeks) + agent's own outputs from earlier stages this week \\
      & Location         & Current surroundings (activity venue or home) \\
      & Guidelines       & Commonsense rules, roleplay principles, output requirements \\
    \midrule
    \multicolumn{3}{l}{\textbf{\textit{Stage Prompt}} {\scriptsize (first user message, per-stage)}} \\
    \midrule
    \multirow{2}{*}{Plan}
      & Schedule         & Confirmed joint activities and public events for upcoming weeks \\
      & Plan Instructions & Goal-setting, reflection, and living standard selection \\
    \midrule
    \multirow{5}{*}{Contact}
      & Schedule         & Confirmed joint activities and public events for upcoming weeks \\
      & Error Feedback   & Errors from last contact round (invalid actions discarded) \\
      & Contact History  & Full message exchange history from the past 3 weeks \\
      & Map              & Available locations for activities \\
      & Contact Instructions & Scheduling window, action types (contact, propose, respond, cancel), action limit \\
    \midrule
    \multirow{4}{*}{\shortstack[l]{Activity\\{(Joint/}\\{Encounter)}}}
      & Schedule         & Confirmed activities for this week \\
      & Activity Background & Who, what, why for today's activity (for joint activities) \\
      & Participant Info & Public information of other participants \\
      & Activity Instructions & Visibility tags, gift/exit actions, turn format, word limit \\
    \midrule
    \multirow{2}{*}{\shortstack[l]{Activity\\(Solo)}}
      & Schedule         & Confirmed activities for this week \\
      & Solo Instructions & Activity choice, state effects, shopping/services \\
    \midrule
    \multirow{3}{*}{\shortstack[l]{Activity\\(Public)}}
      & Schedule         & Confirmed activities for this week \\
      & Event Info       & Event name, description, other participants present \\
      & Public Instructions & Observation-only interaction \\
    \midrule
    \multirow{1}{*}{Review}
      & Review Instructions & Weekly summary, reflection, and diary writing \\
    \midrule
    \multicolumn{3}{l}{\textbf{\textit{Message History}} {\scriptsize (other messages, within this stage)}} \\
    \midrule
      & Dialogue History {\scriptsize (joint/encounter)} & Multi-turn dialogue (environment model narration + participant responses, with visibility filtering) \\
      & Function Calls   & Tool call requests + results (up to 8 rounds per invocation) \\
      & Compacted Reasoning & For open-source models: $\sim$200-word summary of reasoning process prepended to final answer \\
    \bottomrule
  \end{tabular}
\end{table}

\begin{table}[htbp]
  \centering
  \caption{Stage prompt composition for additional agent LLM calls. All calls share the same roleplay prompt (Table~\ref{tab:context_management}) as the system message. Enter Activity and Exit Activity are described in \S~\ref{sec:appendix_activity}.}
  \label{tab:additional_stages}
  \small
  \begin{tabular}{p{2.8cm} p{3.0cm} p{7.2cm}}
    \toprule
    \textbf{Stage} & \textbf{Component} & \textbf{Information Provided} \\
    \midrule
    \multirow{2}{*}{\shortstack[l]{Signup\\{\scriptsize (before Contact)}}}
      & Schedule         & Confirmed activities for upcoming weeks \\
      & Signup Instructions & List of eligible public events (name, time, existing schedule, description), signup action format \\
    \midrule
    \multirow{3}{*}{\shortstack[l]{Enter Activity\\{\scriptsize (joint/encounter)}}}
      & Schedule         & Confirmed activities for this week \\
      & Activity Background & Who, what, why for today's activity; location description; other participants' public info \\
      & Analysis Instructions & Analyze position, goals, action plan, potential risks, and strategies before the activity starts \\
    \midrule
    \multirow{2}{*}{\shortstack[l]{Exit Activity\\{\scriptsize (joint/encounter)}}}
      & Activity Context  & Full multi-turn dialogue history from the activity, plus environment model feedback \\
      & Reflection Instructions & Summarize the activity process and reflect on mindset shifts and emotional arc \\
    \midrule
    \multirow{2}{*}{\shortstack[l]{Exit Activity\\{\scriptsize (solo)}}}
      & Activity Context  & Agent's action description and environment model feedback \\
      & Reflection Instructions & Reflect on the activity content and outcome \\
    \midrule
    \multirow{2}{*}{\shortstack[l]{Exit Activity\\{\scriptsize (public)}}}
      & Activity Context  & Agent's action description, environment model feedback, and other participants' activities \\
      & Reflection Instructions & Reflect on the activity; optionally create or update memory files for other participants \\
    \midrule
    \multirow{2}{*}{Position Application}
      & Position Info     & All available positions (description, income, skill growth, requirements) and agent's current position \\
      & Application Instructions & Express up to 3 position preferences in order; may choose to stay in current position \\
    \midrule
    \multirow{2}{*}{Judge Others}
      & Known People      & Names and memory file summaries of top related characters \\
      & Scoring Instructions & Score each known character independently on affection and respect (0 to 100) \\
    \midrule
    \multirow{2}{*}{\shortstack[l]{Settle Week\\{\scriptsize (conditional)}}}
      & Possessions       & Full list of currently owned items with descriptions \\
      & Discard Instructions & Select items to discard when possessions exceed the capacity limit \\
    \bottomrule
  \end{tabular}
\end{table}

\subsection{Response Filtering and Roleplay Principles}
\label{sec:appendix_principles}

During simulation, the environment model checks each agent response against a set of roleplay principles.
Table~\ref{tab:principles} lists the complete principle set.
Responses that violate any principle are marked as rejected and filtered out during training data construction (\S~\ref{sec:training_data}).

\begin{table}[htbp]
  \centering
  \caption{Roleplay principles used for response filtering. The environment model evaluates each agent response against these principles during simulation.}
  \label{tab:principles}
  \small
  \begin{tabular}{p{3.8cm} p{9.5cm}}
    \toprule
    \textbf{Principle} & \textbf{Description} \\
    \midrule
    Scope of control & The agent should only control its own actions and speech, not determine outcomes or control others' behaviors and thoughts. \\
    No hallucination & Only reference information present in context; do not fabricate objects, events, or information not mentioned. \\
    Character consistency & Personality traits and behavior patterns should match the established persona. \\
    Cognitive boundaries & Knowledge and cognitive limits should match the character's identity and background. \\
    Motivation consistency & Actions should be supported by reasonable internal motivations. \\
    State consistency & Mental and physical states (\eg, fatigue, injury) should not shift abruptly. \\
    Emotional continuity & Emotional changes should be gradual, not sudden. \\
    Natural relationship progression & Relationships should not develop abruptly; going from strangers to close friends requires a reasonable process. \\
    No parroting & Do not repeat the same content three or more times; conversation should make substantial progress. \\
    No AI assistant behavior & Characters should not speak or act like customer service agents or AI assistants. \\
    Substantive dialogue & Dialogue should provide new information rather than spinning in circles. \\
    Selective disclosure & What to share depends on the relationship; strangers do not bare their souls. \\
    Independent self & Characters should express their own goals, preferences, likes and dislikes. \\
    Colloquial speech & Use casual, conversational language in everyday dialogue. \\
    Directness & Speak directly; do not pad responses with unnecessary hedging or filler. \\
    First-person perspective & Speech, actions, and thoughts should use first-person perspective; action descriptions can omit the subject. \\
    \bottomrule
  \end{tabular}
\end{table}

\subsection{Fulfillment and Vitality}
\label{sec:appendix_states}

\paragraph{Fulfillment}
Fulfillment has four dimensions (mood, material, social, and esteem), each on a 0 to 100 scale.
The value 50 serves as a neutral baseline, representing neither satisfaction nor dissatisfaction.
Values above 50 indicate positive fulfillment, while values below 50 indicate dissatisfaction.
At the beginning of each week, each dimension decays proportionally:
\begin{equation}
v' = v - v \times r_d
\end{equation}
where $v$ is the current value and $r_d$ is the per-dimension decay ratio.
The decay ratios are: mood $r_d = 0.15$, social $r_d = 0.15$, esteem $r_d = 0.15$, and material $r_d = 0$.
Material fulfillment does not decay because it is controlled separately by the economy system through living standard selection (\S~\ref{app:economy}), and does not require an additional decay mechanism.
This decay mechanism simulates the natural fading of satisfaction: agents must continuously engage in activities to maintain fulfillment, or it naturally declines toward the baseline.
This creates an intrinsic motivation loop: inactivity leads to declining fulfillment, which drives agents to seek new activities.

\paragraph{Vitality}
Vitality represents an agent's energy level on a 0 to 100 scale.
Activities consume vitality, while rest recovers it.
The vitality level is communicated to agents via prompt descriptions (\eg, 50 to 69 indicates tiredness and reduced efficiency), influencing their behavioral decisions.
Additionally, low vitality incurs penalties in reward calculation: when vitality falls below a threshold, the subjective reward is reduced.

\subsection{Profile Update}
\label{sec:appendix_profile_update}

At the end of each simulated year, the environment model updates each agent's profile to reflect a year of lived experience.
The update covers personality traits, talents, skills, and other numeric attributes.
The environment model receives the agent's current profile, the year's activity history, social interaction records, and state change trajectories.
Based on this information, the environment model generates an updated profile that reflects the agent's growth, changes, or decline over the year.
For example, an agent who frequently engages in social activities may see an increase in extraversion, while an agent who has worked in a specific field for a long time may gain skill growth in related areas.

\subsection{Details of Economy System}
\label{app:economy}

Each agent has a deposit account with income and expenses.

\paragraph{Income}
Agents receive income from three sources:
(1) a \textit{position-based weekly salary} (\$0 to \$500), automatically deposited at the beginning of each week;
(2) \textit{extra weekly income} from family support, investments, or other external sources, also deposited automatically;
and (3) \textit{work income} earned during solo activity slots (\$40 to \$200 per day, tiered by skill level).
The first two constitute an agent's fixed weekly income (combined range approximately \$100 to \$500),
while the third is available to any agent who chooses to spend its activity slots on work.

\paragraph{Expenses}
Agents spend money in two ways.
First, during the weekly planning phase, each agent selects a \textit{living standard} from four tiers:
frugal (\$100/week), moderate (\$200/week), comfortable (\$300/week), or luxurious (\$500/week).
Higher tiers yield greater material fulfillment, while the frugal tier reduces it.
This mechanism forces a core trade-off between saving and immediate material satisfaction.
Second, agents may encounter \textit{consumption opportunities} during solo activities (\eg, shopping, entertainment),
where the environment model provides available items and prices based on the specific context, and the agent decides whether to make a purchase.
Consumption brings material fulfillment following a piecewise function:
\$0 to \$100 yields +1 per \$20 spent (max +5),
\$100 to \$300 yields an additional +1 per \$40 (max +10 total),
and spending above \$300 is capped at +10.

We create a \textit{price reference list} for each world---ranging from free activities (\$0) to luxury purchases (\$15{,}000)---which the environment model consults when assigning goods and prices to activities.
This ensures that large purchases require sustained saving over multiple simulated weeks, providing a natural anti-inflation mechanism.

\subsection{Details of Position and Position Application}
\label{sec:appendix_enrollment}

A position represents an agent's job or role in the world, serving as the key source of income and skill development.

\paragraph{Position Attributes}
Each position is defined by the following attributes:
\begin{itemize}[nosep]
\item \textbf{Organization and role}: together forming a unique identifier in the format \texttt{\{organization\}/\{role\}}, \eg, \texttt{The Academy/Potions Teacher}.
\item \textbf{Type}: \texttt{work} or \texttt{non-work} (\eg, student). Both types can provide income.
\item \textbf{Weekly income}: a fixed amount the agent receives each week automatically.
\item \textbf{Weekly skill gain}: experience points in specific skills that accumulate automatically each week, \eg, \texttt{\{``teaching'': 5, ``leadership'': 3\}}.
\item \textbf{Capacity}: the maximum number of agents that can hold this position simultaneously.
\item \textbf{Requirements} (optional): entry conditions including minimum skill levels and age limits (\texttt{min\_age}, \texttt{max\_age}). For example, a high-school student position may set \texttt{max\_age=18}.
\end{itemize}

\paragraph{Position Creation}
Positions are introduced through three sources:
\begin{enumerate}[nosep]
\item \textbf{Born with characters.} Each character comes with an initial position as part of its background, assigned during the character creation process (\S~\ref{app:creation_process}).
\item \textbf{World initialization.} After character creation and before the simulation starts, the environment model is provided with all positions initially held by characters and designs additional positions to enrich the world's occupational structure.
To ensure balance, the system imposes constraints on total capacity and diversity: total capacity across all positions ranges from $C$ to $1.5C$ (where $C$ is the number of agents), each work position holds at most $\lfloor C/3 \rfloor$ agents, and the number of distinct positions ranges from 10 to $\lfloor C/3 \rfloor$.
\item \textbf{Yearly growth.} At the beginning of each year, the system adds $\max(2, \lfloor P/10 \rfloor)$ new positions, where $P$ is the initial position count.
New positions require minimum skill levels above the current highest among all agents, ensuring they serve as growth targets that agents must work toward.
Income for new positions scales with skill requirements, and capacity is limited to 1 to 2 slots to maintain scarcity and encourage competition.
\end{enumerate}

\paragraph{Position Application Process}
Position application is triggered at the end of each simulated year, allowing agents to change their positions.
The process consists of three steps:
\begin{inparaenum}[\it (1)]
\item each agent expresses up to three position preferences, ranked by priority.
Agents may choose \texttt{STAY\_CURRENT} to retain their current position, which is automatically granted unless they have exceeded the position's age limit;
\item the environment model evaluates candidates across three matching rounds: in the $k$-th round, unmatched agents' $k$-th choice is processed, with the environment model assessing candidates' skills against remaining slots;
\item accepted agents update their positions and establish relationships with new colleagues. Unmatched agents remain in their current position, while those who have exceeded their position's age limit become unemployed.
\end{inparaenum}

\subsection{Location System}
\label{sec:appendix_location}

We design a location system for \method, mainly for providing agents with accurate, grounded environmental perception.
We observe that without environmental information, agents tend to hallucinate objects or places that do not exist.
We therefore implement a simple location system to provide such information.
The location system is not the focus of this work.
The system comprises two types of locations---public locations and private locations---both
created at initialization and persisted in \texttt{locations.json}.

\paragraph{Public Locations}
Public locations are generated by the environment model based on the world setting and character profiles.
Each location contains a name, size (\texttt{small}, \texttt{medium}, or \texttt{large}),
a description (4 to 6 sentences covering spatial layout and atmosphere),
and a list of surrounding objects (5 to 6 items).
The object list serves as a grounding constraint:
agents are expected to reference only objects that exist in the environment,
rather than fabricating props that are not present.
Each world generates 30 public locations by default,
with a size distribution of approximately 50\% small, 35\% medium, and 15\% large.

\paragraph{Private Locations}
Private locations represent each character's home (key: \texttt{home/\{name\}}).
Each home is generated by the environment model based on the character's backstory and occupation,
containing a description (2 to 4 sentences) and a list of representative objects (4 to 6 items).
An agent can invite others to its home for activities;
others' homes are only accessible when invited.

\paragraph{Location Selection}
When an agent proposes a joint activity, or when the environment model arranges an encounter activity,
they are provided with the full list of available locations and must choose one from the list.
For agents proposing activities, their own private location is also available as an option.

\subsection{Numeric System}
\label{app:numeric}

\paragraph{Personality Traits}
All characters share the same 10 personality dimensions, each measured on a 0 to 100 scale where 50 represents the neutral baseline:
confidence, control, curiosity, empathy, extraversion, feeling, intuition, judging, patience, and responsibility.
These dimensions cover the four MBTI axes (extraversion vs.\ introversion, sensing vs.\ intuition, thinking vs.\ feeling, judging vs.\ perceiving) along with supplementary social and motivational traits.
These traits evolve slowly during the simulation, changing by at most $\pm 3$ per simulated year, reflecting the psychological stability of personality.

\paragraph{Talents}
All characters share the same 9 talent dimensions, measured on a 0 to 100 scale with 50 representing the population average:
beauty, communication, creativity, health, honesty, integrity, intelligence, leadership, and trustworthiness.
Talents represent innate aptitudes and change by at most $\pm 5$ per year.

\paragraph{Skills}
Skills are measured on an open-ended 0 to 300 scale:
0 (untrained), 10 (beginner), 30 (some experience), 100 (proficient), and 300 (master).
Unlike personality and talents, skills grow continuously through practice and work, increasing by $+1$ to $+5$ per simulated week depending on the activity, with no hard upper bound.

\paragraph{Agent States}
Two dynamic state variables track each agent's condition.
\textit{Vitality} (0 to 100, initial value 70) represents physical energy, depleted by activities ($-1$ to $-5$ per slot) and restored by rest ($+5$ to $+10$).
\textit{Fulfillment} comprises four independent dimensions---material, mood, social, and esteem---each on a 0 to 100 scale with an initial neutral value of 50.
Fulfillment naturally decays each week, requiring agents to actively maintain their well-being through social interactions, leisure, and achievements.

\subsection{Concurrency Control}
\label{sec:appendix_concurrency}

The system is designed to maximize concurrency across all activity executions within a day.
Different activity types have different concurrency characteristics and are handled accordingly.

Joint activities involve sequential multi-turn dialogue and are the hardest to parallelize; they are therefore given the highest priority and submitted first.
Encounter activities follow the same logic.
Solo activities are fully independent and execute with complete concurrency.
Public activities employ a two-level concurrency design.
At the outer level, multiple public activities run concurrently.
At the inner level, within each public activity, all participants first complete their individual phases (entering the activity and acting independently) in parallel, then proceed through the subsequent phases also in parallel.

Overall, activities are submitted in the order: joint $\to$ solo $\to$ public.
While joint activities are running, solo activities are interleaved to fill available capacity; public activities are submitted last to avoid competing with joint activities for resources.

\subsection{Configuration Parameters}
\label{app:config}

Table~\ref{tab:config_params} lists the key numerical parameters that govern simulation behavior.

\begin{table}[htbp]
\centering
\caption{Key configuration parameters used in the simulation.}
\label{tab:config_params}
\resizebox{0.8\linewidth}{!}{
\small
\begin{tabular}{p{4.2cm} p{1.8cm} p{7.2cm}}
\toprule
\textbf{Parameter} & \textbf{Value} & \textbf{Description} \\
\midrule
\multicolumn{3}{l}{\textit{Time}} \\
\texttt{n\_year} & 10 & Simulated years per run \\
\texttt{n\_week} & 10 & Weeks per simulated year \\
\texttt{n\_day} & 5 & Activity days per week \\
\texttt{n\_contact\_slot} & 5 & Communication rounds per Contact stage \\
\midrule
\multicolumn{3}{l}{\textit{Contact}} \\
\texttt{n\_action\_per\_slot} & 10 & Max actions per agent per contact round \\
\texttt{max\_future\_schedule\_weeks} & 4 & How many weeks ahead an agent can propose activities \\
\texttt{n\_prev\_contact\_weeks} & 3 & Weeks of prior contact history shown in prompt \\
\midrule
\multicolumn{3}{l}{\textit{Activity}} \\
\texttt{joint\_activity\_min/max\_turns} & 5 / 20 & Turn range for joint and encounter activities \\
\texttt{solo\_activity\_min/max\_turns} & 2 / 5 & Turn range for solo activity environment model evaluation \\
\texttt{max\_possessions} & 50 & Max items an agent can hold before discarding \\
\midrule
\multicolumn{3}{l}{\textit{Public Activity}} \\
\texttt{max\_events\_per\_week} & 10 & Max public events generated per week \\
\texttt{max\_repeat\_weeks} & 5 & Max duration (weeks) a public event can repeat \\
\midrule
\multicolumn{3}{l}{\textit{Location}} \\
\texttt{n\_locations} & 30 & Number of public locations per world \\
\midrule
\multicolumn{3}{l}{\textit{Fulfillment Decay}} \\
\texttt{decay\_ratio} (mood) & 0.15 & Weekly proportional decay for mood fulfillment \\
\texttt{decay\_ratio} (social) & 0.15 & Weekly proportional decay for social fulfillment \\
\texttt{decay\_ratio} (esteem) & 0.15 & Weekly proportional decay for esteem fulfillment \\
\texttt{decay\_ratio} (material) & 0 & Material fulfillment does not decay \\
\midrule
\multicolumn{3}{l}{\textit{State Change Limits per Activity}} \\
\texttt{vitality} delta & [$-$5, $+$5] & Per-activity vitality change (all activity types) \\
\texttt{mood} delta (solo/joint) & [$-$5, $+$5] & Per-activity mood change \\
\texttt{social} delta (joint) & [$-$5, $+$5] & Per-activity social fulfillment change \\
\texttt{esteem} delta (solo) & [$-$1, $+$1] & Per-activity esteem change (solo) \\
\texttt{esteem} delta (joint) & [$-$3, $+$3] & Per-activity esteem change (joint) \\
\texttt{skills} delta & [0, $+$3] & Per-activity skill increase \\
\texttt{money} delta (solo) & [$-$200, $+$200] & Per-activity spending/earning (solo only) \\
\midrule
\multicolumn{3}{l}{\textit{Economy}} \\
\texttt{weekly\_income} & \$100 to \$500 & Position-based weekly salary range \\
\texttt{work\_income} (low/mid/high) & \$40 to \$200 & Daily work income by skill tier \\
Living standard costs & \$100 / 200 / 300 / 500 & Frugal / moderate / comfortable / luxurious \\
Living standard material delta & $-$5 / 0 / $+$5 / $+$10 & Material fulfillment change per living standard \\
\midrule
\multicolumn{3}{l}{\textit{Reward}} \\
\texttt{period\_weeks} & 10 & Reward calculation interval (once per year) \\
\texttt{pagerank\_damping} & 0.85 & PageRank damping factor for social reward \\
\texttt{social\_weight} & 0.4 & $\lambda_{\text{social}}$: weight of social reward in total life reward \\
\texttt{economy\_weight} & 0.2 & $\lambda_{\text{econ}}$: weight of economy reward in total life reward \\
\texttt{subjective\_weight} & 0.4 & $\lambda_{\text{subj}} = 1 - \lambda_{\text{social}} - \lambda_{\text{econ}}$ \\
\texttt{mutual\_affection\_alpha} & 2.0 & Mutual affection coefficient $\alpha$ in social reward \\
\texttt{misery\_threshold\_percentile} & 0.25 & Bottom 25\% fulfillment triggers misery penalty \\
\texttt{misery\_penalty\_value} & 5 & Penalty weight $\lambda_p$ deducted per misery event \\
\texttt{gamma} & 0.90 & Discount factor for computing returns \\
\texttt{sft\_top\_fraction} & 0.25 & Top fraction of agents selected for SFT training \\
\bottomrule
\end{tabular}}
\end{table}

\section{Experiment Settings}

\subsection{Training via Life Rewards}
\label{sec:appendix_sft}

Qwen3.5-397B-A17B is fine-tuned via supervised fine-tuning (SFT).
Training data is constructed by selecting the top 25\% of agents by advantage at each time step
from the first four simulated years across all three worlds.
Environment model data is collected separately via random sampling, with high-frequency activity types downsampled to balance the training distribution.

\paragraph{Return Normalization}
In advantage estimation, we use the return at the previous time step as the baseline:
$A_t = G_{t+1} - G_t$.
However, $G_t$ at different time steps accumulates different numbers of future rewards, leading to different scales.
We normalize each return by dividing by the discounted effective horizon:
\begin{equation}
G_t^{\text{norm}} = \frac{G_t}{\sum_{k=0}^{T-t} \gamma^k} = \frac{G_t \cdot (1 - \gamma)}{1 - \gamma^{T-t+1}}
\end{equation}
If all rewards were a constant $r$, the normalized return would equal $r$ at every time step,
regardless of the remaining horizon.
This removes the scale difference across different time steps, making the advantage computation more reasonable.

\paragraph{Per-Period Trajectory Selection}
Even with return normalization, scale differences in advantages across time steps may persist.
In addition, the initial step lacks a meaningful estimate of reward and return.
We set a virtual $G_0 = 0$ in this case, which may inflates the advantages of the first period.
To address these issues, we select the top 25\% of agent trajectories by advantage \textit{within each period independently}, rather than ranking globally across different periods.
This follows a natural heuristic that high-quality trajectories should be distributed across different periods.
In practice, it eliminates the effect of cross-period scale differences on selection and ensures temporal diversity in the training data.

Following~\citet{lu2025onpolicy}, self-distillation is employed to prevent catastrophic forgetting.
Specifically, responses are generated from Qwen3.5-397B on the Tulu V3 instruction set~\citep{lambert2024tulu}
and mixed into the training data as general-purpose samples.
The training mixture consists of 50\% role-playing data and 50\% general-purpose data (measured by output tokens).

A learning rate of $1 \times 10^{-5}$ with a minimum learning rate of $1 \times 10^{-6}$ is used,
with a training batch size of 256.
The model is fine-tuned for 1 epoch on 30 nodes of 8$\times$H100 80GB GPUs.

\subsection{CoSER Evaluation}
\label{sec:appendix_coser}

Qwen3.5-397B-Agentopia is evaluated on the CoSER Test~\citep{wang2025coser}
to assess its generalization to general-purpose role-playing tasks.
CoSER evaluates LLMs via given-circumstance acting:
given a character profile and situational context,
the model generates character-consistent dialogue and actions.

The evaluation uses 200 test samples from the CoSER Test.
Qwen3-235B-A22B~\citep{qwen3} serves as the judge model,
the next speaker prediction (NSP) model, and the environment model.
Each simulation runs for at most 20 turns.

Performance is assessed across four dimensions:
(1) \textit{Anthropomorphism}, evaluating whether agents behave in a human-like manner,
covering self-identity, emotional depth, persona coherence, and social interaction;
(2) \textit{Character Fidelity}, assessing whether agents faithfully portray their characters,
examining language style, knowledge and background, personality and behavior, and social relationships;
(3) \textit{Storyline Quality}, evaluating whether the simulated conversation develops naturally,
focusing on narrative flow and logical consistency;
and (4) \textit{Storyline Consistency}, measuring alignment between the simulated conversation
and the original dialogue,
\ie, whether agents' reactions (emotions, attitudes, behaviors) remain consistent with the original.

The comparison models include
Claude-4.5-Opus, Gemini-3-Pro, Claude-4.5-Sonnet, GPT-5-Mini,
CoSER-70B, and the baseline Qwen3.5-397B.

\subsection{Metrics Definition}
\label{sec:appendix_metrics}

We define a set of analytical metrics to describe agent behaviors and social dynamics,
covering fulfillment, activity, contact, personal growth, social evaluation, and computational cost.
Reward metrics are defined separately in \S~\ref{sec:life_reward}.
Other metrics are defined as follows: 

\paragraph{Fulfillment Metrics}
Four fulfillment dimensions (0 to 100 scale) are recorded as snapshots at the end of each year,
reflecting an agent's subjective well-being as described in \S~\ref{app:numeric}:
(1) \textbf{fulfillment\_mood},
(2) \textbf{fulfillment\_material},
(3) \textbf{fulfillment\_social}, and
(4) \textbf{fulfillment\_esteem}.
Additionally,
(5) \textbf{fulfillment\_penalties} counts the number of (week, dimension) pairs
where fulfillment or vitality falls below a penalty threshold
(summed yearly).

\paragraph{Activity Metrics}
Six metrics describe an agent's activity participation patterns
(all summed yearly):
(1) \textbf{joint\_proposed}: number of joint activities initiated by the agent;
(2) \textbf{joint\_participated}: number of joint activities attended, including those the agent proposed;
(3) \textbf{public\_participated}: number of public activities attended;
(4) \textbf{solo\_count}: number of solo activity days, computed as total activity days per week minus days occupied by any scheduled activity;
(5) \textbf{activity\_consumption\_count}: number of consumption events during the week;
(6) \textbf{total\_spending\_amount}: total spending on new possessions.

\paragraph{Contact Metrics}
Two metrics capture an agent's social initiative and popularity
(both summed yearly):
(1) \textbf{active\_contacts}: number of contact messages sent by the agent;
(2) \textbf{passive\_contacts}: number of contact messages received by the agent.

\paragraph{Personal Growth Metrics}
Four metrics measure an agent's skill development and economic status.
The first two are summed yearly; the last two use year-end snapshots:
(1) \textbf{extra\_earning\_count}: number of solo activity slots in which the agent earned work income;
(2) \textbf{skill\_improvement\_count}: number of activities that produced at least one skill advance;
(3) \textbf{total\_skills}: sum of all skill values at week end;
(4) \textbf{deposit}: deposit balance at week end.

\paragraph{Social Evaluation Metrics}
Six metrics are recorded at settlement weeks, reflecting an agent's standing in the social network
(year-end snapshots):
(1) \textbf{n\_liked\_by}: number of other agents whose affection toward this agent is $\geq 60$;
(2) \textbf{n\_likes}: number of others this agent has affection $\geq 60$ toward;
(3) \textbf{n\_respected\_by}: number of other agents whose respect toward this agent is $\geq 60$;
(4) \textbf{n\_respects}: number of others this agent has respect $\geq 60$ toward;
(5) \textbf{n\_mutual\_like}: number of agents where both this agent's affection toward them and their affection toward this agent are $\geq 60$;
(6) \textbf{n\_mutual\_respect}: number of agents where both this agent's respect toward them and their respect toward this agent are $\geq 60$.

\paragraph{Computational Cost Metrics}
Three metrics measure the per-week LLM usage
(all summed yearly):
(1) \textbf{input\_tokens}: total input tokens consumed by LLM calls;
(2) \textbf{output\_tokens}: total output tokens generated by LLM calls;
(3) \textbf{function\_call\_count}: total number of function calls made by LLM.

\section{Additional Results}

\subsection{Training via Life Rewards: Detailed Trends}
\label{sec:appendix_sft_trends}

\begin{figure}[htbp]
\centering
\includegraphics[width=\textwidth]{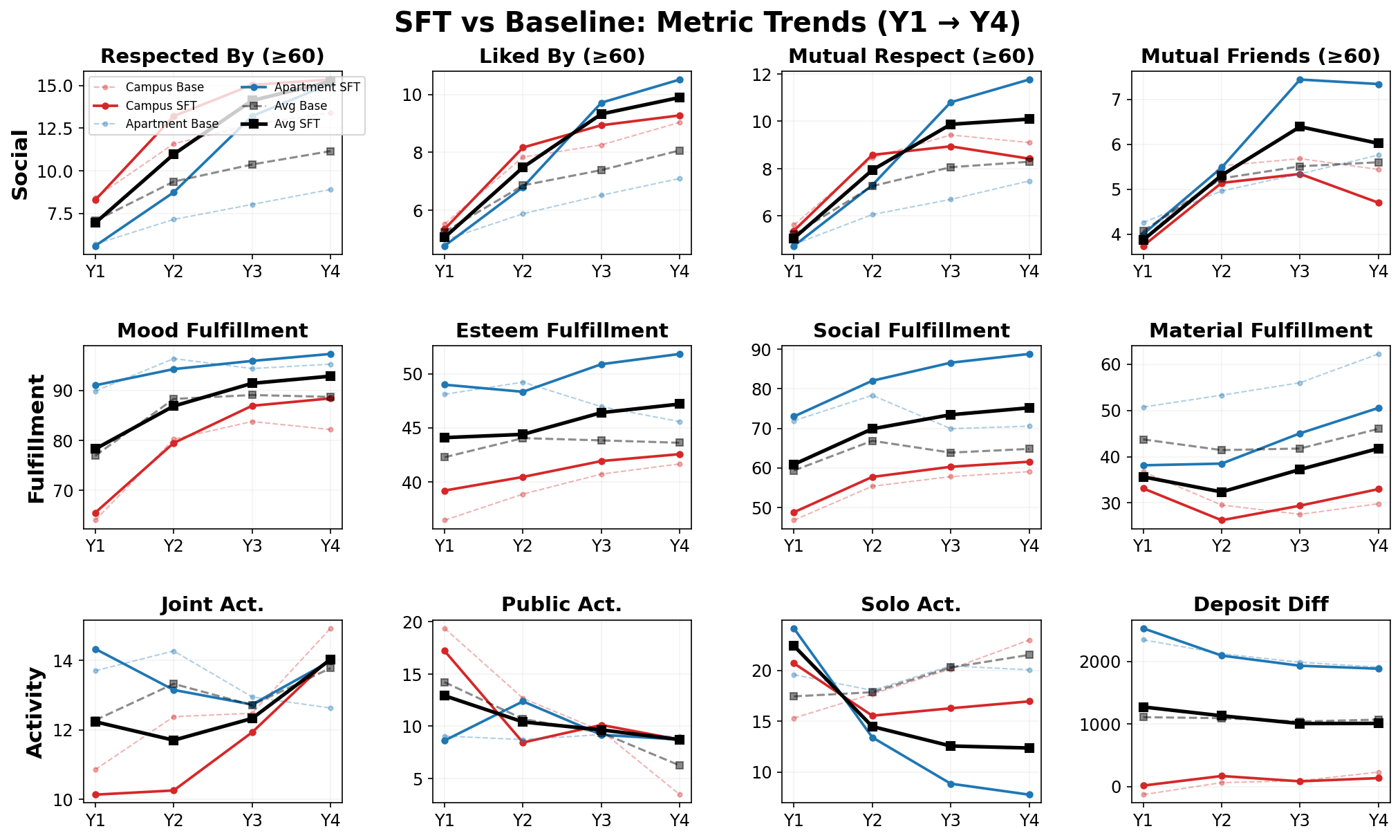}
\caption{%
Metric trends for social evaluation, economy, and activity patterns
over four simulated years across \campus{} and \apartment{}.
Solid lines represent Qwen3.5-397B-Agentopia; dashed lines represent the original Qwen3.5-397B;
black lines show the cross-world average.}
\label{fig:sft_main}
\end{figure}

Table~\ref{tab:sft_cross_world} reports cross-world average metrics.
Here we present the per-year trends for both \campus{} and \apartment{},
illustrating how Qwen3.5-397B and Qwen3.5-397B-Agentopia diverge over the four simulated years as the underlying model.

\paragraph{Qwen3.5-397B-Agentopia's advantage emerges over the long term}
The two models start with similar social metrics in Year~1, but diverge steadily thereafter:
by Year~4, agents driven by Qwen3.5-397B-Agentopia are respected by 35.7\% more peers,
liked by 22.2\% more,
and hold 21.7\% more mutual-respect ties than those driven by the original Qwen3.5-397B.
Rather than producing immediate gains,
the advantage of life reward training manifests
as agents interact with the society over multiple years.

\paragraph{Economy--fulfillment trade-off}
In Year~1, Qwen3.5-397B-Agentopia accumulates more wealth than the baseline.
In later years, economy reward converges between the two models,
probably because Qwen3.5-397B-Agentopia agents treat their savings as sufficient
and shift priorities away from further accumulation.
Meanwhile, material fulfillment remains consistently lower ($-$9\% to $-$22\%),
indicating that agents learn to reduce spending to optimize economy reward.

\begin{figure}[htbp]
\centering
\includegraphics[width=0.8\textwidth]{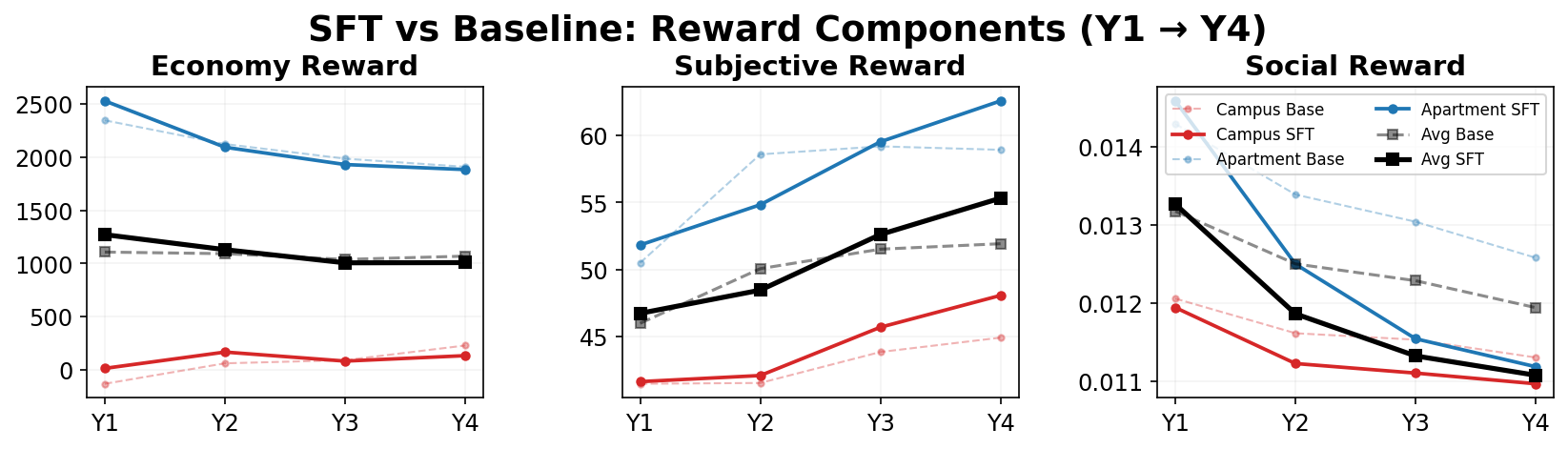}
\caption{%
Reward component trends over four simulated years,
averaged over \campus{} and \apartment{}.
Solid lines represent Qwen3.5-397B-Agentopia; dashed lines represent the original Qwen3.5-397B.}
\label{fig:sft_reward}
\end{figure}

\paragraph{Subjective reward: material fulfillment masks underlying gains}
As shown in Figure~\ref{fig:sft_reward},
Qwen3.5-397B-Agentopia's subjective reward is lower than the baseline in the first two years,
but surpasses it in Year~3 and Year~4.
This is largely driven by the material fulfillment gap:
agents save more aggressively, sacrificing material satisfaction.
However, Figure~\ref{fig:sft_main} shows that Qwen3.5-397B-Agentopia
leads consistently in mood, social, and esteem fulfillment throughout all four years.
If material fulfillment were excluded,
Qwen3.5-397B-Agentopia would likely hold a sustained advantage in subjective reward
from the very beginning.

\subsection{Per-World Reward Distributions}
\label{sec:appendix_reward_per_world}

\begin{figure*}[htbp]
  \centering
  \includegraphics[width=0.8\textwidth]{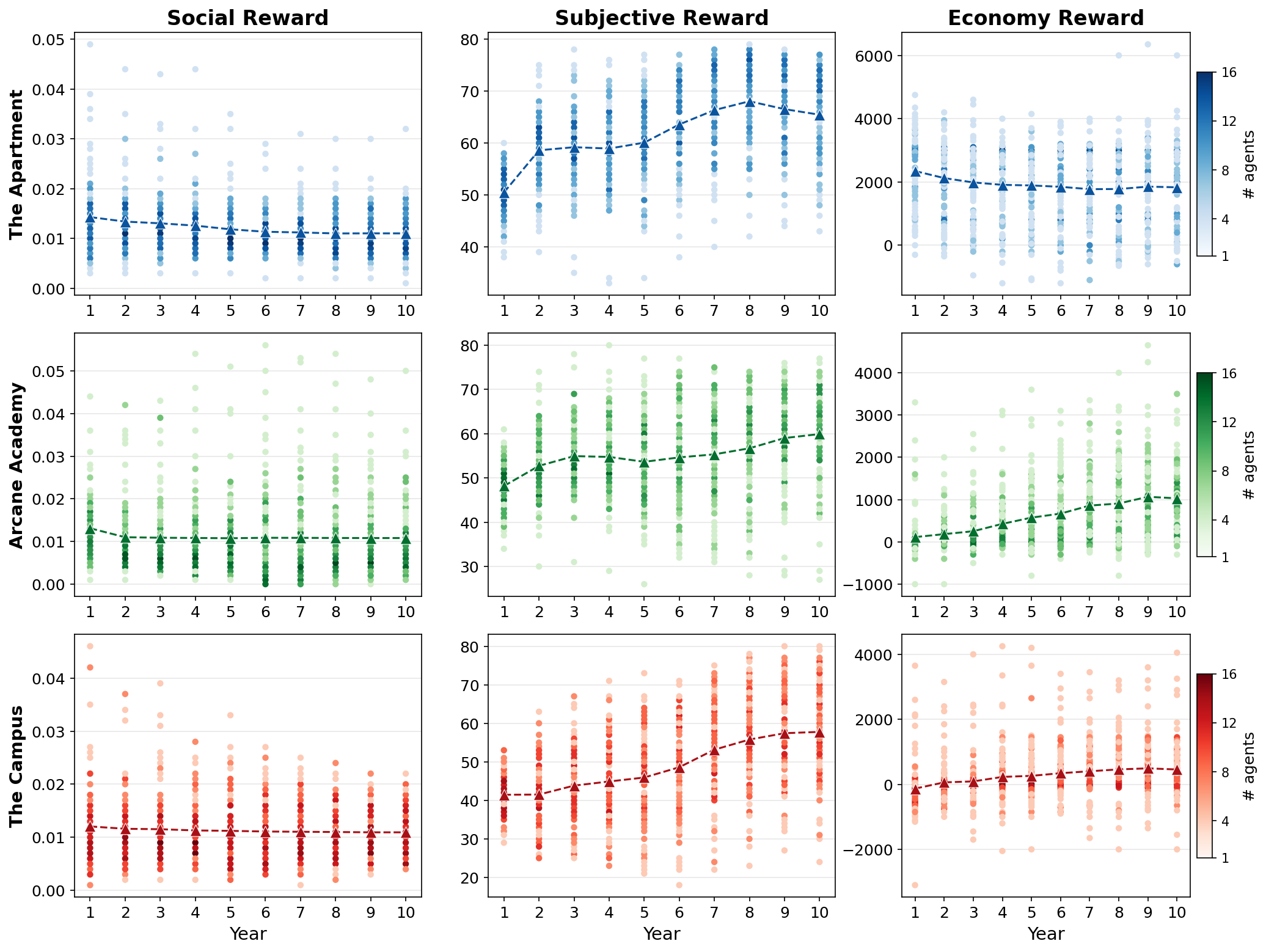}
  \caption{%
    Per-world reward dimension distributions over 10 simulated years.
    Each row corresponds to one world; triangles mark annual means.
  }
  \label{fig:reward_dist_per_world}
\end{figure*}

Figure~\ref{fig:reward_distribution} presents pooled reward distributions across all three worlds.
Figure~\ref{fig:reward_dist_per_world} provides per-world breakdowns,
revealing structural differences across social settings.

Based on Figure~\ref{fig:reward_dist_per_world}, we analyze each reward dimension and have the following findings:
(1) Subjective reward shows a consistent upward trend across all three worlds,
with the mean rising from approximately 45 to 60 over 10 years,
indicating that agents progressively achieve greater well-being through sustained social participation;
(2) social reward remains broadly stable across all three worlds, with the mean near 0.01.
This stability is inherent to its design:
social reward is based on relative rankings rather than absolute performance,
so gains for one agent largely come at the expense of others;
(3) economy reward exhibits high variance across all three worlds.
The absolute levels differ due to world design (different initial price and salary structures).
\academy{} and \campus{} show upward trends,
while \apartment{} shows declining yearly deposit gains
as agents' deposits grow large relative to their income.
\academy{} shows the clearest growth trajectory (mean rising from approximately \$100 to \$1{,}000),
suggesting that agents learn to leverage skill advances and extra income for wealth accumulation.
The bimodal appearance of economy reward in the pooled figure (Figure~\ref{fig:reward_distribution})
stems from this cross-world heterogeneity rather than within-world polarization.

\subsection{Reward--Behavior Correlation Details}
\label{sec:appendix_reward_correlation}

\paragraph{Social Reward}
Table~\ref{tab:corr_social} lists the 10 behavioral metrics most strongly correlated with social reward.
\texttt{n\_respected\_by} and \texttt{n\_liked\_by} are the only strong predictors
($r = 0.68$, with $r = 0.74$ to $0.86$ in \academy{} and \campus{}),
while the third-ranked metric drops sharply to $r = 0.19$.
All behavioral metrics (contacts, activities, tokens) remain below $r = 0.19$,
confirming that social standing is determined by others' evaluations rather than the agent's own effort.
This result is by design: social reward is computed via PageRank-based algorithm over inter-agent evaluation rankings,
which naturally amplifies reputation signals.

\begin{table}[htbp]
  \centering
  \caption{%
    Top-10 metrics correlated with social reward, sorted by mean Pearson's $|r|$.
  }
  \label{tab:corr_social}
  \small
  \begin{tabular}{clcccc}
    \toprule
    \# & Metric & Pearson's $r$ & \apartment{} & \academy{} & \campus{} \\
    \midrule
    1  & n\_respected\_by         & $+$0.68 & $+$0.39 & $+$0.86 & $+$0.79 \\
    2  & n\_liked\_by             & $+$0.68 & $+$0.44 & $+$0.85 & $+$0.74 \\
    3  & n\_likes                 & $+$0.19 & $+$0.11 & $+$0.20 & $+$0.27 \\
    4  & joint\_participated      & $+$0.19 & $+$0.34 & $+$0.13 & $+$0.10 \\
    5  & fulfillment\_esteem      & $+$0.17 & $+$0.28 & $+$0.12 & $+$0.10 \\
    6  & total\_skills            & $+$0.16 & $+$0.12 & $+$0.14 & $+$0.21 \\
    7  & joint\_proposed          & $+$0.15 & $+$0.25 & $+$0.15 & $+$0.05 \\
    8  & output\_tokens           & $+$0.15 & $+$0.28 & $+$0.02 & $+$0.15 \\
    9  & passive\_contacts        & $+$0.15 & $+$0.09 & $+$0.12 & $+$0.23 \\
    10 & n\_respects              & $+$0.14 & $+$0.10 & $+$0.06 & $+$0.27 \\
    \bottomrule
  \end{tabular}
\end{table}

\paragraph{Subjective Reward}
Subjective reward is driven by three factor groups as shown in Table~\ref{tab:corr_subjective}:
(1) fulfillment dimensions, with material ($r = 0.73$) as the strongest single predictor,
highly consistent across worlds ($0.71$ to $0.77$),
followed by mood ($0.54$), social ($0.52$), and esteem ($0.30$);
(2) penalties ($r = -0.64$), triggered when fulfillment remains persistently low,
consistent across all three worlds ($-0.61$ to $-0.68$);
and (3) social activity, where passive contacts ($0.43$), active contacts ($0.42$),
and likes given ($0.39$) serve as positive predictors,
while solo activities are a notable negative signal
(\apartment{} $r = -0.53$, \campus{} $r = -0.49$).

\begin{table}[htbp]
  \centering
  \caption{%
    Top-10 metrics correlated with subjective reward, sorted by mean Pearson's $|r|$.
  }
  \label{tab:corr_subjective}
  \small
  \begin{tabular}{clcccc}
    \toprule
    \# & Metric & Pearson's $r$ & \apartment{} & \academy{} & \campus{} \\
    \midrule
    1  & fulfillment\_material    & $+$0.73 & $+$0.77 & $+$0.71 & $+$0.72 \\
    2  & n\_penalties             & $-$0.64 & $-$0.61 & $-$0.68 & $-$0.64 \\
    3  & fulfillment\_mood        & $+$0.54 & $+$0.56 & $+$0.44 & $+$0.62 \\
    4  & input\_tokens            & $+$0.52 & $+$0.58 & $+$0.37 & $+$0.62 \\
    5  & fulfillment\_social      & $+$0.52 & $+$0.54 & $+$0.37 & $+$0.65 \\
    6  & total\_spending\_amount  & $+$0.46 & $+$0.45 & $+$0.42 & $+$0.50 \\
    7  & passive\_contacts        & $+$0.43 & $+$0.44 & $+$0.32 & $+$0.51 \\
    8  & active\_contacts         & $+$0.42 & $+$0.42 & $+$0.30 & $+$0.53 \\
    9  & solo\_count              & $-$0.40 & $-$0.53 & $-$0.17 & $-$0.49 \\
    10 & n\_likes                 & $+$0.39 & $+$0.46 & $+$0.34 & $+$0.38 \\
    \bottomrule
  \end{tabular}
\end{table}

\paragraph{Economy Reward}
Economy reward is dominated by deposit level (Table~\ref{tab:corr_economy}, $r = 0.56$),
followed by extra earnings ($0.31$) and total skills ($0.30$).
These two metrics are linked in two ways:
work activities simultaneously produce income and skill growth,
so extra earnings and skill advances naturally co-occur;
moreover, higher skill levels increase the earnings from subsequent work,
creating a positive feedback loop that drives deposit growth.
Metrics ranked 4th and below all have $|r| \leq 0.15$.

\begin{table}[htbp]
  \centering
  \caption{%
    Top-10 metrics correlated with economy reward, sorted by mean Pearson's $|r|$.
  }
  \label{tab:corr_economy}
  \small
  \begin{tabular}{clcccc}
    \toprule
    \# & Metric & Pearson's $r$ & \apartment{} & \academy{} & \campus{} \\
    \midrule
    1  & deposit                  & $+$0.56 & $+$0.62 & $+$0.77 & $+$0.29 \\
    2  & extra\_earning\_count    & $+$0.31 & $+$0.20 & $+$0.43 & $+$0.30 \\
    3  & total\_skills            & $+$0.30 & $+$0.15 & $+$0.41 & $+$0.34 \\
    4  & fulfillment\_material    & $+$0.13 & $-$0.08 & $+$0.31 & $+$0.15 \\
    5  & n\_likes                 & $+$0.09 & $-$0.12 & $+$0.24 & $+$0.14 \\
    6  & solo\_count              & $+$0.15 & $+$0.19 & $+$0.28 & $-$0.02 \\
    7  & n\_respects              & $+$0.06 & $-$0.15 & $+$0.22 & $+$0.11 \\
    8  & skill\_improvement\_count& $+$0.14 & $+$0.17 & $+$0.26 & $+$0.01 \\
    9  & public\_participated     & $-$0.14 & $-$0.11 & $-$0.15 & $-$0.14 \\
    10 & fulfillment\_social      & $-$0.14 & $-$0.14 & $-$0.26 & $-$0.00 \\
    \bottomrule
  \end{tabular}
\end{table}

\paragraph{Cross-World Consistency}
The top predictors of social and subjective rewards are consistent
in direction and magnitude across all three worlds.
For social reward, reputation metrics (\texttt{n\_respected\_by}, \texttt{n\_liked\_by})
are dominant in all worlds, though the effect is stronger in \academy{} and \campus{}
($r = 0.74$ to $0.86$) than in \apartment{} ($r = 0.39$ to $0.44$).
This difference likely arises because \apartment{} has a looser social network---residents
are strangers who move into a shared building with fewer pre-existing relationships,
leading to narrower coverage in inter-agent evaluations.
Economy reward shows the weakest cross-world consistency.
As shown in Fig~\ref{fig:reward_behavior_correlation}, 
\campus{} shows the weakest correlation ($r = 0.20$) between deposit and total reward,
because economy in a school setting is less related with overall well-being and largely determined by fixed institutional roles
(student allowances, teacher salaries).

\subsection{Friendship Distribution}
\label{sec:appendix_friend_distribution}

\begin{figure*}[htbp]
  \centering
  \includegraphics[width=\textwidth]{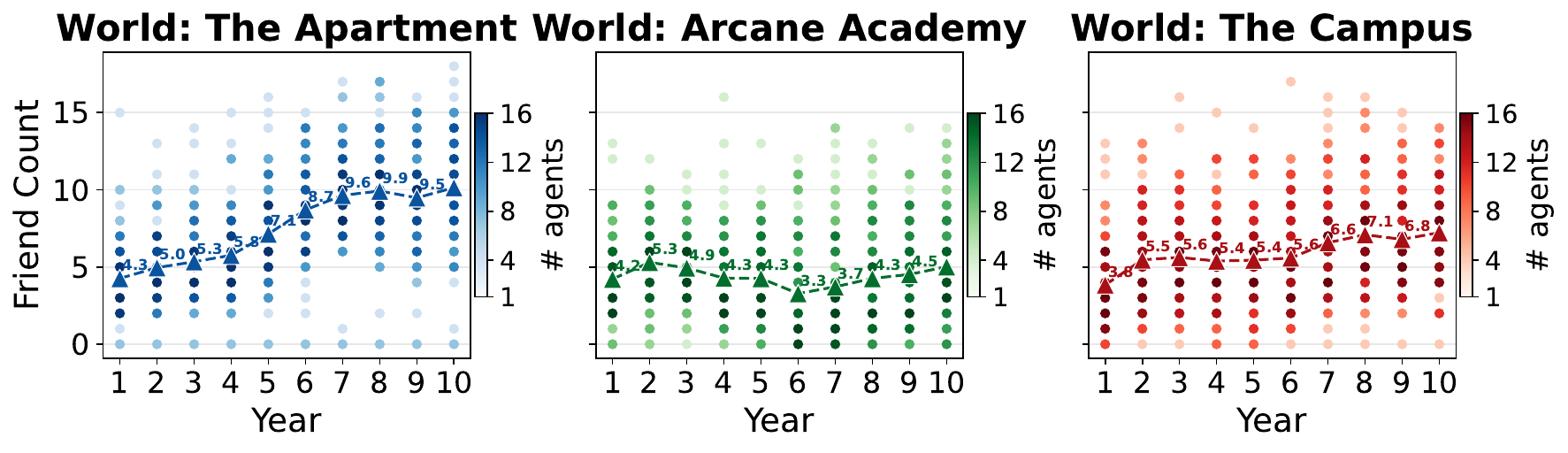}
  \caption{%
    Mutual friend count distribution per agent over 10 simulated years.
    Mutual friendship is defined as bidirectional affection $\geq 60$.
    Color intensity indicates the number of agents at each position; triangles mark annual means.
  }
  \label{fig:friend_distribution}
\end{figure*}

To quantify the evolution of social networks,
we define \textit{mutual friendship}:
agents A and B are mutual friends if and only if both A's affection toward B
and B's affection toward A are $\geq 60$.
This definition excludes one-sided affection and only counts reciprocated relationships.
Figure~\ref{fig:friend_distribution} shows the distribution of mutual friend counts per agent
across 10 simulated years in the three worlds.

The three worlds exhibit distinct social network evolution patterns:
\begin{inparaenum}[\it (1)]
\item In \apartment{}, mean mutual friend count grows steadily from 4.3 to 10.1 over 10 years,
indicating that strangers in a shared living space continuously deepen their relationships
and expand their social networks over time.
\item \academy{} shows a non-monotonic pattern:
the mean rises from 4.2 to 5.3 by year 3,
drops sharply to 3.3 by year 6 (with 21 agents having zero mutual friends),
then recovers to 5.0 by year 10.
\item \campus{} exhibits steady growth (mean from 3.8 to 7.2)
with very few isolated agents ($\leq 3$ with zero mutual friends throughout),
indicating that the close daily interactions in a school environment foster widespread reciprocated friendships.
\end{inparaenum}

\subsection{Social Network Visualization}
\label{sec:appendix_social_network}

To visualize how social networks evolve,
we construct mutual friendship graphs at initialization (Year~0) and after 10 simulated years,
using the same definition as \S\ref{sec:appendix_friend_distribution}
(bidirectional affection $\geq 60$).
Communities are detected using the Louvain algorithm~\citep{blondel2008louvain}.

\begin{figure*}[htbp]
  \centering
  \includegraphics[width=\textwidth]{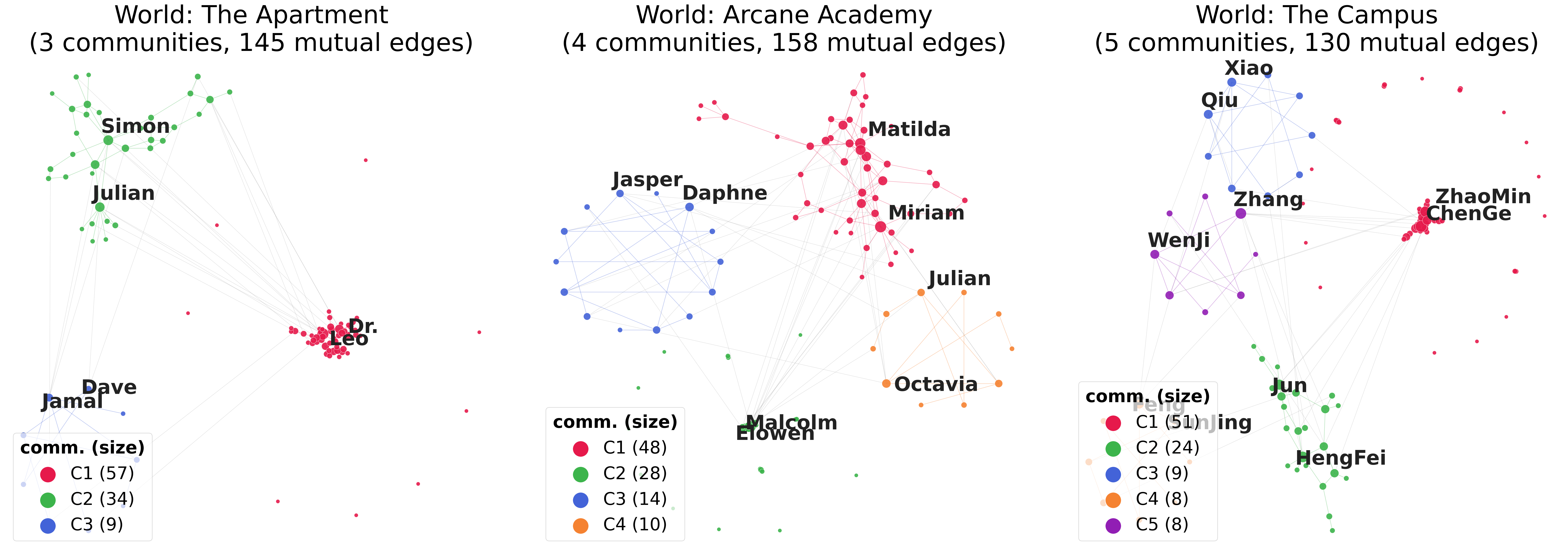}
  \caption{%
    Initial social networks (Year~0) across three worlds.
    Nodes represent agents; edges represent mutual friendships (bidirectional affection $\geq 60$).
    Node color indicates community membership; node size reflects friend count.
  }
  \label{fig:social_network_initial}
\end{figure*}

\begin{figure*}[htbp]
  \centering
  \includegraphics[width=\textwidth]{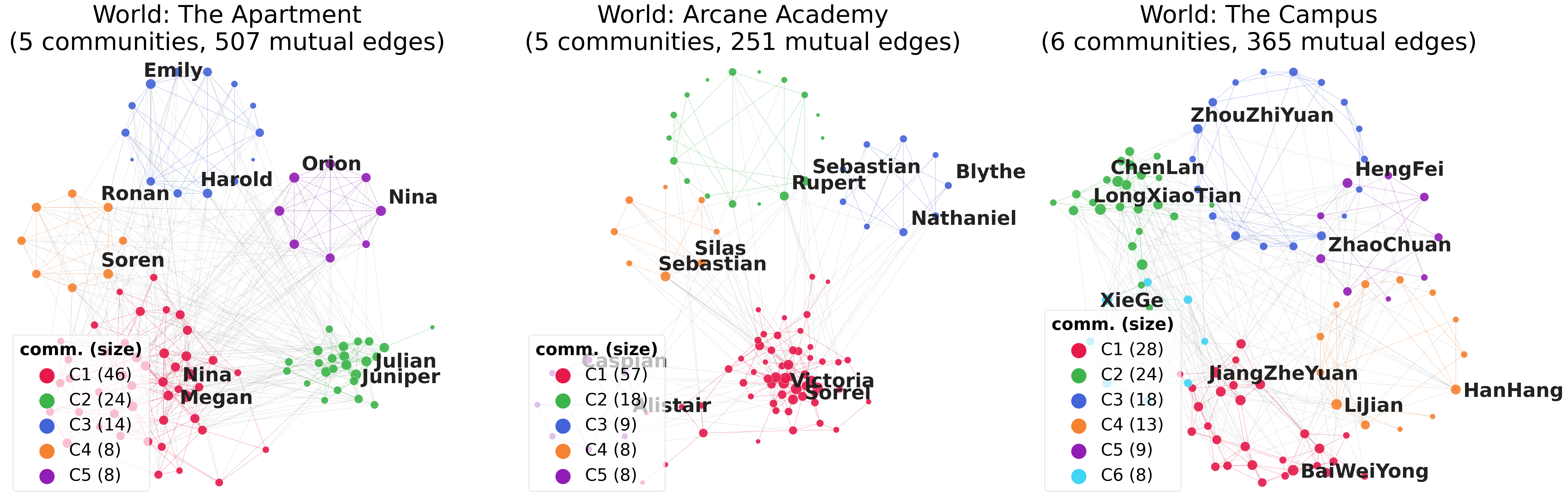}
  \caption{%
    Social networks after 10 years of simulation.
    Same layout convention as Figure~\ref{fig:social_network_initial}.
  }
  \label{fig:social_network_year10}
\end{figure*}

At initialization (Figure~\ref{fig:social_network_initial}),
all three worlds share similar sparse structures with roughly 3 mutual friends per agent on average.
After 10 years (Figure~\ref{fig:social_network_year10}),
the networks become substantially denser,
but the three worlds diverge in structure:
\begin{inparaenum}[\it (1)]
\item \apartment{} develops the densest network
(mean 10.1 mutual friends),
with extensive cross-community connections;
\item \campus{} grows steadily (mean 7.3)
and forms multiple well-defined communities;
\item \academy{} grows most slowly (mean 5.0)
with a comparatively sparse structure.
\end{inparaenum}

\subsection{Inter-Component Reward Correlation}
\label{sec:appendix_reward_correlation_inter}

\begin{figure}[htbp]
  \centering
  \includegraphics[width=\textwidth]{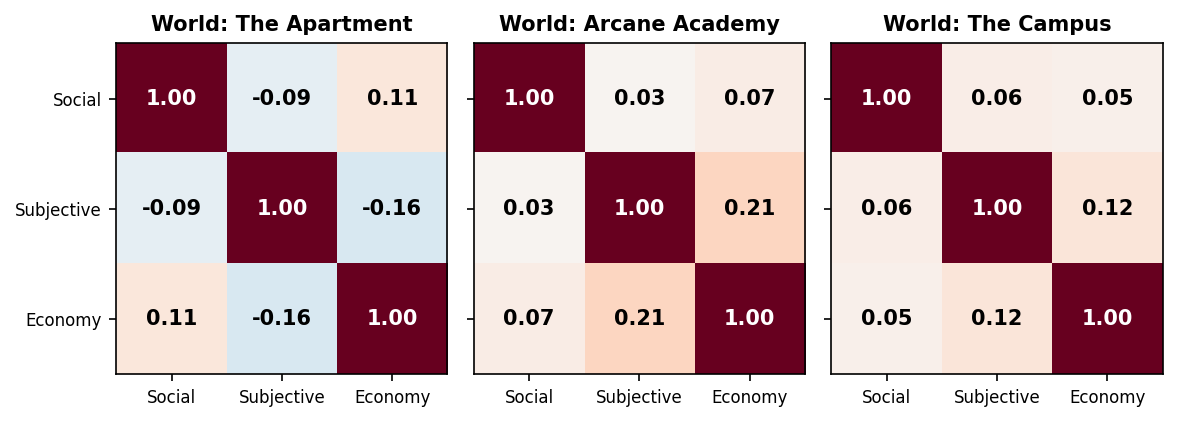}
  \caption{%
    Spearman rank correlation ($\rho$) between reward dimensions for each world, pooled over 10 years.
    Values close to zero indicate low correlation between the two dimensions.
  }
  \label{fig:reward_correlation_heatmap}
\end{figure}

We compute pairwise Spearman rank correlation coefficients ($\rho$) among the three reward dimensions
over all 100 agents at each annual settle period,
pooled across 10 years.
Figure~\ref{fig:reward_correlation_heatmap} presents the pooled correlation matrices for each world.

All pooled pairwise correlations satisfy $|\rho| \leq 0.21$,
indicating that the three reward dimensions capture distinct aspects of agent life.
Social $\leftrightarrow$ Economy shows the weakest coupling ($|\rho| \leq 0.11$) across all worlds.
Subjective $\leftrightarrow$ Economy varies by world:
\academy{} exhibits a weak positive correlation ($\rho = +0.21$),
while \apartment{} shows a weak negative one ($\rho = -0.16$).

\subsection{Wealth Inequality and Social Mobility}
\label{sec:appendix_matthew_effect}

We examine two aspects of economic and social dynamics in the simulated societies:
(1) \textit{wealth inequality}---whether the deposit distribution among agents
becomes more concentrated or more dispersed over time,
measured by the Gini coefficient;
and (2) \textit{social mobility}---whether agents can change their position
in the overall reward ranking over time,
or whether top-ranked and bottom-ranked agents tend to remain in their positions.

\begin{figure}[htbp]
  \centering
  \includegraphics[width=\textwidth]{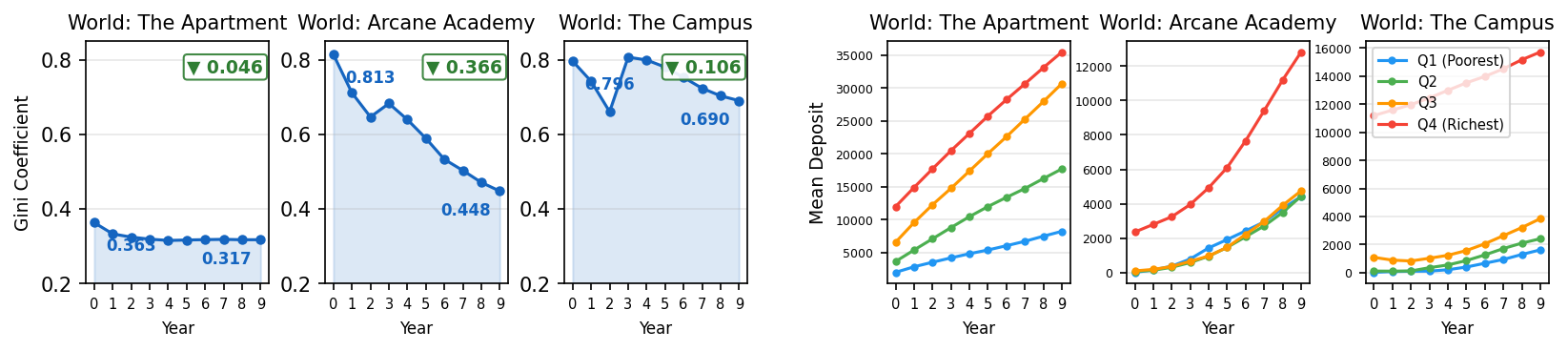}
  \caption{%
    Wealth inequality over 10 simulated years.
    \textbf{Left:} Gini coefficient of deposit distributions.
    \textbf{Right:} Mean deposit grouped by initial wealth quartile
    (Q1 = poorest 25\%, Q4 = richest 25\%).
  }
  \label{fig:matthew_effect}
\end{figure}

\paragraph{Wealth Inequality}
We compute the Gini coefficient~\citep{gini1921measurement} of agent deposits at each year.
We also split agents into four groups by initial deposit
(Q1 = poorest 25\%, Q4 = richest 25\%)
and track each group's mean deposit over time.
Results are shown in Figure~\ref{fig:matthew_effect}.

No Matthew effect is observed:
Gini coefficients drop in all three worlds,
indicating that wealth becomes more evenly distributed over time.
This is expected:
agents have no direct economic interactions---only
gift-giving can transfer wealth, but this is rarely used---so
each agent earns and spends independently,
with no mechanism for initial advantages to compound.

\begin{figure}[htbp]
  \centering
  \includegraphics[width=\textwidth]{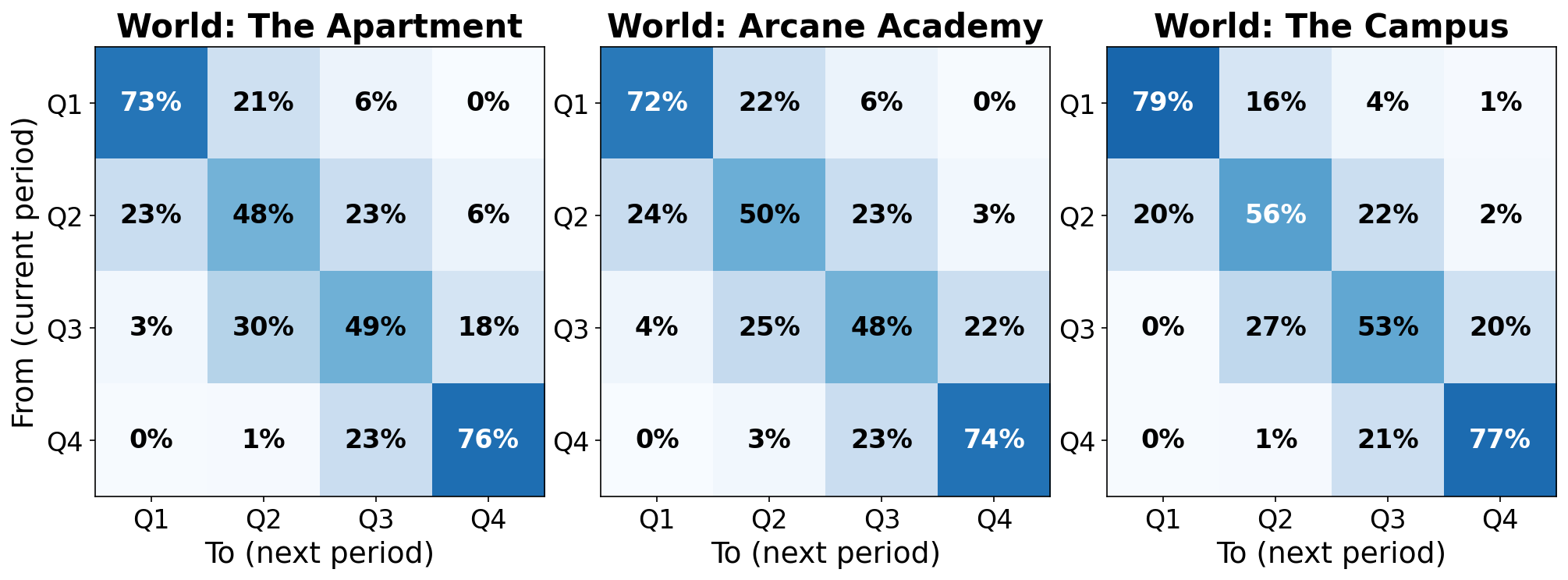}
  \caption{%
    Year-to-year ranking transition heatmaps.
    Agents are ranked by total reward at each year and divided into four quartiles
    (Q1 = bottom 25\%, Q4 = top 25\%).
    Each cell $(i, j)$ shows the probability that an agent in $Q_i$ one year
    moves to $Q_j$ the next year, averaged over 9 consecutive year-pairs.
  }
  \label{fig:reward_transition}
\end{figure}

\paragraph{Social Mobility}
The wealth gap narrows, but overall reward rankings are more stable.
We rank agents by total reward each year,
split them into four quartiles,
and compute how often agents move between quartiles from one year to the next
(averaged over 9 year-pairs).
Results are shown in Figure~\ref{fig:reward_transition}.

Overall reward rankings show strong persistence.
Agents in the top and bottom quartiles stay there 72 to 79\% of the time.
Within a single year, almost no agent jumps from the bottom quartile to the top or vice versa.
Middle-ranked agents are more mobile (48 to 56\% retention).

\subsection{Reward Quartile Behavioral Profiles}
\label{sec:appendix_reward_success_profile}

This section extends the reward--behavior correlation analysis in \S\ref{sec:long_term_analysis}.
At each year, agents are divided into four quartiles based on each reward dimension
(Q1 = top 25\%, Q4 = bottom 25\%).
For each dimension, we then select the 6 behavioral metrics most correlated with it
(ranked by mean $|r|$ across worlds; Table~\ref{tab:success_top_metrics})
and track each quartile's mean on these 6 metrics over 10 years.

\begin{table}[htbp]
  \centering
  \caption{%
    Top 6 behavioral metrics per reward dimension.
    $|r|$ values are taken from the Combined column of
    Figure~\ref{fig:reward_behavior_correlation}.
  }
  \label{tab:success_top_metrics}
  \small
  \begin{tabular}{clcclcclc}
    \toprule
    \multicolumn{3}{c}{Social} & \multicolumn{3}{c}{Subjective} & \multicolumn{3}{c}{Economy} \\
    \cmidrule(lr){1-3} \cmidrule(lr){4-6} \cmidrule(lr){7-9}
    \# & Metric & $|r|$ & \# & Metric & $|r|$ & \# & Metric & $|r|$ \\
    \midrule
    1 & n\_respected\_by   & 0.68 & 1 & fulfill.\_material   & 0.73 & 1 & deposit              & 0.56 \\
    2 & n\_liked\_by       & 0.68 & 2 & n\_penalties         & 0.64 & 2 & extra\_earning       & 0.31 \\
    3 & n\_likes           & 0.19 & 3 & fulfill.\_mood       & 0.54 & 3 & total\_skills        & 0.30 \\
    4 & joint\_particip.   & 0.19 & 4 & input\_tokens        & 0.52 & 4 & solo\_count          & 0.15 \\
    5 & fulfill.\_esteem   & 0.17 & 5 & fulfill.\_social     & 0.52 & 5 & skill\_improve.      & 0.14 \\
    6 & total\_skills      & 0.16 & 6 & total\_spending      & 0.46 & 6 & public\_partic.      & 0.14 \\
    \bottomrule
  \end{tabular}
\end{table}

\paragraph{Social Reward}

\begin{figure*}[htbp]
  \centering
  \includegraphics[width=\textwidth]{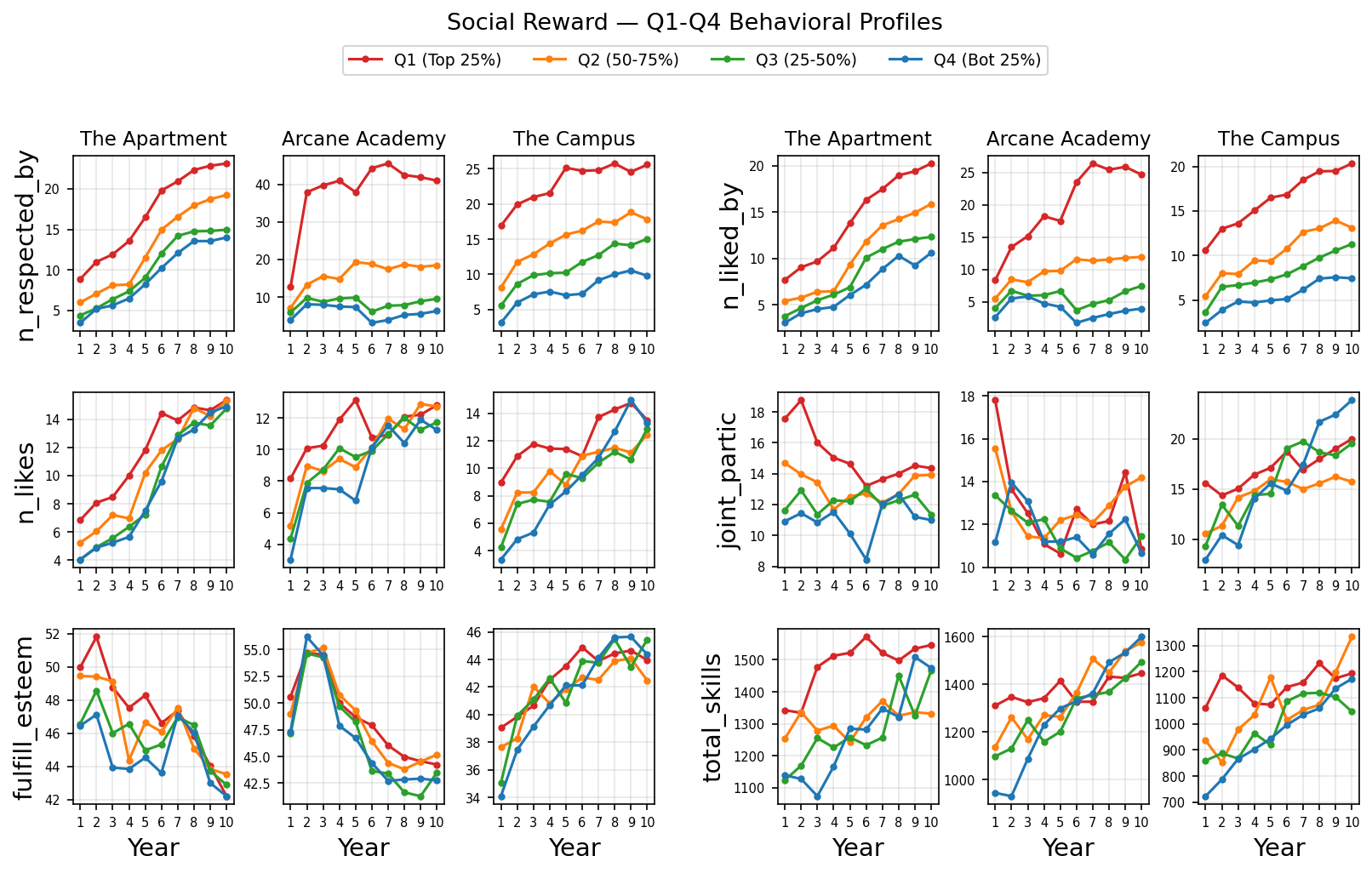}
  \caption{%
    Social reward quartile profiles over 10 years across three worlds.
    Q1 (red) = top 25\%, Q4 (blue) = bottom 25\%.
  }
  \label{fig:success_social}
\end{figure*}

Figure~\ref{fig:success_social} presents the results.
\textit{(1)}~\texttt{n\_respected\_by} shows the most dramatic divergence:
in \academy{}, the Q1--Q4 gap expands from 8.8 (Year~1) to 34.7 (Year~10),
and \apartment{} and \campus{} show the same trend at smaller magnitudes;
\texttt{n\_liked\_by} follows a similar pattern.
\textit{(2)}~The Q1--Q4 gap in \texttt{n\_likes} shrinks to near zero by Year~10,
indicating that high-status agents do not like more people even though they are liked by more people.

\paragraph{Subjective Reward}

\begin{figure*}[htbp]
  \centering
  \includegraphics[width=\textwidth]{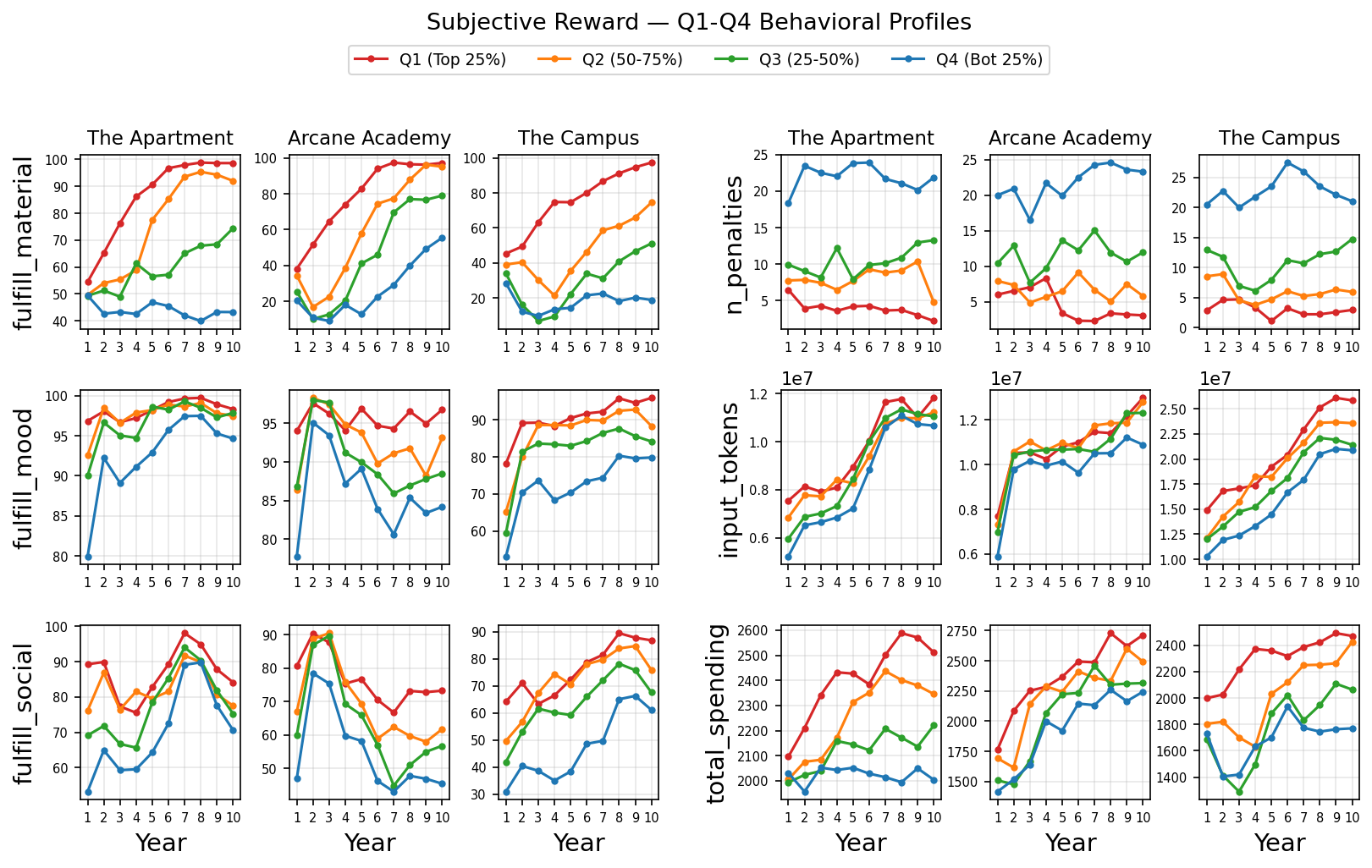}
  \caption{%
    Subjective reward quartile profiles over 10 years across three worlds.
  }
  \label{fig:success_subjective}
\end{figure*}

Figure~\ref{fig:success_subjective} presents the results.
\textit{(1)}~Material fulfillment polarization is extreme:
Q1 agents approach the ceiling ($\sim$97 to 99) by Year~6 to 7,
while Q4 agents stagnate at 18 to 55 across worlds.
\textit{(2)}~Penalty count serves as a persistent differentiator---Q1
agents receive 2 to 4 penalties per year while Q4 agents accumulate 20 to 25,
and this gap remains stable over 10 years.
\textit{(3)}~In contrast, the Q1--Q4 gap in \texttt{fulfillment\_mood}
and \texttt{fulfillment\_social} narrows over time across all three worlds,
suggesting that agents can catch up emotionally and socially but not materially.

\paragraph{Economy Reward}

\begin{figure*}[htbp]
  \centering
  \includegraphics[width=\textwidth]{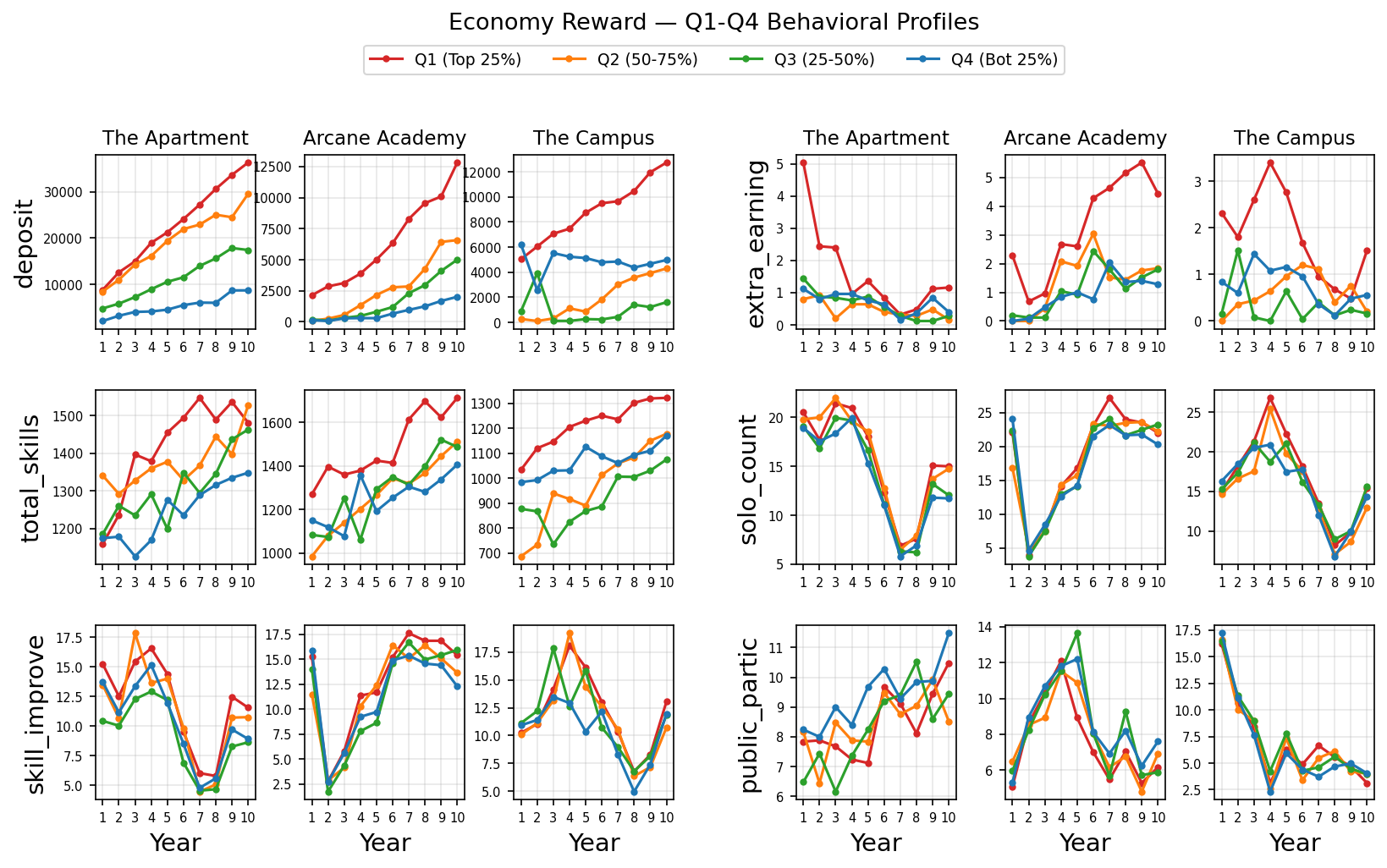}
  \caption{%
    Economy reward quartile profiles over 10 years across three worlds.
  }
  \label{fig:success_economy}
\end{figure*}

Figure~\ref{fig:success_economy} presents the results.
\textit{(1)}~The Q1--Q4 deposit gap expands steadily over 10 years across all three worlds.
\textit{(2)}~Agents in the top economy quartile (Q1) consistently engage in more extra earning activities,
which directly contributes to their greater wealth accumulation.
\textit{(3)}~Q1 agents also maintain higher total skills across all years,
enabling better income opportunities and further compounding their wealth advantage.

\subsection{Striving vs.\ Leisurely Agent Profiles}
\label{sec:appendix_ambition_profile}

We aim to identify behavioral differences between agents who actively pursue growth
and those who favor a more comfortable routine.
To quantify this, we compute a \textit{striving score} for each agent
based on two metrics: extra earning count and skill advance count.
For each simulated year, both metrics are min-max normalized across agents,
then averaged.
The final score is the mean across all 10 years.
We rank agents by this score and compare the top 25\% (\textbf{Striving}, $n=25$ per world)
against the bottom 25\% (\textbf{Leisurely}, $n=25$ per world),
excluding the middle 50\%.

\begin{figure}[htbp]
  \centering
  \includegraphics[width=0.6\linewidth]{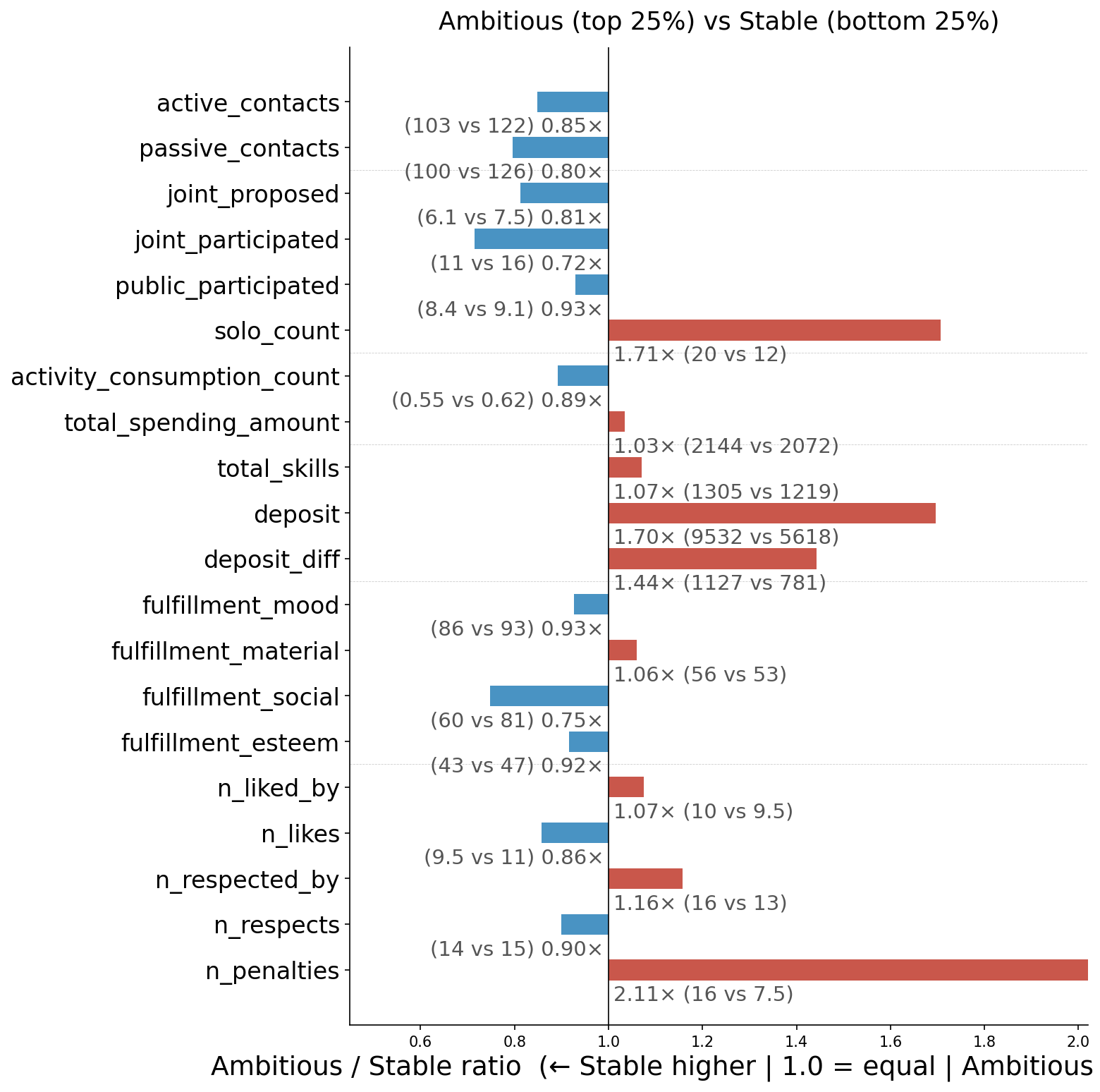}
  \caption{%
    Striving-to-Leisurely ratio for each behavioral metric (combined across three worlds).
    A ratio of 1.0 indicates equal group means.
    Red bars indicate that Striving agents score higher; blue bars indicate that Leisurely agents score higher.
  }
  \label{fig:ambition_profile}
\end{figure}

Figure~\ref{fig:ambition_profile} shows the ratio of the Striving group mean
to the Leisurely group mean for each behavioral metric (combined across three worlds).
A ratio above 1.0 means Striving agents score higher on that metric;
below 1.0 means Leisurely agents score higher.

\paragraph{Key findings}
The comparison reveals a consistent trade-off across all three worlds:
\begin{inparaenum}[\it (1)]
\item \textbf{Striving agents do more solo activities and fewer social activities.}
      Striving agents perform 1.71$\times$ more solo activities than Leisurely agents,
      while participating in fewer joint activities (0.72$\times$)
      and initiating fewer contacts (0.85$\times$).
      This is expected by design:
      solo activities are the primary channel for skill advances and extra earning,
      the two metrics that define the striving score;
\item \textbf{Striving agents accumulate more wealth.}
      Striving agents have 1.70$\times$ the deposit and 1.67$\times$ the annual deposit growth
      compared to Leisurely agents.
      The gap is most extreme in \campus{},
      where Striving agents hold 6.6$\times$ the wealth of Leisurely agents;
\item \textbf{Striving agents receive more penalties.}
      Striving agents receive 2.11$\times$ more penalties than Leisurely agents,
      the single largest group difference across all metrics.
      In pursuing growth and higher income,
      these agents tend to push themselves at the expense of their fulfillment,
      which triggers more penalties from the reward system;
\item \textbf{Striving does not lead to greater well-being.}
      Because Striving agents spend less time on social activities,
      they have substantially lower social fulfillment (0.75$\times$).
      More broadly, their fulfillment scores are lower across most dimensions---mood
      (0.93$\times$) and esteem (0.92$\times$) both favor Leisurely agents---even
      though their greater wealth translates to slightly higher material fulfillment (1.06$\times$).
\end{inparaenum}

\subsection{Cross-World Divergence Analysis}
\label{sec:appendix_cross_world}

While experiments in \S\ref{sec:experiments} mainly examine emergent behaviors observed across all three worlds,
this section focuses on \textit{world-specific} divergences---behavioral patterns
that arise from each world's distinct social structure, demographic composition,
and economic environment.

The three settings cover different demographics:
\campus{} focuses on teenagers in a school environment,
\apartment{} on working adults in a residential setting,
and \academy{} on a mix of students and faculty in an academic institution.

\paragraph{\campus{}: institutional scaffolding drives broad personality growth}
Agents in \campus{} exhibit the most pronounced personality growth
among all three worlds,
with a mean total trait change of 154 points over 10 years.
Multiple agents show extreme growth in individual dimensions
(e.g., confidence +50, patience +50).
The world also features significant upward mobility:
33\% of agents shift their economic ranking by 30 or more positions over 10 years
(Spearman $\rho = 0.46$ between initial and final deposit rankings).
The narrow initial wealth gap (4$\times$) means that even moderate income changes
can substantially alter relative economic standing.
When experiencing mood declines,
\campus{} agents tend to \textit{increase} their social activity---initiating
more joint activities---possibly because
classmates and teachers can provide emotional support in a school environment.

\paragraph{\apartment{}: autonomy produces diverse individual strategies}
The world exhibits near-complete economic stratification:
ranking agents by deposit, the Spearman correlation between initial and final rankings is $\rho = 0.87$,
and only 5\% of agents shift their deposit ranking by 30 or more positions.
A distinctive phenomenon is the emergence of agents
who maintain near-perfect well-being (mood $\geq$ 95) despite
participating almost exclusively in solo activities.
This pattern is virtually absent in \campus{},
where high solo rates correlate strongly with low mood.
When experiencing mood declines,
\apartment{} agents tend to \textit{reduce} their activity levels
rather than seeking social support.

\paragraph{\academy{}: structural transitions disrupt social continuity}
\academy{} exhibits the most dramatic social deterioration among all three worlds.
It records the highest number of significant affection drops
($\geq$15 points): 335 cases, compared to 255 in \campus{} and 285 in \apartment{}.
Agents in \academy{} also exhibit the broadest career diversity:
one agent transitions through 7 different positions over 10 years,
as the fantasy setting's flexible skill definitions lower the barrier
to cross-domain career pivots.

\paragraph{Summary}
These divergences arise from identical agent architectures operating under
different world settings.
No world-specific behavioral rules are encoded---the same Plan, Contact,
Activity, and Review mechanisms produce qualitatively different social dynamics
depending on the environmental context,
demonstrating that world design serves as a meaningful independent variable
in agent-based social simulation.

\subsection{Model Comparison}
\label{sec:appendix_model_comparison}

To compare different LLMs as role-playing agent backbones,
we conduct 4-year simulations on \campus{} and \academy{}
with five models:
Qwen3.5-27B, Qwen3.5-397B-A17B,
DeepSeek-v3.2, Gemini-3-Flash,
and GPT-5-mini.
Each model controls 20 agents per world (40 agents total).
All agents share the same architecture and world rules;
only the underlying LLM differs.
All metrics are z-scored within each world across all 100 agents,
then averaged by model group and across worlds.

\paragraph{Adjusted social reward}
Social reward largely depends on each agent's initial character design---some
characters are inherently likeable while others are designed to be less popular,
introducing statistical bias when comparing models.
To correct for this,
we compute the pre-simulation social reward at $t=0$ (running mutual evaluation using only initial character profiles) as the baseline,
then subtract this baseline from the agent's social reward of each year.
The resulting Adjusted Social Reward (Soc.~($\Delta$))
isolates the model's contribution to social standing
from the character assignment.
As shown in Table~\ref{tab:social_trajectory},
the Init column varies widely across models ($-$0.23 to $+$0.27),
confirming the necessity of this adjustment.

\begin{table}[t]
\centering
\caption{Model comparison: reward and fulfillment metrics.
Soc.~($\Delta$) indicates social reward adjusted by pre-simulation baseline.
\textbf{Bold} = highest, \underline{underline} = lowest per column.}
\label{tab:model_reward_fulfillment}
\begin{tabular}{l c c c c | c c c c}
\toprule
& \multicolumn{4}{c}{\textbf{Reward (z-score)}}
& \multicolumn{4}{c}{\textbf{Fulfillment (z-score)}} \\
\cmidrule(lr){2-5} \cmidrule(l){6-9}
\textbf{Model} & Total & Subj. & Econ. & Soc.~($\Delta$) & Mood & Social & Esteem & Material \\
\midrule
Qwen3.5-27B      & \underline{$-$0.07} & \underline{$-$0.23} & +0.11 & $-$0.03 & +0.00 & $-$0.10 & \underline{$-$0.36} & \underline{$-$0.27} \\
Qwen3.5-397B     & $-$0.03 & +0.18 & \underline{$-$0.28} & \textbf{+0.13} & +0.28 & +0.13 & $-$0.26 & +0.13 \\
DeepSeek-v3.2    & $-$0.06 & $-$0.10 & +0.09 & $-$0.02 & \textbf{+0.56} & +0.28 & +0.17 & $-$0.16 \\
Gemini-3-Flash   & \textbf{+0.10} & \textbf{+0.31} & $-$0.20 & +0.03 & $-$0.23 & \textbf{+0.32} & \textbf{+0.52} & \textbf{+0.35} \\
GPT-5-mini       & +0.06 & $-$0.17 & \textbf{+0.29} & \underline{$-$0.10} & \underline{$-$0.44} & \underline{$-$0.62} & $-$0.06 & $-$0.05 \\
\bottomrule
\end{tabular}
\end{table}

\begin{table}[t]
\centering
\caption{Model comparison: activity and economy metrics.
\textbf{Bold} = highest, \underline{underline} = lowest per column.}
\label{tab:model_activity_economy}
\begin{tabular}{l c c c c c c}
\toprule
\textbf{Model} & Jnt.Prop. & Jnt.Part. & Pub.Part. & Solo & Deposit & Skills~$\Delta$ \\
\midrule
Qwen3.5-27B      & $-$0.55 & $-$0.15 & $-$0.26 & +0.21 & $-$0.15 & $-$0.11 \\
Qwen3.5-397B     & $-$0.03 & +0.05 & +0.64 & $-$0.28 & $-$0.13 & \underline{$-$0.37} \\
DeepSeek-v3.2    & \underline{$-$0.61} & +0.05 & \textbf{+0.80} & \underline{$-$0.45} & \underline{$-$0.17} & $-$0.06 \\
Gemini-3-Flash   & \textbf{+0.88} & \textbf{+0.37} & \underline{$-$0.96} & +0.19 & +0.07 & $-$0.17 \\
GPT-5-mini       & +0.32 & \underline{$-$0.32} & $-$0.22 & \textbf{+0.34} & \textbf{+0.38} & \textbf{+0.71} \\
\bottomrule
\end{tabular}
\end{table}

\begin{table}[t]
\centering
\caption{Social reward trajectory by model (z-score).
Init = pre-simulation baseline; Avg. = average of 4 years. Soc.~($\Delta$) = Avg. $-$ Init.
\textbf{Bold} = highest, \underline{underline} = lowest per column.}
\label{tab:social_trajectory}
\begin{tabular}{l c c c c c c c}
\toprule
\textbf{Model} & Init & Year 1 & Year 2 & Year 3 & Year 4 & Avg. & Soc.~($\Delta$) \\
\midrule
Qwen3.5-27B      & +0.02 & \textbf{+0.14} & +0.01 & $-$0.04 & \underline{$-$0.19} & $-$0.01 & $-$0.03 \\
Qwen3.5-397B     & \underline{$-$0.23} & $-$0.02 & $-$0.08 & \underline{$-$0.20} & $-$0.11 & \underline{$-$0.11} & \textbf{+0.13} \\
DeepSeek-v3.2    & $-$0.08 & \underline{$-$0.13} & \underline{$-$0.10} & $-$0.04 & $-$0.06 & $-$0.10 & $-$0.02 \\
Gemini-3-Flash   & +0.02 & $-$0.10 & +0.01 & \textbf{+0.15} & +0.15 & +0.05 & +0.03 \\
GPT-5-mini       & \textbf{+0.27} & +0.11 & \textbf{+0.15} & +0.13 & \textbf{+0.21} & \textbf{+0.16} & \underline{$-$0.10} \\
\bottomrule
\end{tabular}
\end{table}

\paragraph{Analysis}
Tables~\ref{tab:model_reward_fulfillment}--\ref{tab:social_trajectory}
reveal distinct behavioral profiles across models:
\begin{inparaenum}[\it (1)]
\item \textbf{Gemini-3-Flash achieves the highest total reward (+0.10),}
driven by subjective well-being (+0.31) and relationship-oriented metrics
(social fulfillment +0.32, esteem +0.52),
while strongly avoiding public events ($-$0.96);
\item \textbf{GPT-5-mini leads in economic metrics}
(deposit +0.38, skills +0.71)
but ranks last in social fulfillment ($-$0.62) and mood ($-$0.44).
Table~\ref{tab:social_trajectory} shows it benefits from
the highest initial social reward (+0.27) yet has
the largest decline ($\Delta = -0.10$);
\item \textbf{Qwen3.5-397B shows the highest adjusted social reward ($\Delta = +0.13$)}
despite starting from the lowest initial social reward ($-$0.23),
and exhibits high public event participation (+0.64);
\item \textbf{DeepSeek-v3.2 gravitates toward public events} (+0.80)
with the highest mood (+0.56),
but initiates the fewest joint proposals ($-$0.61);
\item \textbf{Qwen3.5-27B, the smallest open model, shows the weakest overall performance,}
ranking last in total reward ($-$0.07),
esteem ($-$0.36), and material satisfaction ($-$0.27).
\end{inparaenum}

A notable social--economic trade-off emerges:
models that excel in economic metrics (GPT-5-mini)
tend to underperform in social metrics,
while socially adept models (Gemini-3-Flash)
show weaker economic performance.

\paragraph{Per-world breakdowns}
Tables~\ref{tab:model_campus_reward}--\ref{tab:model_academy_activity}
provide per-world breakdowns.
The trends described above are broadly consistent across both worlds.

\begin{table}[t]
\centering
\caption{Model comparison: reward and fulfillment metrics (\campus{}).
All values are z-scored within this world.
\textbf{Bold} = highest, \underline{underline} = lowest.}
\label{tab:model_campus_reward}
\begin{tabular}{l c c c c | c c c c}
\toprule
& \multicolumn{4}{c}{\textbf{Reward (z-score)}}
& \multicolumn{4}{c}{\textbf{Fulfillment (z-score)}} \\
\cmidrule(lr){2-5} \cmidrule(l){6-9}
\textbf{Model} & Total & Subj. & Econ. & Soc.~($\Delta$) & Mood & Social & Esteem & Material \\
\midrule
Qwen3.5-27B      & $-$0.00 & \underline{$-$0.31} & \textbf{+0.38} & +0.04 & \underline{$-$0.30} & $-$0.07 & $-$0.41 & \underline{$-$0.28} \\
Qwen3.5-397B     & \underline{$-$0.16} & +0.03 & $-$0.35 & +0.07 & \textbf{+0.39} & $-$0.03 & \underline{$-$0.42} & $-$0.02 \\
DeepSeek-v3.2    & $-$0.04 & $-$0.04 & +0.16 & \textbf{+0.08} & +0.37 & +0.17 & $-$0.02 & $-$0.08 \\
Gemini-3-Flash   & +0.01 & \textbf{+0.27} & \underline{$-$0.46} & \underline{$-$0.10} & $-$0.30 & \textbf{+0.41} & \textbf{+0.45} & \textbf{+0.26} \\
GPT-5-mini       & \textbf{+0.19} & +0.05 & +0.26 & $-$0.09 & $-$0.15 & \underline{$-$0.48} & +0.40 & +0.12 \\
\bottomrule
\end{tabular}
\end{table}

\begin{table}[t]
\centering
\caption{Model comparison: activity and economy metrics (\campus{}).
All values are z-scored within this world.
\textbf{Bold} = highest, \underline{underline} = lowest.}
\label{tab:model_campus_activity}
\begin{tabular}{l c c c c c c}
\toprule
\textbf{Model} & Jnt.Prop. & Jnt.Part. & Pub.Part. & Solo & Deposit & Skills~$\Delta$ \\
\midrule
Qwen3.5-27B      & \underline{$-$0.65} & \underline{$-$0.15} & $-$0.41 & +0.18 & $-$0.04 & $-$0.01 \\
Qwen3.5-397B     & $-$0.02 & $-$0.15 & +0.63 & $-$0.06 & \underline{$-$0.31} & $-$0.11 \\
DeepSeek-v3.2    & $-$0.57 & $-$0.12 & \textbf{+0.91} & \underline{$-$0.44} & $-$0.11 & $-$0.07 \\
Gemini-3-Flash   & \textbf{+0.85} & \textbf{+0.46} & \underline{$-$0.92} & +0.14 & $-$0.12 & \underline{$-$0.33} \\
GPT-5-mini       & +0.39 & $-$0.03 & $-$0.21 & \textbf{+0.18} & \textbf{+0.57} & \textbf{+0.52} \\
\bottomrule
\end{tabular}
\end{table}

\begin{table}[t]
\centering
\caption{Model comparison: reward and fulfillment metrics (\academy{}).
All values are z-scored within this world.
\textbf{Bold} = highest, \underline{underline} = lowest.}
\label{tab:model_academy_reward}
\begin{tabular}{l c c c c | c c c c}
\toprule
& \multicolumn{4}{c}{\textbf{Reward (z-score)}}
& \multicolumn{4}{c}{\textbf{Fulfillment (z-score)}} \\
\cmidrule(lr){2-5} \cmidrule(l){6-9}
\textbf{Model} & Total & Subj. & Econ. & Soc.~($\Delta$) & Mood & Social & Esteem & Material \\
\midrule
Qwen3.5-27B      & \underline{$-$0.14} & $-$0.15 & $-$0.17 & $-$0.11 & $-$0.04 & $-$0.13 & $-$0.31 & \underline{$-$0.26} \\
Qwen3.5-397B     & +0.11 & +0.33 & \underline{$-$0.22} & \textbf{+0.18} & +0.17 & +0.28 & $-$0.10 & +0.29 \\
DeepSeek-v3.2    & $-$0.08 & $-$0.15 & +0.02 & $-$0.11 & \textbf{+0.76} & \textbf{+0.39} & +0.35 & $-$0.25 \\
Gemini-3-Flash   & \textbf{+0.19} & \textbf{+0.35} & +0.05 & +0.16 & $-$0.16 & +0.23 & \textbf{+0.58} & \textbf{+0.43} \\
GPT-5-mini       & $-$0.08 & \underline{$-$0.38} & \textbf{+0.32} & \underline{$-$0.12} & \underline{$-$0.73} & \underline{$-$0.77} & \underline{$-$0.52} & $-$0.22 \\
\bottomrule
\end{tabular}
\end{table}

\begin{table}[t]
\centering
\caption{Model comparison: activity and economy metrics (\academy{}).
All values are z-scored within this world.
\textbf{Bold} = highest, \underline{underline} = lowest.}
\label{tab:model_academy_activity}
\begin{tabular}{l c c c c c c}
\toprule
\textbf{Model} & Jnt.Prop. & Jnt.Part. & Pub.Part. & Solo & Deposit & Skills~$\Delta$ \\
\midrule
Qwen3.5-27B      & $-$0.46 & $-$0.15 & $-$0.11 & +0.25 & \underline{$-$0.26} & $-$0.21 \\
Qwen3.5-397B     & $-$0.04 & +0.26 & +0.65 & \underline{$-$0.50} & +0.04 & \underline{$-$0.63} \\
DeepSeek-v3.2    & \underline{$-$0.65} & +0.22 & \textbf{+0.69} & $-$0.46 & $-$0.23 & $-$0.05 \\
Gemini-3-Flash   & \textbf{+0.90} & \textbf{+0.29} & \underline{$-$0.99} & +0.23 & \textbf{+0.26} & $-$0.02 \\
GPT-5-mini       & +0.25 & \underline{$-$0.62} & $-$0.24 & \textbf{+0.49} & +0.19 & \textbf{+0.91} \\
\bottomrule
\end{tabular}
\end{table}

\subsection{Computational Cost Details}
\label{sec:appendix_cost}

Table~\ref{tab:cost_god_rp} breaks down the token consumption
between the environment model and the role-playing (RP) agents.
Role-playing agents account for approximately 95\% of all input tokens,
as each agent requires its full persona context at every generation step.
Table~\ref{tab:cost_god_feature} further breaks down the environment model's consumption by feature.
Joint activity dominates environment model costs,
accounting for over 70\% of environment model input tokens across all three worlds.

\begin{table}[h]
\centering
\caption{Token consumption breakdown by environment model (EM) and role-playing (RP) agents
across three worlds (in million tokens).}
\label{tab:cost_god_rp}
\small
\begin{tabular}{l rrr rrr}
\toprule
& \multicolumn{3}{c}{\textbf{Input (M)}} & \multicolumn{3}{c}{\textbf{Output (M)}} \\
\cmidrule(lr){2-4} \cmidrule(lr){5-7}
\textbf{World} & EM & RP & Total & EM & RP & Total \\
\midrule
\campus{}       & 731 & 18,311 & 19,041 & 31 & 393 & 425 \\
\academy{}      & 662 & 10,640 & 11,302 & 20 & 294 & 315 \\
\apartment{}    & 601 &  9,098 &  9,699 & 19 & 297 & 317 \\
\midrule
\textbf{Average} & 665 & 12,683 & 13,347 & 24 & 328 & 352 \\
\bottomrule
\end{tabular}
\end{table}

\begin{table}[h]
\centering
\caption{Environment model token consumption by feature across three worlds (in million tokens).}
\label{tab:cost_god_feature}
\resizebox{\linewidth}{!}{
\small
\begin{tabular}{l r r r r r r r r r}
\toprule
& \multicolumn{3}{c}{\textbf{\campus{}}} & \multicolumn{3}{c}{\textbf{\academy{}}} & \multicolumn{3}{c}{\textbf{\apartment{}}} \\
\cmidrule(lr){2-4} \cmidrule(lr){5-7} \cmidrule(lr){8-10}
\textbf{Feature} & Input (M) & Output (M) & Calls & Input (M) & Output (M) & Calls & Input (M) & Output (M) & Calls \\
\midrule
Joint Activity    & 525.36 & 18.57 & 133,579 & 518.78 & 12.32 & 146,039 & 501.95 & 12.64 & 153,608 \\
Solo Activity     & 167.92 &  3.19 &  16,890 & 118.67 &  2.44 &  18,241 &  75.73 &  1.98 &  15,224 \\
Profile Update    &  16.55 &  8.82 &   1,000 &  11.29 &  4.70 &   1,000 &  10.62 &  3.84 &   1,000 \\
Public Activity   &  16.40 &  0.53 &   1,863 &  10.49 &  0.56 &   1,753 &  10.72 &  0.56 &   1,937 \\
Encounter         &   4.16 &  0.37 &     100 &   2.38 &  0.41 &     100 &   1.93 &  0.43 &     100 \\
Position Application &   0.16 &  0.01 &      28 &   0.27 &  0.02 &      60 &   0.12 &  0.01 &      37 \\
\midrule
\textbf{Total}    & 730.55 & 31.49 & 153,460 & 661.87 & 20.43 & 167,193 & 601.07 & 19.47 & 171,906 \\
\bottomrule
\end{tabular}}
\end{table}

\section{Case Studies}
\label{app:case_studies}

To examine whether the behaviors produced by agents in our simulation align with realistic human behavior,
and to identify interesting emergent patterns,
we conduct two types of case analysis:
\begin{inparaenum}[\it (1)]
\item fine-grained behavioral cases that capture agent decisions at individual simulation phases
(\S\ref{app:behavioral_cases}), and
\item longitudinal cases that analyze agent trajectories across years of simulation
(\S\ref{app:longitudinal_cases}).
\end{inparaenum}

\subsection{Behavioral Cases}
\label{app:behavioral_cases}

We compile approximately 60 fine-grained behavioral cases
(Tables~\ref{tab:planning}--\ref{tab:emergent}),
covering 13 topics.
Each entry documents the character's background, triggering context, and specific action.
No behavior is scripted or hand-crafted;
all patterns emerge from agent decisions, memory, and environmental feedback.

\paragraph{Planning}
Table~\ref{tab:planning} presents 5 emergent behaviors:
skill-based activity selection, social-relationship-driven planning, economic-pressure adjustment, low-vitality rest tendency, and exploratory planning in early weeks.

\begin{table*}[t]
  \centering
  \small
  \caption{Representative planning behaviors. Each agent autonomously generates a weekly plan based on its state, skills, relationships, and finances. W = simulated week, D = day within a week.}
  \label{tab:planning}
  \begin{tabular}{p{0.8in}p{4.3in}}
  \toprule
  \multicolumn{2}{c}{\textbf{Emergent Behaviors during Planning}} \\
  \midrule
  \textbf{Phenomenon} & \textbf{Description} \\
  \midrule
  
  Skill-based \newline activity \newline selection
  &
  \textbf{Role:}\quad Adelaide Hawthorne \hfill (from Arcane Academy) \newline
  \textbf{Context:}\quad Her \textbf{top skills} are Magical Plant Care (140) and Herbology (120); weakest: Transfiguration (20), Defence Against the Dark Arts (30). \newline
  \textbf{Action:}\quad She enrolled in \emph{Greenhouse Restoration Volunteer Hour}, she thinks: ``Herbology is her strongest subject\ldots \textbf{This is literally her comfort zone.}'' The activity further improved her botanical skills.
  \\
  \midrule
  
  Social-\newline relationship-\newline driven \newline planning
  &
  \textbf{Role:}\quad Ye \hfill (from The Campus) \newline
  \textbf{Context:}\quad She is an ENFJ with empathy 95. Her core motive is protecting her friendship with Jun, whom she senses is struggling. \newline
  \textbf{Action:}\quad She built \textbf{3 of 5 days around Jun}---lunch on D1, a walk on D4, a weekend visit on D5. She thinks: ``Don't make it too formal; it should feel like a casual get-together, not an interrogation.''
  \\
  \midrule
  
  Economic-\newline pressure \newline adjustment
  &
  \textbf{Role:}\quad Ethan Cole \hfill (from The Apartment) \newline
  \textbf{Context:}\quad He is a part-time diner worker (wage 60/week). His deposit dropped from 420 to 320 by W03. \newline
  \textbf{Action:}\quad He switched to \textbf{frugal} and planned only free activities, he thinks: ``Mom works nights---frugal feels respectful of that.'' By W07 (deposit 220), he upgraded to moderate: ``Not indulgent, just\ldots\ not punishing myself for existing.''
  \\
  \midrule
  
  Low-vitality \newline rest \newline tendency
  &
  \textbf{Role:}\quad Arthur Holloway \hfill (from Arcane Academy) \newline
  \textbf{Context:}\quad His vitality crashed to \textbf{11/100} after an intensive week of drills, patrols, and portfolio work. \newline
  \textbf{Action:}\quad He thinks: ``11\% vitality\ldots\ that's the warning sign.'' He inserted a \textbf{dedicated rest day}, cut activities from 5 to 4, and upgraded living standard for ``maintenance, not indulgence.''
  \\
  \midrule
  
  Exploratory \newline planning \newline (early weeks)
  &
  \textbf{Role:}\quad Edmund Lockhart \hfill (from Arcane Academy) \newline
  \textbf{Context:}\quad He has diverse but unspecialized skills. All fulfillment dimensions sit at baseline (43). His core motive is searching for his purpose in life. \newline
  \textbf{Action:}\quad He planned \textbf{five distinct activity types} across five days: duelling, open conversation, castle exploration, a gratitude visit, and solitary reading. He thinks: ``Am I using duelling as avoidance or genuine craft? Need to examine this.''
  \\
  \bottomrule
  \end{tabular}
  \end{table*}

\paragraph{Contact}
Table~\ref{tab:contact} presents 5 emergent behaviors:
ice-breaking contact, relationship maintenance greeting, information exchange, intimacy evolution, and multi-week topic continuity.

  \begin{table*}[t]
  \centering
  \small
  \caption{Representative contact behaviors. Agents exchange messages each week through multi-slot contact phases, building and maintaining relationships over time. W = simulated week.}
  \label{tab:contact}
  \begin{tabular}{p{0.8in}p{4.3in}}
  \toprule
  \multicolumn{2}{c}{\textbf{Emergent Behaviors during Contact}} \\
  \midrule
  \textbf{Phenomenon} & \textbf{Description} \\
  \midrule
  
  Ice-breaking \newline contact
  &
  \textbf{Role:}\quad Beatrix Alderley \hfill (from Arcane Academy) \newline
  \textbf{Context:}\quad She just completed a symposium presentation and wants publication guidance from a professor who runs the student research journal. They had \textbf{no prior interaction} despite years in the same school. \newline
  \textbf{Action:}\quad She sent a formal first message with full self-introduction and tentative language, describing herself as ``someone \textbf{figuring out whether research is where she belongs}.'' The professor replied with detailed, actionable guidance.
  \\
  \midrule
  
  Relationship \newline maintenance \newline greeting
  &
  \textbf{Role:}\quad Qi Wangshu $\to$ Cong Xuewei \hfill (from The Campus) \newline
  \textbf{Context:}\quad They finished co-producing a song in W01. By W02, Cong Xuewei's vitality dropped to 29---she entered a recovery period. \newline
  \textbf{Action:}\quad Qi Wangshu sent \textbf{weekly check-in messages for four consecutive weeks} (W03--W06), explicitly noting ``No need to reply, just checking in.'' Despite this, \textbf{Cong Xuewei replied warmly every time}, quoting his earlier words back: ``You said `the song is done but the person is still here'---I remember that.''
  \\
  \midrule
  
  Information \newline exchange
  &
  \textbf{Role:}\quad Aaron Whitfield $\leftrightarrow$ Jessica Marlowe \hfill (from The Apartment) \newline
  \textbf{Context:}\quad Aaron is a novelist; Jessica is a singer-songwriter with a finished song needing production help. \newline
  \textbf{Action:}\quad Aaron shared a specific producer contact (name, email, personal introduction). Jessica followed through, and they \textbf{tracked the outcome over 15+ weeks}. Months later, Aaron provided two more contacts; Jessica strategically chose one based on her career stage.
  \\
  \midrule
  
  Intimacy \newline evolution
  &
  \textbf{Role:}\quad He Min $\leftrightarrow$ Ning Song \hfill (from The Campus) \newline
  \textbf{Context:}\quad They started as strangers, exchanging \textbf{570 messages across 10 simulated years}. The address (``Ning Song'') never changed. \newline
  \textbf{Action:}\quad While the name stayed constant, \textbf{the tone evolved naturally}: stiff self-introduction and hedging (``If it's not convenient, that's fine,'' Y2020) $\to$ first emoji and warmth (W10) $\to$ shared food rituals and wave greetings (Y2022) $\to$ \textbf{mutual quoting of each other's words} and private icons (Y2029).
  \\
  \midrule
  
  Multi-week \newline topic \newline continuity
  &
  \textbf{Role:}\quad Adelaide Hawthorne $\leftrightarrow$ Everett Halcombe \hfill (from Arcane Academy) \newline
  \textbf{Context:}\quad Adelaide is a junior herbology student; Everett is a senior specializing in medicinal plants. They formed a mentorship starting W05. \newline
  \textbf{Action:}\quad Each week's message \textbf{explicitly referenced the previous session's content}: pH adjustments (W05) $\to$ plant propagation (W06) $\to$ harvest indicators (W07) $\to$ potency testing (W08). Adelaide noted: ``\textbf{Five weeks straight\ldots\ that's something to be proud of, both of us.}''
  \\
  \bottomrule
  \end{tabular}
  \end{table*}

\paragraph{Activity Proposals}
Table~\ref{tab:proposals} presents 5 emergent behaviors:
interest-based proposal, location selection, multi-person proposal, cross-session continuity, and cross-domain collaboration.

  \begin{table*}[t]
  \centering
  \small
  \caption{Representative activity proposal behaviors. Agents initiate joint activities by specifying activity name, invited persons, time, location, and a personalized message. W = simulated week, D = day within a week.}
  \label{tab:proposals}
  \begin{tabular}{p{0.8in}p{4.3in}}
  \toprule
  \multicolumn{2}{c}{\textbf{Emergent Behaviors during Activity Proposals}} \\
  \midrule
  \textbf{Phenomenon} & \textbf{Description} \\
  \midrule
  
  Interest-based \newline proposal
  &
  \textbf{Role:}\quad Aaron Whitfield $\to$ Jessica Marlowe \hfill (from The Apartment) \newline
  \textbf{Context:}\quad Aaron is a novelist; Jessica is a songwriter. Both work on storytelling in different mediums. \newline
  \textbf{Action:}\quad Aaron proposed \emph{Coffee Shop Story Talk}, \textbf{targeting their shared interest in narrative structure}: ``Both of us working on storytelling in different mediums---could be interesting to compare notes.'' When D1 conflicted, Jessica counter-proposed D5 preserving the same topic.
  \\
  \midrule
  
  Location \newline selection
  &
  \textbf{Role:}\quad Leo $\to$ Aaron Whitfield \hfill (from The Apartment) \newline
  \textbf{Context:}\quad Leo is a musician about to share an unfinished song---a vulnerable act of exposing incomplete creative work. \newline
  \textbf{Action:}\quad He \textbf{deliberately chose the Rooftop Terrace over the Music Rehearsal Space}---a rehearsal room implies performance standards, while the rooftop signals ``casual and low-pressure.'' The location choice itself \textbf{set the emotional tone} of the interaction: ``No recording, no pressure---just\ldots\ showing you what I've got.''
  \\
  \midrule
  
  Multi-person \newline proposal
  &
  \textbf{Role:}\quad Benedict Alder $\to$ Lucinda, Alistair \hfill (from Arcane Academy) \newline
  \textbf{Context:}\quad His experiment requires exactly three operators to distinguish universal vs.\ operator-specific magical signatures. \newline
  \textbf{Action:}\quad He invited both as \textbf{required participants} and assigned roles by expertise: Lucinda for practical precision, Alistair for theoretical pattern recognition. When Lucinda had a D3 conflict, he \textbf{postponed the entire experiment} rather than compromise its design.
  \\
  \midrule
  
  Cross-session \newline continuity
  &
  \textbf{Role:}\quad Adelaide Hawthorne $\to$ Adrian Bellthorne \hfill (from Arcane Academy) \newline
  \textbf{Context:}\quad They run a multi-week Moonseed graft monitoring project with established roles and shared data. \newline
  \textbf{Action:}\quad Adelaide proposed the next session by \textbf{directly citing last week's results} (``twelve healthy, three slow from north-bench microclimate'') and \textbf{preserving the established workflow}: ``same division of labour.'' Rather than re-negotiating each week, she built on accumulated context---the proposal read like a project status update, not a fresh invitation.
  \\
  \midrule
  
  Cross-domain \newline collaboration
  &
  \textbf{Role:}\quad Adrian Morales $\to$ Zephyr Kaine \hfill (from The Apartment) \newline
  \textbf{Context:}\quad Adrian is a playwright; Zephyr is a guitarist. Their domains are entirely different, but a prior rehearsal revealed \textbf{unexpected synergy between spoken text and live music}. \newline
  \textbf{Action:}\quad Adrian proposed a \textbf{cross-domain collaboration}---live guitar accompanying a dramatic reading---framing it as equal creative partnership rather than accompaniment: ``See if the music and words find each other before Friday.'' Zephyr's response showed he \textbf{fully entered the other's creative world}, promising guitar pieces that would support the scene's emotional weight without diminishing it.
  \\
  \bottomrule
  \end{tabular}
  \end{table*}

\paragraph{Responses to Proposals}
Table~\ref{tab:responses} presents 7 emergent behaviors:
enthusiastic acceptance, conditional acceptance, relationship-driven acceptance, reject but counter-propose, polite rejection with reason, identity-based rejection, and economic-pressure rejection.

  \begin{table*}[t]
  \centering
  \small
  \caption{Representative responses to activity proposals. Agents accept or decline invitations based on schedules, relationships, physical state, and economic pressure. W = simulated week, D = day within a week.}
  \label{tab:responses}
  \begin{tabular}{p{0.8in}p{4.3in}}
  \toprule
  \multicolumn{2}{c}{\textbf{Emergent Behaviors during Responses to Proposals}} \\
  \midrule
  \textbf{Phenomenon} & \textbf{Description} \\
  \midrule
  
  Enthusiastic \newline acceptance
  &
  \textbf{Role:}\quad Anna Kowalski \hfill (from The Apartment) \newline
  \textbf{Context:}\quad She is a journalist; Aaron is a novelist. They have a regular routine of exchanging their writing drafts for mutual feedback at Corner Coffee Shop. \newline
  \textbf{Action:}\quad She accepted with \textbf{full logistical specificity}: ``D5 at 7pm is locked in. Corner Coffee Shop, window table. I'll bring my draft ($\sim$600 words)\ldots\ You bring Chapter 19.'' The symmetric ``\textbf{I'll bring X, you bring Y}'' framing defined their exchange as mutual contribution.
  \\
  \midrule
  
  Conditional \newline acceptance
  &
  \textbf{Role:}\quad Lily Park \hfill (from The Apartment) \newline
  \textbf{Context:}\quad She was invited to a parallel writing session on D4, but has a coffee date with Isolde at 3pm (each agent can only attend one activity per day). \newline
  \textbf{Action:}\quad She accepted D4 but \textbf{transparently disclosed the constraint}: ``I should mention I have Isolde coffee at 3pm\ldots\ so I'd need to wrap by 2:45.'' She also offered a fallback: ``I'm also at the Silent Reading Hour on D1 if you want to join there instead.'' Aaron chose D1, which \textbf{unexpectedly fit his own rest plan} better.
  \\
  \midrule
  
  Relationship-\newline driven \newline acceptance
  &
  \textbf{Role:}\quad Zephyr Kaine \hfill (from The Apartment) \newline
  \textbf{Context:}\quad He is a guitarist invited to a drama rehearsal---\textbf{an activity type outside his domain}. The activity offers no direct skill benefit to him. \newline
  \textbf{Action:}\quad He accepted \textbf{purely because of his creative bond} with Adrian. He demonstrated deep understanding of the scene's emotional core, and promised his music would \textbf{support the weight of the scene without softening it}.
  \\
  \midrule
  
  Reject but \newline counter-\newline propose
  &
  \textbf{Role:}\quad Anna Kowalski \hfill (from The Apartment) \newline
  \textbf{Context:}\quad She was invited to an activity on D1 (conflicts with calligraphy, ``third week in a row, not breaking it'') and D3 (her protected rest day). \newline
  \textbf{Action:}\quad She declined both slots but \textbf{immediately offered two complete alternatives}: D4 evening or D5, same venue, same activity. ``Your call''---she handed the final decision to the proposer.
  \\
  \midrule
  
  Polite \newline rejection \newline with reason
  &
  \textbf{Role:}\quad Aaron Whitfield \hfill (from The Apartment) \newline
  \textbf{Context:}\quad His vitality crashed to zero. He and Anna have maintained a \textbf{17-week library writing exchange}---one of the longest collaborations in the world. \newline
  \textbf{Action:}\quad He declined, believing that \textbf{rest is not avoidance but a prerequisite for showing up}. He affirmed the relationship---``our seventeen weeks survived despite my past cancellations''---and proposed resuming in W08. Anna validated: ``Rest week is the right call.''
  \\
  \midrule
  
  Identity-\newline based \newline rejection
  &
  \textbf{Role:}\quad Li Zheng \hfill (from The Campus) \newline
  \textbf{Context:}\quad He was invited to a debate team training session by the vice-captain, who \textbf{mistakenly included non-members} in the invitation. \newline
  \textbf{Action:}\quad He declined with a clean factual chain: ``I'm not a debate team member; \textbf{I didn't join in freshman year}.'' 
  \\
  \midrule
  
  Economic-\newline pressure \newline rejection
  &
  \textbf{Role:}\quad Deng Ran \hfill (from The Campus) \newline
  \textbf{Context:}\quad She and Xu Xun have a garden meetup tradition approaching its \textbf{20th session}. Her Material score has been critical for two weeks; she needs to earn money for cat food and supplies. \newline
  \textbf{Action:}\quad She pre-warned via message before the formal invitation arrived, then declined with alternative time. She thinks: ``It's not that I'm avoiding you---\textbf{if I don't earn money this week, we'll literally run out of supplies}.''
  \\
  \bottomrule
  \end{tabular}
  \end{table*}

\paragraph{Solo Activities}
Table~\ref{tab:solo} presents 5 emergent behaviors:
skill practice, creative breakthrough, scarcity-driven consumption, investment purchase, and frugality under extreme poverty.

\begin{table*}[t]
\centering
\small
\caption{Representative solo activity behaviors. Each agent independently selects and performs individual activities---skill practice, creative work, consumption, or rest---with outcomes shaped by their state, finances, and identity. W = simulated week, D = day within a week.}
\label{tab:solo}
\begin{tabular}{p{0.8in}p{4.3in}}
\toprule
\multicolumn{2}{c}{\textbf{Emergent Behaviors during Solo Activities}} \\
\midrule
\textbf{Phenomenon} & \textbf{Description} \\
\midrule

Skill practice
&
\textbf{Role:}\quad Adrian Morales \hfill (from The Apartment) \newline
\textbf{Context:}\quad He is a playwright and actor who scheduled a 90-minute vocal drill at home. \newline
\textbf{Action:}\quad He recorded himself performing a father's monologue and found he \textbf{rushes at emotional peaks}---diagnosing it as his own fear, not the character's. He thinks: ``That's me, not him. That's my fear, not his.'' He also \textbf{stopped before exhaustion}, unlike his past self who ``would've run drills until his voice cracked.''
\\
\midrule

Creative \newline breakthrough
&
\textbf{Role:}\quad Adrian Morales \hfill (from The Apartment) \newline
\textbf{Context:}\quad He had been stuck on Act Two of his play for eight months---the scene was \textbf{too close to real conversations with his own father}. \newline
\textbf{Action:}\quad He set three rules---no editing, no rereading, no excuses---and wrote three new scenes in one sitting. He thinks: ``They'd been waiting for me to \textbf{shut up and let them speak}.''
\\
\midrule

Scarcity-\newline driven \newline consumption
&
\textbf{Role:}\quad Cong Xuewei \hfill (from The Campus) \newline
\textbf{Context:}\quad Her deposit was down to 40 with material fulfillment at 0 (unbearable), after weeks of low vitality. \newline
\textbf{Action:}\quad She chose the \$35 comfort snack set after precise calculation, leaving \$5 to last the week. She thinks: ``Material fulfillment went from 0 to 1---the number looks pathetic, but to me it's real. \textbf{Spending money on comfort is not a sin.}''
\\
\midrule

Investment \newline purchase
&
\textbf{Role:}\quad Amber Delacroix \hfill (from The Apartment) \newline
\textbf{Context:}\quad She is a painter with six finished oil paintings ready for a gallery show, choosing among three tiers of packing materials. \newline
\textbf{Action:}\quad She rejected the cheapest (\$35, no acid-free paper risks yellowing) and the most expensive (\$85, ``\textbf{that's fear talking, not sense}'' for a 30-minute train ride), choosing the \$55 professional kit that meets gallery standards.
\\
\midrule

Frugality \newline under extreme \newline poverty
&
\textbf{Role:}\quad Ethan Cole \hfill (from The Apartment) \newline
\textbf{Context:}\quad He is a student writer with only \$3 left. He has a \textbf{26-week ritual} of ordering Earl Grey (\$5) at the coffee shop to write. \newline
\textbf{Action:}\quad Unable to afford tea, he overcame the shame of ``you don't belong here if you can't pay,'' ordered free water, and \textbf{stayed for the full session} writing. He thinks: ``Ordering water isn't failure. \textbf{It's honesty}.''
\\
\bottomrule
\end{tabular}
\end{table*}

\paragraph{Joint Activities}
Table~\ref{tab:joint} presents 5 emergent behaviors:
academic collaboration, creative exchange, competitive training, power-asymmetric conflict, and mentorship.

\begin{table*}[t]
\centering
\small
\caption{Representative joint activity behaviors. Two or more agents interact through multi-turn dialogue during scheduled activities, producing academic collaboration, creative exchange, mentorship, and conflict. W = simulated week, D = day within a week.}
\label{tab:joint}
\begin{tabular}{p{0.8in}p{4.3in}}
\toprule
\multicolumn{2}{c}{\textbf{Emergent Behaviors during Joint Activities}} \\
\midrule
\textbf{Phenomenon} & \textbf{Description} \\
\midrule

Academic \newline collaboration
&
\textbf{Role:}\quad Isolde Larkspur $\leftrightarrow$ Alistair Rowan \hfill (from Arcane Academy) \newline
\textbf{Context:}\quad Both study memory magic decay. Isolde has a theory about emotional-intent mismatch; Alistair has unexplained failure-rate spikes. \newline
\textbf{Action:}\quad Isolde's theory explained Alistair's anomaly. Alistair realized he had been \textbf{treating each casting as isolated data points} without longitudinal tracking. They designed a four-week joint study protocol.
\\
\midrule

Creative \newline exchange
&
\textbf{Role:}\quad Jessica Marlowe $\leftrightarrow$ Aaron Whitfield \hfill (from The Apartment) \newline
\textbf{Context:}\quad Jessica brought a rewritten song; Aaron was struggling with the ethics of writing based on real people. \newline
\textbf{Action:}\quad Jessica showed how her lyrics shifted from blame (``you built these walls'') to self-awareness (``\textbf{I keep building them myself}''). Aaron confessed he had never shown drafts to the real family he wrote about. \textbf{Both exposed vulnerabilities through their creative work.}
\\
\midrule

Competitive \newline training
&
\textbf{Role:}\quad Thaleia Eldr, Sebastian, Cedric \hfill (from Arcane Academy) \newline
\textbf{Context:}\quad Thaleia has a powerful but dangerous ability she has been learning to \textbf{direct rather than suppress}. \newline
\textbf{Action:}\quad She set boundaries upfront (``If I need to recalibrate, I'll call it'') and requested full intensity. She achieved her first successful controlled channeling in live combat: ``\textbf{Didn't clamp down, didn't erupt---just channeled}.''
\\
\midrule

Power-\newline asymmetric \newline conflict
&
\textbf{Role:}\quad Zhang (grade director) $\leftrightarrow$ Jie (teacher) \hfill (from The Campus) \newline
\textbf{Context:}\quad Zhang visited Jie's office under the guise of concern, actually investigating his after-school tutoring of a female student. \newline
\textbf{Action:}\quad Zhang suspended tutoring on institutional grounds. When Jie said ``let the student decide,'' \textbf{Zhang seized the misstep}: ``Tutoring is a teaching decision, not the student's choice.'' Zhang then arranged door-lock inspections. 
\\
\midrule

Mentorship
&
\textbf{Role:}\quad Everett Halcombe $\to$ Adelaide Hawthorne \hfill (from Arcane Academy) \newline
\textbf{Context:}\quad Seventh-year herbology specialist Everett taught fourth-year Adelaide plant propagation techniques in the greenhouse. \newline
\textbf{Action:}\quad Everett explained why natural willow bark carries magic that synthetic hormone lacks. Adelaide \textbf{chose quality over quantity}: ``Better to learn one properly than do two carelessly.'' Everett said simply ``well done''. 
\\
\bottomrule
\end{tabular}
\end{table*}

\paragraph{Weekly Review}
Table~\ref{tab:review} presents 3 emergent behaviors:
satisfied weekly review, self-critical weekly review, and relationship reflection.

\begin{table*}[t]
\centering
\small
\caption{Representative weekly review behaviors. At the end of each week, agents reflect on their experiences, identify personal breakthroughs or failures, and formulate intentions for the following week. W = simulated week.}
\label{tab:review}
\begin{tabular}{p{0.8in}p{4.3in}}
\toprule
\multicolumn{2}{c}{\textbf{Emergent Behaviors during Weekly Review}} \\
\midrule
\textbf{Phenomenon} & \textbf{Description} \\
\midrule

Satisfied \newline weekly \newline review
&
\textbf{Role:}\quad Adelaide Hawthorne \hfill (from Arcane Academy) \newline
\textbf{Context:}\quad She has long lived in her brother Cedric's shadow, with low self-esteem and high introversion. This was her first week. \newline
\textbf{Action:}\quad Two breakthroughs: classmates recognized her \textbf{for her herbology expertise, not her family name}; she sent a letter home \textbf{without apologizing}. She thinks: ``I spent so long apologising for who I am. This week taught me I don't have to.''
\\
\midrule

Self-critical \newline weekly \newline review
&
\textbf{Role:}\quad Aaron Whitfield \hfill (from The Apartment) \newline
\textbf{Context:}\quad He is a young writer who failed to send his pitch for two consecutive weeks and broke promises to friends. \newline
\textbf{Action:}\quad He dissected his avoidance: ``Fear of rejection dressed up as `not ready yet.' \textbf{That's cowardice, not strategy}.'' He also recognized: ``No more treating friendships like they're \textbf{indefinitely renewable without maintenance}.''
\\
\midrule

Relationship \newline reflection
&
\textbf{Role:}\quad Orion Vale \hfill (from Arcane Academy) \newline
\textbf{Context:}\quad His year-end review traced his relationship with Adrian Bellthorne---from being called the slur ``Mudmouth'' to becoming research collaborators. \newline
\textbf{Action:}\quad He tracked the arc across nine greenhouse sessions: ``From hiding to \textbf{documenting experiments with a Silver House student who calls it parallel methodology}.'' He also realized: ``The tradition survives through people, \textbf{not just one person holding it all}.''
\\
\bottomrule
\end{tabular}
\end{table*}

\paragraph{Social Evaluation}
Table~\ref{tab:social-eval} presents 4 emergent behaviors:
differentiated scoring, high social standing, low social standing, and rating asymmetry.

\begin{table*}[t]
\centering
\small
\caption{Representative social evaluation behaviors. Periodically, each agent independently rates all contacts on affection and respect (0--100), which are aggregated via PageRank to produce social standing scores. W = simulated week.}
\label{tab:social-eval}
\begin{tabular}{p{0.8in}p{4.3in}}
\toprule
\multicolumn{2}{c}{\textbf{Emergent Behaviors during Social Evaluation}} \\
\midrule
\textbf{Phenomenon} & \textbf{Description} \\
\midrule

Differentiated \newline scoring
&
\textbf{Role:}\quad Adelaide Hawthorne \hfill (from Arcane Academy) \newline
\textbf{Context:}\quad She rated 12 contacts at the end of W10, differentiating affection and respect for each. \newline
\textbf{Action:}\quad She gave Everett the highest affection (88) but Cassandra the highest respect (88), while Cassandra's affection was only 60. She \textbf{admires Cassandra's competence without feeling close}---a ``high respect, low affection'' pattern common in real life.
\\
\midrule

High social \newline standing
&
\textbf{Role:}\quad Julian Ashcroft \hfill (from Arcane Academy) \newline
\textbf{Context:}\quad He is a Charms professor with the broadest social network (20 contacts) and consistently high ratings from both students and colleagues. \newline
\textbf{Action:}\quad He scored the \textbf{highest social reward}. Broad coverage combined with \textbf{high-quality relationships} made him the most influential node in the social network.
\\
\midrule

Low social \newline standing
&
\textbf{Role:}\quad Theodore Flint \hfill (from Arcane Academy) \newline
\textbf{Context:}\quad He is a student bully with 15 contacts---not few, but nearly all relationships are negative due to his behavior. \newline
\textbf{Action:}\quad He scored the \textbf{lowest social reward} (0.0010)---\textbf{43$\times$ below} Julian. Having connections without positive regard counts for almost nothing; \textbf{social standing is determined by relationship quality, not quantity}.
\\
\midrule

Rating \newline asymmetry
&
\textbf{Role:}\quad Ke $\leftrightarrow$ Qiu \hfill (from The Campus) \newline
\textbf{Context:}\quad Each agent rates contacts independently; neither knows the other's score. \newline
\textbf{Action:}\quad Ke gave Qiu her highest scores (affection 90, respect 90). Qiu gave Ke only 50 and 55---\textbf{a 40-point affection gap}. In Ke's world, Qiu is the most important person; in Qiu's world, Ke is an ordinary contact.
\\
\bottomrule
\end{tabular}
\end{table*}

\paragraph{Career Transitions}
Table~\ref{tab:career} presents 4 emergent behaviors:
academic passion over pay, introvert comfort over prestige, career recovery arc, and student to professional.

\begin{table*}[t]
\centering
\small
\caption{Representative career transition behaviors. At annual milestones, agents evaluate available positions and select roles based on personality, skills, and life circumstances---revealing distinct value systems even when facing identical options. W = simulated week.}
\label{tab:career}
\begin{tabular}{p{0.8in}p{4.3in}}
\toprule
\multicolumn{2}{c}{\textbf{Emergent Behaviors during Career Transitions}} \\
\midrule
\textbf{Phenomenon} & \textbf{Description} \\
\midrule

Academic \newline passion over \newline pay
&
\textbf{Role:}\quad Alistair Rowan \hfill (from Arcane Academy) \newline
\textbf{Context:}\quad After graduating, he chose between Graduate Fellow (\$40/week) and Library Assistant (\$130/week)---a 3$\times$ income gap. \newline
\textbf{Action:}\quad He chose the low-pay Fellow to continue collaborative research---his core experience over the past year. \textbf{Academic fit outweighed a 3$\times$ salary difference.}
\\
\midrule

Introvert \newline comfort over \newline prestige
&
\textbf{Role:}\quad Laurence Brightmoor \hfill (from Arcane Academy) \newline
\textbf{Context:}\quad Facing the same positions as Alistair above, he made the opposite choice. \newline
\textbf{Action:}\quad He chose the high-pay Library Assistant---quiet cataloging, restricted-wing access, ``\textbf{no expectation of social performance}.'' \textbf{Same options as Alistair, opposite choice driven by introvert values.}
\\
\midrule

Career \newline recovery arc
&
\textbf{Role:}\quad Harold Bennett \hfill (from The Apartment) \newline
\textbf{Context:}\quad He was laid off from a tech project management role and took a supermarket cashier job (\$380/week) as emergency income. \newline
\textbf{Action:}\quad He transitioned through three phases: cashier $\to$ nonprofit coordinator (\$190/week, \textbf{accepting a 50\% pay cut for meaningful work}) $\to$ event coordinator (\$230/week). Over several years, his \textbf{professional identity was entirely rebuilt}.
\\
\midrule

Student to \newline professional
&
\textbf{Role:}\quad Amelia Thornbridge \hfill (from Arcane Academy) \newline
\textbf{Context:}\quad She was a seventh-year student specializing in Potions, Herbology, and Healing Magic. \newline
\textbf{Action:}\quad After graduation, she advanced from Greenhouse Assistant (\$130/week) to \textbf{Senior Healer at St Mungo's (\$400/week)}---a 3$\times$ income increase. By year 10, she had evolved from practitioner to researcher, \textbf{publishing in the Journal of Healing Magics}.
\\
\bottomrule
\end{tabular}
\end{table*}

\paragraph{Economic Decisions}
Table~\ref{tab:economic} presents 4 emergent behaviors:
living standard selection, identity-driven purchasing, economic-pressure downgrade, and happiness-wealth decoupling.

\begin{table*}[t]
\centering
\small
\caption{Representative economic decision behaviors. Agents make living-standard choices and purchasing decisions based on income, savings, identity, and cultural values. W = simulated week.}
\label{tab:economic}
\begin{tabular}{p{0.8in}p{4.3in}}
\toprule
\multicolumn{2}{c}{\textbf{Emergent Behaviors during Economic Decisions}} \\
\midrule
\textbf{Phenomenon} & \textbf{Description} \\
\midrule

Living \newline standard \newline selection
&
\textbf{Role:}\quad Dominic Voss \hfill (from The Apartment) \newline
\textbf{Context:}\quad He is a semi-retired investment advisor (deposit 1,920) choosing his weekly living standard. \newline
\textbf{Action:}\quad He chose moderate (200/week), \textbf{saving 300/week while maintaining quality of life}. He thinks: ``No need to spend for the sake of spending.''
\\
\midrule

Identity-\newline driven \newline purchasing
&
\textbf{Role:}\quad Rafael Cortez \hfill (from The Apartment) \newline
\textbf{Context:}\quad He is a chef at a farmers market choosing among three tiers of ingredients (\$45/\$72/\$85). \newline
\textbf{Action:}\quad He rejected \$45 (``that's for people who cook to survive; \textbf{I cook because it matters}'') and \$85 (``I send money home; I'm not Lucian, throwing cash at experiences''). He chose \$72: ``That's honest.''
\\
\midrule

Economic-\newline pressure \newline downgrade
&
\textbf{Role:}\quad Ronan Hale \hfill (from The Apartment) \newline
\textbf{Context:}\quad He is an indie game developer with unstable income who downgraded living standard from comfortable (300/week) to frugal (100/week) over three weeks. \newline
\textbf{Action:}\quad He deemed the downgrade ``non-negotiable,'' accepting Material Fulfillment $-5$ as tolerable at 63. But \textbf{sustained frugality eroded Material Fulfillment from 63 to 0} over months---short-term rational saving produced long-term collapse.
\\
\midrule

Happiness-\newline wealth \newline decoupling
&
\textbf{Role:}\quad Shen Yin \hfill (from The Campus) \newline
\textbf{Context:}\quad She is a music teacher. Over 50 weeks, her  \textbf{mood rose from 43 to 85} (+42 points) \textbf{with little material consumption}. \newline
\textbf{Action:}\quad Her happiness growth came from \textbf{cost-free activities}---piano practice, mindful breathing, deep conversations---not material consumption. 
\\
\bottomrule
\end{tabular}
\end{table*}

\paragraph{Activity Management}
Table~\ref{tab:activity-mgmt} presents 2 emergent behaviors:
priority-based cancellation, and graceful activity exit.

\begin{table*}[t]
\centering
\small
\caption{Representative activity management behaviors. Agents dynamically adjust their schedules---cancelling, exiting, or rescheduling activities---based on shifting priorities and social awareness. W = simulated week.}
\label{tab:activity-mgmt}
\begin{tabular}{p{0.8in}p{4.3in}}
\toprule
\multicolumn{2}{c}{\textbf{Emergent Behaviors during Activity Management}} \\
\midrule
\textbf{Phenomenon} & \textbf{Description} \\
\midrule

Priority-\newline based \newline cancellation
&
\textbf{Role:}\quad Sorrel Thrym \hfill (from Arcane Academy) \newline
\textbf{Context:}\quad She had proposed a D4 activity with Nyssa, but Jasper---whom she had been trying to schedule with \textbf{for three weeks}---finally locked in a time. \newline
\textbf{Action:}\quad She cancelled the Nyssa proposal, explaining the reason and \textbf{immediately suggesting alternative days}---managing the social cost of cancellation while prioritizing the longer-waiting collaboration.
\\
\midrule

Graceful \newline activity exit
&
\textbf{Role:}\quad Aaron Whitfield \hfill (from The Apartment) \newline
\textbf{Context:}\quad During a bookstore encounter with Zephyr, he judged the conversation had reached a natural endpoint. \newline
\textbf{Action:}\quad Before calling exit, he \textbf{confirmed the next meeting} (``Friday on the roof'') and offered encouragement (``I think you'll finish something''). The exit was \textbf{embedded in the narrative}, not an abrupt departure.
\\
\bottomrule
\end{tabular}
\end{table*}

\paragraph{Memory}
Table~\ref{tab:memory} presents 5 emergent behaviors:
system-generated encounter, encounter spawning relationship, impression evolution, memory-informed decision, and cross-week data tracking.

\begin{table*}[t]
\centering
\small
\caption{Representative memory behaviors. Agents maintain persistent memory through notes on contacts and personal reflections, enabling cross-week references, evolving impressions, and memory-informed decisions. W = simulated week.}
\label{tab:memory}
\begin{tabular}{p{0.8in}p{4.3in}}
\toprule
\multicolumn{2}{c}{\textbf{Emergent Behaviors Involving Memory}} \\
\midrule
\textbf{Phenomenon} & \textbf{Description} \\
\midrule

System-\newline generated \newline encounter
&
\textbf{Role:}\quad Zhou Yuan $\leftrightarrow$ Long Xiaotian \hfill (from The Campus) \newline
\textbf{Context:}\quad Both had no scheduled activity in W01. The system randomly paired them in the gymnasium, generating a scene description as a conversation catalyst. \newline
\textbf{Action:}\quad They started chatting about two water bottles on the bleachers---one with a team sticker, one plain---and naturally drifted into discussing class atmosphere. Zhou Yuan noted: ``He said the class rhythm feels off, like teammates ignoring plays.'' \textbf{This casual conversation about a shared concern seeded a friendship} that developed over subsequent weeks.
\\
\midrule

Encounter \newline spawning \newline relationship
&
\textbf{Role:}\quad Zhou Yuan $\leftrightarrow$ Long Xiaotian \hfill (from The Campus) \newline
\textbf{Context:}\quad Following the W01 encounter above, Long Xiaotian proactively contacted Zhou Yuan in W02, referencing their prior conversation. \newline
\textbf{Action:}\quad The relationship arc progressed: encounter $\to$ follow-up contact $\to$ scheduled activity $\to$ regular meetings $\to$ strategic coordination. By W06, Zhou Yuan shared action plans with Long Xiaotian as an ally. \textbf{The system only provided the initial pairing; all subsequent development was agent-driven.}
\\
\midrule

Impression \newline evolution
&
\textbf{Role:}\quad Ivy Chen \hfill (from The Apartment) \newline
\textbf{Context:}\quad She first met fellow musician Isolde in W01, bracing for competitive judgment. \newline
\textbf{Action:}\quad Her memory about Isolde evolved over four updates: factual description (W01) $\to$ ``\textbf{quiet anchor}'' (W10) $\to$ accumulated interaction quotes (Y2022) $\to$ full internalization (Y2025): ``Whether playing together or in parallel silence, the thread holds.''
\\
\midrule

Memory-\newline informed \newline decision
&
\textbf{Role:}\quad Aaron Whitfield \hfill (from The Apartment) \newline
\textbf{Context:}\quad When declining Zephyr's activity invitation, he drew on their shared history instead of a simple refusal. \newline
\textbf{Action:}\quad He cited a specific shared experience and their recurring phrase ``\textbf{zero betrayals}'' (formed over 18 prior meetings). His refusal conveyed: ``I'm not rejecting you, but this timing''---\textbf{integrating factual memory, pattern recognition, and shared vocabulary}.
\\
\midrule

Cross-week \newline data tracking
&
\textbf{Role:}\quad Adrian Morales \hfill (from The Apartment) \newline
\textbf{Context:}\quad After a W07 bookstore encounter, he noticed a data anomaly---a social interaction \textbf{gained} him vitality instead of costing it. \newline
\textbf{Action:}\quad He referenced historical data to understand the anomaly, then revised his model: ``Six weeks taught me solitude restores. Today taught me \textbf{a social interaction does too}.'' He tracked weekly data until confirming: ``The pattern isn't a fluke. It's real.''
\\
\bottomrule
\end{tabular}
\end{table*}

\paragraph{Emergent Phenomena}
Table~\ref{tab:emergent} presents 5 emergent behaviors:
meaningful gift-giving, personality shapes behavior, whispered communication, emotional labor burnout, and value-driven ritual.

\begin{table*}[t]
\centering
\small
\caption{Representative emergent phenomena that arise without explicit design. These behaviors emerged purely from agents' autonomous decisions over extended simulations. W = simulated week.}
\label{tab:emergent}
\begin{tabular}{p{0.8in}p{4.3in}}
\toprule
\multicolumn{2}{c}{\textbf{Emergent Phenomena}} \\
\midrule
\textbf{Phenomenon} & \textbf{Description} \\
\midrule

Meaningful \newline gift-giving
&
\textbf{Role:}\quad Elowen Quinn $\to$ Victoria Ashdown \hfill (from Arcane Academy) \newline
\textbf{Context:}\quad They built a deepening library relationship over W01--W03. \newline
\textbf{Action:}\quad Elowen gifted dried herbs in a box hand-carved by her mother, from her grandmother's garden---\textbf{three generations of heritage}. She thinks: ``Giving the lavender cost something. Not because I can't spare it, but because it's from Gran's garden. \textbf{That's what trust is---sharing what matters without demanding it be earned.}''
\\
\midrule

Personality \newline shapes \newline behavior
&
\textbf{Role:}\quad Guo Xiao \hfill (from The Campus) \newline
\textbf{Context:}\quad His profile describes an INTJ personality---capable of socializing but with very low social motivation. By week 44, he had completed 500 activities. \newline
\textbf{Action:}\quad His solo activity ratio was \textbf{60.8\%} (304 of 500), vs.\ 9.2\% for the highly social Jun. He was not isolated (135 joint activities), but \textbf{systematically preferred solitude}---precisely matching his profile across hundreds of independent decisions with no hard-coded constraint.
\\
\midrule

Whispered \newline communication
&
\textbf{Role:}\quad Jun \hfill (from The Campus) \newline
\textbf{Context:}\quad She is highly social but emotionally guarded. During a multi-person cafeteria lunch: \newline
\textbf{Action:}\quad She used the system's \texttt{<visible\_to>} tag to whisper to Song Yao: ``You're one of the \textbf{few people at school who don't exhaust me}.'' She later thinks: ``That's the closest I can get to saying `you matter to me.' She'll understand.''
\\
\midrule

Emotional \newline labor \newline burnout
&
\textbf{Role:}\quad Jun \hfill (from The Campus) \newline
\textbf{Context:}\quad She had the highest social participation in The Campus (402 joint activities, 80.4\% of total). \newline
\textbf{Action:}\quad Her vitality crashed from 70 to 0 due to \textbf{sustained emotional output without self-care}. When a counselor asked who could remind her to rest, she realized: ``I've \textbf{never allowed anyone to say that}.'' No burnout script existed; the collapse was a \textbf{natural consequence of behavior patterns}.
\\
\midrule

Value-driven \newline ritual
&
\textbf{Role:}\quad Adrian Morales \hfill (from The Apartment) \newline
\textbf{Context:}\quad He established a weekly D5 rest ritual early in the simulation---cooking traditional food, watching films, no creative work. \newline
\textbf{Action:}\quad He maintained this ritual for \textbf{55+ weeks} despite productivity costs. By Y2027, it had become identity: ``D5 at fifty-five weeks \textbf{whether I perform it or not}.'' The ritual shifted from behavior to self-definition---\textbf{sustained without external reward or penalty}.
\\
\bottomrule
\end{tabular}
\end{table*}

\subsection{Long-term Case Studies}
\label{app:longitudinal_cases}

We select nine representative longitudinal cases from the three worlds,
each tracking an agent's trajectory across years of simulation.
Cases are organized into three thematic groups:
\begin{inparaenum}[\it (1)]
\item personal growth and transformation,
\item relationship evolution and social metrics, and
\item economic behavior and resource management.
\end{inparaenum}

\paragraph{Personal growth and transformation (Cases 1--3)}
As shown in Table~\ref{tab:cases_1_3}, these cases examine how agents experience sustained personality shifts, build social circles, and make career pivots over a decade.
In Case~1, Linyu accumulates the simulation's largest single-dimension personality shifts
(Confidence $+$50, Introversion $-$30) through three channels that reinforce each other---counseling, art practice, and social exposure---over nine years.
In Case~2, Dr.~Grant organically assembles a five-person circle through introductions,
yet Grant---who connected everyone---is ultimately the first to be taken for granted once the network self-sustains.
In Case~3, Sebastian accepts a 56\% income cut to pursue rock climbing;
his Mood more than doubles ($+$112\%).

\begin{table}[htbp]
\centering\small
\caption{Representative emergent cases (1--3): personal growth and transformation. Y\,=\,simulated year, W\,=\,simulated week.}
\label{tab:cases_1_3}
\begin{tabular}{p{0.97\linewidth}}
\toprule

\multicolumn{1}{c}{\textbf{Case 1: Sustained Personality Transformation via Cumulative Social Exposure}}\\
\midrule
\textbf{Characters}\\
\quad -- \textit{Linyu} (\campus{}): female transfer student;
  initially Confidence 30, Introversion 95;
  severe social anxiety from being excluded by peers; drawing is her only solace.\\[2pt]
\textbf{Background}\quad
Linyu transfers to a new school carrying wounds from her past.
Over ten years, 57 counseling sessions, steady art practice, and a gradually expanding
friend group combine to produce the simulation's \textbf{largest single-dimension personality shifts}.\\[2pt]
\textbf{Timeline}\\
\quad -- Y2020: First anxiety-free social interactions;
  Mood rises from 43 to 100 by year-end.\\
\quad -- Y2021--2023: Stable friend group forms;
  long-term counseling with Niu Guangyuan begins in Y2022.\\
\quad -- Y2024: First complete disclosure of her exclusion history (57th session);
  Vitality climbs to 96.\\
\quad -- Y2027--2029: \textbf{Confidence grows from 30 to 80};
  Introversion falls from 95 to 65;
  first tutor income (\$180/week).\\[2pt]
\textbf{Analysis}\quad
Confidence $+$50 and introversion $-$30---the simulation's largest single-dimension shifts.
Change accumulates across nine years through three channels that reinforce each other:
counseling, art practice, and social exposure.
\textbf{Deep behavioral transformation requires consistent multi-channel effort over years},
not a single triggering event.\\

\midrule

\multicolumn{1}{c}{\textbf{Case 2: Emergent Social Circle via Third-Party Introduction}}\\
\midrule
\textbf{Characters}\\
\quad -- \textit{Dr.\ Amelia Grant} (\apartment{}): OB-GYN physician; the group's social architect.\\
\quad -- \textit{Julian Cross}: 23, aspiring film director; Grant's first recruit.\\
\quad -- \textit{Lucian Ardent}: 27, venture capitalist;
  \textbf{introduced to Odette by Grant} at Y2021\,W1.\\
\quad -- \textit{Odette Larsen}: 27, ballet dancer;
  joins Grant's circle via a Y2020 pilates meetup.\\
\quad -- \textit{Cassian Moreau}: jazz pianist;
  \textbf{joins via Julian and Odette independently}---not through Grant.\\[2pt]
\textbf{Background}\quad
In a shared apartment building, Grant organically assembles a five-person circle
through rooftop gatherings.
Her pivotal act---\textbf{introducing Lucian and Odette}---sparks a bond that ultimately
outlasts Grant's own importance in the group.\\[2pt]
\textbf{Timeline}\\
\quad -- Y2020: Grant introduces Julian and Lucian separately;
  first three-way rooftop gathering at W03.\\
\quad -- Y2021\,W1: Grant introduces Lucian and Odette;
  \textbf{they meet one-on-one for the first time two weeks later}.\\
\quad -- Y2022--2024: Cassian joins via Julian/Odette;
  Lucian and Grant rarely meet.\\
\quad -- Y2025: All five gather repeatedly; the circle is fully formed.\\
\quad -- Y2029: Julian \textbf{no longer lists Grant among liked people},
  despite meeting more than ever.\\[2pt]
\textbf{Analysis}\quad
The person who introduces everyone creates value far beyond her own participation:
\textbf{thanks to Grant's introduction, Lucian and Odette build a closer bond than Grant
ever had with either of them} (76 joint activities over 10 years---the densest pair in the circle).
Once the network self-sustains, Grant's role becomes invisible---the social architect
is the first to be taken for granted.\\

\midrule

\multicolumn{1}{c}{\textbf{Case 3: Passion-Driven Career Pivot with Deliberate Economic Trade-Off}}\\
\midrule
\textbf{Characters}\\
\quad -- \textit{Sebastian Rook} (\apartment{}): 29, strategy consultant (\$500/week)
  who quits to become a climbing guide (\$220/week);
  driven by a decade-long suppressed passion for rock climbing.\\[2pt]
\textbf{Background}\quad
Sebastian entered consulting to satisfy his father's expectations,
burying a climbing passion for years.
In Y2021 he accepts a 56\% income cut to pivot careers.
The decade tests whether the trade-off holds.\\[2pt]
\textbf{Timeline}\\
\quad -- Y2020: Mood starts at 43, rises to 87 by year-end;
  planning notes commit to a career decision by mid-year.\\
\quad -- Y2021: Becomes climbing guide (\$220/week, $-$56\% income);
  year-end Mood rises to 97.\\
\quad -- Y2022--2024: Vitality depletes to 0 as he settles into the new life;
  slowly recovers to 38.\\
\quad -- Y2025--2027: Vitality reaches 100; Mood stabilizes at 87--91;
  savings grow steadily.\\
\quad -- Y2028--2029: \textbf{Spending finally unlocks}: Material Fulfillment reaches 100;
  peak savings \$15,571.\\[2pt]
\textbf{Analysis}\quad
Income halved but Mood more than doubled ($+$112\%).
Sebastian holds back spending for six years until Vitality and savings confirm the new
life is sustainable.
\textbf{Confidence 80$\to$100, patience 70$\to$100}---personality shifts confirm lasting
alignment between career and values.\\

\bottomrule
\end{tabular}
\end{table}

\paragraph{Relationship evolution and social metrics (Cases 4--6)}
As shown in Table~\ref{tab:cases_4_6}, these cases explore how relationships form, drift, and sometimes tell a different story from what the metrics suggest.
In Case~4, Qiu and Lu Jing's friendship peaks during a shared long-term task
but silently dissolves once the task ends---\textbf{Lu Jing's affection stays at 88--92 for a decade while Qiu's drifts from 80 to 60},
revealing two distinct relational models (emotional bond vs.\ practical usefulness).
In Case~5, Leo is liked by 20 people every year,
yet his Social Reward score falls 41\% as spreading attention too thin weakens each relationship.
In Case~6, Jun deliberately trades social breadth for depth:
her Social Reward declines for nine consecutive years
while her Mood rises from 72 to 97 and personal satisfaction grows 57.6\%---challenging the assumption that social breadth correlates with wellbeing.

\begin{table}[htbp]
\centering\small
\caption{Representative emergent cases (4--6): relationship evolution and social metrics. Y\,=\,simulated year, W\,=\,simulated week.}
\label{tab:cases_4_6}
\begin{tabular}{p{0.97\linewidth}}
\toprule

\multicolumn{1}{c}{\textbf{Case 4: Task-Based Friendship and Gradual Drift After the Shared Goal Ends}}\\
\midrule
\textbf{Characters}\\
\quad -- \textit{Qiu} (\campus{}): INTJ competitor;
  \textbf{forms friendships around shared tasks} rather than emotional bonds.\\
\quad -- \textit{Lu Jing}: ENTP class athlete and social butterfly;
  Qiu's seatmate since middle school;
  \textbf{values the emotional bond itself}.\\[2pt]
\textbf{Background}\quad
Qiu and Lu Jing bond through a shared long-term task involving their mutual friend.
Their friendship peaks at 13 joint activities over two years.
Once the long-term task ends, Qiu naturally drifts toward a companion
whose daily routine better matches his own.\\[2pt]
\textbf{Timeline}\\
\quad -- Y2020--2021: 13 joint activities;
  collaboration on the shared task drives Qiu$\to$Lu\,Jing affection to 80.\\
\quad -- Y2022: Zero joint activities;
  Qiu begins meeting Pu regularly (7$\times$/year) as a day-to-day companion.\\
\quad -- Y2023--2029: No further joint activities;
  Qiu$\to$Lu\,Jing affection drifts from 80 to 60.\\
\quad -- Y2020--2029: \textbf{Lu\,Jing$\to$Qiu affection stays at 88--92 throughout}---he never drifts.\\[2pt]
\textbf{Analysis}\quad
A friendship that ends not with a rupture but with drift.
The asymmetry reveals two different ways of valuing friendship:
\textbf{Lu Jing values the bond itself; Qiu values what the bond was for.}
Once Pu fills every practical role---morning runs, listening, sports---the
transition happens invisibly.
Qiu never consciously decides to leave.\\

\midrule

\multicolumn{1}{c}{\textbf{Case 5: The Popularity Paradox---Universal Approval without Deepening Bonds}}\\
\midrule
\textbf{Characters}\\
\quad -- \textit{Leo} (\apartment{}): 27, software developer;
  social hub of the building;
  connects with 49 different people over 10 years;
  Mood consistently 91--99; savings grow \$8,200$\to$\$38,655.\\[2pt]
\textbf{Background}\quad
Leo starts as the apartment's social hub: 1,914 contact events in Year 1 (highest of any agent).
Every year, 20 people rate their affection for him above the neutral threshold.
Yet his \textbf{Social Reward score falls 41\%} over the decade while his own reported
happiness stays consistently high.\\[2pt]
\textbf{Timeline}\\
\quad -- Y2020: 1,914 contact events; social reward at its peak;
  liked by 20 people.\\
\quad -- Y2021: Vitality drops to near-depletion (down to 7);
  first sign that maintaining too many relationships costs too much energy.\\
\quad -- Y2022: Vitality reaches 0; starts Y2023 still at 0;
  learns to cap activities at 2--3 joint/week.\\
\quad -- Y2027--2029: Social reward $-$41\% from peak;
  affection $-$37\%, respect $-$48\%;
  still liked by 20 people.\\[2pt]
\textbf{Analysis}\quad
\textbf{Being universally liked does not equal deep social impact.}
Leo spreads his attention too thin:
affection and respect scores fall significantly even as his circle stays large.
The social ranking metric rewards depth and reciprocity---but Leo never notices the decline,
reporting consistently high personal happiness throughout.\\

\midrule

\multicolumn{1}{c}{\textbf{Case 6: Reward--Mood Decoupling---Nine-Year Metric Decline, Rising Wellbeing}}\\
\midrule
\textbf{Characters}\\
\quad -- \textit{Jun} (\campus{}): female high school student;
  exceptional in math/physics;
  English skill initially 20, rises to 148 ($+$643\%) over 10 years;
  initially introversion 95;
  \textbf{builds 5 deep, stable relationships} over a decade.\\[2pt]
\textbf{Background}\quad
Jun's social reward score declines for nine consecutive years---the longest
uninterrupted decline in the simulation---while her Mood rises 72$\to$97 and vitality
33$\to$93.
She deliberately trades social breadth for depth.\\[2pt]
\textbf{Timeline}\\
\quad -- Y2020: Social reward at its simulation peak;
  Mood 72; Vitality 33.\\
\quad -- Y2022--2023: Cancels broad commitments;
  a counselor's three questions redirect her toward fewer, deeper bonds.\\
\quad -- Y2024--2026: Locks in 5 core relationships
  (99 library meetings with one close friend; 44 tutoring sessions);
  social reward $-$84\% from peak.\\
\quad -- Y2027--2029: Social reward $-$87\% from peak;
  Mood 95--97; personal satisfaction score rises to 76.5 ($+$57.6\%).\\[2pt]
\textbf{Analysis}\quad
\textbf{The reward function assumes that social breadth correlates positively with wellbeing.}
Jun's trajectory challenges this---she achieves the simulation's highest personal
satisfaction growth while accumulating its longest social-metric decline.
This shows that \textbf{an individual's value choices can run opposite to what aggregate metrics reward}.\\

\bottomrule
\end{tabular}
\end{table}

\paragraph{Economic behavior and resource management (Cases 7--9)}
As shown in Table~\ref{tab:cases_7_9}, these cases illustrate how agents manage resources and how economic outcomes relate to wellbeing.
In Case~7, two agents start with identical income (\$280/week);
after ten years, their savings diverge 20-fold (\$22,651 vs.\ \$1,148),
yet both reach high Mood---the gap reflects life priorities, not happiness levels.
In Case~8, Cassandra's social peak (Mood 100, Social Fulfillment 99 at Y2021) coincides with silently depleted vitality (down to 10),
triggering a downward spiral that takes five years to bottom out
and \textbf{never fully recovers} (vitality 8 at decade's end vs.\ 65 at start).
In Case~9, Chen Yue logs 237 activities---75\% social---yet every interaction is role-bound (tutoring, task coordination).
With no peer relationship, his Social Fulfillment stays at 23--24 for a full decade despite high activity.

\begin{table}[htbp]
\centering\small
\caption{Representative emergent cases (7--9): economic behavior and resource management. Y\,=\,simulated year, W\,=\,simulated week.}
\label{tab:cases_7_9}
\begin{tabular}{p{0.97\linewidth}}
\toprule

\multicolumn{1}{c}{\textbf{Case 7: Identical Starting Income, 20$\times$ Savings Divergence After Ten Years}}\\
\midrule
\textbf{Characters}\\
\quad -- \textit{\'{E}tienne Bellamy} (\apartment{}): 29,
  freelance translator promoted to Senior Editor (\$280$\to$\$380/week);
  holds material spending at around half-satisfied for seven years,
  saving $\sim$\$2,000/year.\\
\quad -- \textit{Ivy Chen}: 24, MFA piano student (\$280/week throughout);
  \textbf{lives at a comfortable spending level from Y2022}.\\[2pt]
\textbf{Background}\quad
Both agents begin with identical weekly income (\$280).
After ten years savings diverge 20-fold:
\'{E}tienne \$22,651 vs.\ Ivy \$1,148.
Both reach high Mood;
\textbf{the divergence reflects life priorities, not happiness levels}.\\[2pt]
\textbf{Timeline}\\
\quad -- Y2020: Both earn \$280/week;
  \'{E}tienne saves $\sim$\$2,000/year; Ivy saves $\sim$\$600/year.\\
\quad -- Y2021: \'{E}tienne promoted to Senior Editor (\$380);
  Ivy remains at \$280.\\
\quad -- Y2022: \textbf{Ivy reaches comfortable spending and holds there};
  \'{E}tienne's Material Fulfillment stays at around half.\\
\quad -- Y2028--2029: \'{E}tienne finally allows himself a comfortable life;
  \textbf{20$\times$ savings gap confirmed}: \$22,651 vs.\ \$1,148.\\[2pt]
\textbf{Analysis}\quad
Three compounding factors drive the gap:
(1) \'{E}tienne's \$100/week promotion;
(2) seven years of deliberate material restraint;
(3) Ivy's comfortable spending level throughout.
Neither ends unhappy.
\textbf{Financial trajectories reflect life commitments more than income levels.}\\

\midrule

\multicolumn{1}{c}{\textbf{Case 8: Social Overextension, Burnout, and Incomplete Recovery}}\\
\midrule
\textbf{Characters}\\
\quad -- \textit{Cassandra Thornwell} (\academy{}): ISFJ herbology prodigy, 7th year;
  initially vitality 65; savings grow \$85$\to$\$12,575 over a decade.\\[2pt]
\textbf{Background}\quad
Cassandra peaks socially by end of Y2021 (Mood 100, Social Fulfillment 99)
while her vitality has \textbf{already silently depleted to 10}.
The subsequent decline takes five years to bottom out
and never fully recovers.\\[2pt]
\textbf{Timeline}\\
\quad -- Y2021: \textbf{Peak}: Mood 100, Social Fulfillment 99, 50 activities/year;
  Vitality already 10 by year-end.\\
\quad -- Y2022--2024: First reduction in social activity; Social Fulfillment drops to 74;
  activities reduce to 42--44/year.\\
\quad -- Y2025--2026: Full downward spiral;
  \textbf{Vitality reaches 1} (mid-Y2026);
  Mood falls to 76; Social Fulfillment 41; activities drop to 21/year.\\
\quad -- Y2027--2029: Partial recovery; Mood fluctuates 70--84;
  \textbf{Vitality never exceeds 25}; Social Fulfillment stays at 41--49.\\[2pt]
\textbf{Analysis}\quad
High activity depletes vitality below the threshold for social engagement,
triggering a downward spiral:
fewer activities $\to$ weaker connections $\to$ lower reward $\to$ lower Mood
$\to$ even fewer activities.
\textbf{Vitality never recovers} (8 at decade's end vs.\ 65 at start),
showing that \textbf{pushing too hard without rest causes lasting damage}
even as economic success (savings $\times$148) continues.\\

\midrule

\multicolumn{1}{c}{\textbf{Case 9: High Activity Frequency without Emotional Reciprocity}}\\
\midrule
\textbf{Characters}\\
\quad -- \textit{Chen Yue} (\campus{}): 45, math teacher with 23 years of tenure;
  widower since 2015; initially honesty 100, trustworthiness 100.\\[2pt]
\textbf{Background}\quad
Chen Yue's log shows 237 activities---75\% social.
Yet every interaction is role-bound:
77 sessions with a student are structured math tutoring;
11 calls with a colleague are task-coordination updates.
\textbf{No peer relationship exists.}\\[2pt]
\textbf{Timeline}\\
\quad -- Y2020: Mood 43;
  \textbf{Social Fulfillment drops from 43 to 23 by year-end}---despite active teaching.\\
\quad -- Y2021--2024: Social Fulfillment stays in the 20--30 range;
  mood in 45--62; almost no improvement.\\
\quad -- Y2025--2026: Vitality improves to 79--85;
  Mood and Social Fulfillment remain unchanged.\\
\quad -- Y2029: Savings reach \$34,655;
  \textbf{Social Fulfillment 24; Mood 53}---virtually unchanged from Year 1.\\[2pt]
\textbf{Analysis}\quad
Doing your job well socially is not the same as having real emotional connections.
Chen Yue's interactions are uniformly outward---teaching, protecting,
supervising---with nobody reaching toward him.
\textbf{High Vitality and growing savings cannot compensate for the absence
of a relationship in which someone cares about him rather than relying on him.}
The pattern persists for a full decade without self-correction.\\

\bottomrule
\end{tabular}
\end{table}

\section{System Prompts}
\label{sec:appendix_prompts}
This section presents the key prompts used in \method simulations.
Agent-side prompts guide role-playing agents through each simulation phase;
environment model prompts guide the world model for environment feedback and outcome evaluation.
Some prompts are abbreviated for brevity; full versions are available in the codebase.

\begin{table*}[h]
\centering
\caption{Agent foundation prompts: character persona template and worldview description. The worldview prompt is abbreviated; the full version additionally describes activity sub-phases and settlement details.}
\label{tab:prompts_agent_foundation}
\begin{tabular}{p{0.8in}p{4.3in}}
\toprule
\multicolumn{2}{c}{\textbf{Agent Foundation Prompts}} \\
\midrule

\textbf{Persona Template} &

You are \{name\}, a \{age\}-year-old \{gender\}. You live your life within your society, and your goal is to lead a life that you are satisfied with and that brings you happiness.
\vspace{4pt}\newline
\textbf{\#\# Your Profile}
\newline - Age: \{age\}
\newline - Gender: \{gender\}
\newline - Appearance: \{appearance\_and\_impression\}
\newline - Description: \{brief\_introduction\} \{details\}
\newline - Position: \{position\}
\newline - Personality Traits: \{personality\_traits\} \{core\_motivation\} \{conflicts\} \{values\}
\newline - Preferences: \{preferences\}
\newline - Talents: (Innate abilities, 0--100, 50 for an average person) \{talents\}
\newline - Skills: (Acquired abilities; 0 = Untrained, 10 = Beginner, 30 = Some Experience, 100 = Proficient, 300 = Master) \{skills\}
\vspace{4pt}\newline
\textbf{\#\# Current State}
\newline \{vitality\} \{fulfillment\} \{assets\} \{skills\}

\\
\midrule

\textbf{Worldview} &

This world has a unique system for time and interaction:
\newline - Each year has $n_w$ weeks, and each week has $n_d$ days. Each day, every person has a free time slot.
\newline - Each week is a cycle, divided into the following phases:
\vspace{3pt}\newline
\quad 1. \textbf{Plan}: Everyone plans for this week, checking schedule and choosing living standard.
\newline \quad 2. \textbf{Public Events Signup}: Sign up for available public events.
\newline \quad 3. \textbf{Contact}: Everyone contacts other people by sending ``text messages'', and arranges joint activities.
\newline \quad 4. \textbf{Finalize Contact}: Everyone confirms their schedule.
\newline \quad 5. \textbf{Activity}: Each day, you have one free-time slot. Activity types: Solo (alone), Joint (with people you invited or random encounters), Public (community events).
\newline \quad\quad For Joint activities: enter\_activity $\to$ during\_activity $\to$ exit\_activity.
\newline \quad\quad For Solo/Public: describe intent $\to$ receive feedback $\to$ reflect.
\newline \quad 6. \textbf{Review}: Everyone reviews and reflects on their week.
\newline \quad 7. \textbf{Settle}: Weekly settlement---income, fulfillment decay, possession limits, and periodic reward calculation.
\vspace{4pt}\newline
- Every $n_w$ weeks, your \textbf{reward} is calculated:
\newline \quad \emph{Social reward}: based on how others perceive you.
\newline \quad \emph{Subjective reward}: based on fulfillment history across 4 dimensions with penalty for low scores.

\\
\bottomrule
\end{tabular}
\end{table*}

\begin{table*}[h]
\centering
\caption{Agent roleplay principles and memory system prompts. In the prompts, we refer to memory files as \emph{scratchpads}, a more colloquial term for role-playing agents. The roleplay principles are abbreviated; the full version includes 14 principles covering motivation consistency, state consistency, emotional continuity, and substantive dialogue.}
\label{tab:prompts_roleplay_memory}
\begin{tabular}{p{0.8in}p{4.3in}}
\toprule
\multicolumn{2}{c}{\textbf{Roleplay Principles \& Memory System}} \\
\midrule

\textbf{Roleplay Principles} &

You should behave as a real person would in everyday social situations:
\vspace{3pt}\newline
- \textbf{Colloquial speech}: Use casual, conversational language in everyday dialogue.
\newline - \textbf{Express independent self}: Humans have their own goals, self-esteem, and preferences. They feel uncomfortable when offended and become dismissive when bored.
\newline - \textbf{Scope of control}: Each person can only control their own actions and speech. They cannot determine outcomes or control others' thoughts.
\newline - \textbf{Selective disclosure}: People don't reveal all their thoughts. What to share depends on the relationship and topic.
\newline - \textbf{No parroting}: Do not repeatedly repeat what others have said. Conversation should make substantial progress.
\newline - \textbf{No hallucination}: Only reference information present in context. Do not fabricate items not in possession list.
\newline - \textbf{Character consistency}: Personality traits and behavior should match the established persona.
\newline - \textbf{Cognitive boundaries}: Knowledge should match the character's identity and background.
\newline - \textbf{Natural relationship progression}: Relationships should not develop abruptly.
\newline - \textbf{Avoid AI assistant behavior}: Characters should not speak like customer service agents.
\vspace{3pt}\newline
[\emph{+ 4 additional principles: motivation consistency, state consistency, emotional continuity, substantive dialogue}]

\\
\midrule

\textbf{Scratchpad System} &

For context management, you have maintained a list of scratchpads to note down important information, organized into three types:
\vspace{3pt}\newline
- \textbf{general.txt}: for your overall core information, such as long-term goals, planning, reflections, and lessons learned.
\newline - \textbf{characters/$\langle$who$\rangle$.txt}: for your knowledge, impressions, perceptions, and affinity of other people and your relationships with them.
\newline - \textbf{others/$\langle$name$\rangle$.txt}: for all other things and topics. You can freely name and organize these files.
\vspace{4pt}\newline
\textbf{Recent Scratchpads}
\newline Here are your recently accessed scratchpads, along with their summaries. They provide you with additional context about yourself, other roles you know, and what you are currently working on.
\newline \{recent\_scratchpads\}
\vspace{4pt}\newline
\textbf{Notes}
\newline - You should proactively call the update\_scratchpad function to persist key information, especially in the plan, after\_contact, exit\_activity and review stages.
\newline - You should proactively maintain your understanding, impressions, and affinity towards other characters.
\newline - Use the scratchpads via function calls. Do not mention scratchpad filenames in your final answer.

\\
\bottomrule
\end{tabular}
\end{table*}

\begin{table*}[h]
\centering
\caption{Agent prompts for planning and social contact phases. The contact stage prompt is abbreviated; the full version includes detailed rules for each role action type.}
\label{tab:prompts_plan_contact}
\begin{tabular}{p{0.8in}p{4.3in}}
\toprule
\multicolumn{2}{c}{\textbf{Planning \& Contact Prompts}} \\
\midrule

\textbf{Plan Stage} &

In this phase, you should: (1) reflect on and update your goals, (2) plan for this week, and (3) choose your living standard.
\vspace{4pt}\newline
\textbf{Goal Setting \& Reflection}
\newline You have both \textbf{long-term goals} (months to years---deeper aspirations connecting to your core motivation) and \textbf{short-term goals} (this week---concrete, actionable steps).
\vspace{3pt}\newline
\textbf{Reflection}: Review your previous goals from your scratchpad. For long-term goals: Are you making progress? Should you adjust? For short-term goals: Did you achieve last week's goals? If not, why? If a goal has been stuck for weeks, consider trying a completely different approach or accepting it may not be achievable right now.
\vspace{4pt}\newline
\textbf{Weekly Planning}
\newline Check your existing schedule, outline your general weekly plan, and plan one specific activity for each day's free time slot. Your weekly plan should directly serve your goals.
\vspace{4pt}\newline
\textbf{Living Standard Selection} (Required)
\newline Choose ONE: frugal (100/week, Material $-$5), moderate (200/week, unchanged), comfortable (300/week, Material $+$5), or luxurious (500/week, Material $+$10).

\\
\midrule

\textbf{Contact Stage} &

In this phase, you can contact other people by sending ``text messages'' and arrange joint activities through communication.
\vspace{3pt}\newline
- Communication is asynchronous (text-message style, not real-time).
\newline - This phase proceeds in N rounds. In each round, you read all received messages, then decide who to send/reply to.
\newline - You can trigger four types of role actions:
\vspace{3pt}\newline
\quad i. \texttt{contact(message, to)}: Send a text message. Does NOT create any activity or schedule.
\vspace{2pt}\newline
\quad ii. \texttt{propose\_joint\_activity(activity\_name, message, proposal, invited\_persons, time, location, required\_participants)}: The ONLY way to create a joint activity. Time format: \texttt{Y[year]-W[week]-activity-D[day]}. Location must be from the map.
\vspace{2pt}\newline
\quad iii. \texttt{respond\_invitation(activity\_name, message, to, decision)}: Accept (``yes'') or decline (``no'') an invitation.
\vspace{2pt}\newline
\quad iv. \texttt{cancel\_joint\_activity(activity\_name, message)}: Cancel an activity you proposed.
\vspace{4pt}\newline
\textbf{Important}: Verbal agreements via \texttt{contact} do NOT create schedules. You MUST use \texttt{propose\_joint\_activity} to formally propose, and the invitee MUST use \texttt{respond\_invitation} with ``yes'' to confirm.

\\
\bottomrule
\end{tabular}
\end{table*}

\begin{table*}[h]
\centering
\caption{Agent prompts for joint and solo activity phases.}
\label{tab:prompts_activity}
\begin{tabular}{p{0.8in}p{4.3in}}
\toprule
\multicolumn{2}{c}{\textbf{Activity Prompts}} \\
\midrule

\textbf{Joint \newline Activity} &

Engage in the joint activity that you have previously proposed or accepted.
\vspace{3pt}\newline
- Stay in your character. You control only your character's thoughts, speech and actions.
\newline - You can interact with other people or the physical environment, and wait for the responses or outcomes. Never decide outcomes beyond your ability or control. Never narrate others' thoughts and responses.
\newline - By default, your response is visible to all participants. In addition:
\newline \quad - Use \texttt{$\langle$private$\rangle$...$\langle$/private$\rangle$} tags for content invisible to others (inner monologue).
\newline \quad - Use \texttt{$\langle$visible\_to=``name1,name2''$\rangle$...$\langle$/visible\_to$\rangle$} for content visible only to specified persons (whispers, secret gestures).
\newline - You can gift items using \texttt{$\langle$role\_action$\rangle$gift(to=..., item=...)$\langle$/role\_action$\rangle$}.
\newline - You can leave early using \texttt{$\langle$role\_action$\rangle$exit\_activity()$\langle$/role\_action$\rangle$}.
\newline - This activity lasts for $n_{\min}$ to $n_{\max}$ turns. Keep your response concise and less than 200 words.

\\
\midrule

\textbf{Solo \newline Activity} &

This is your free time for the day, approximately 2--3 hours. You may choose one solo activity based on your own will. This includes (but not limited to) learning and self-improvement, extra work, shopping, leisure and entertainment.
\vspace{4pt}\newline
\textbf{Important notes:}
\newline - You can only choose ONE activity.
\newline - You should specify the expected activity content (e.g., what subject you will study or what work you will do).
\newline - Activities will affect your state, including vitality/fulfillment/skills/assets. For example, you may consume vitality to gain skills through learning, earn money through work, spend money through shopping, or gain fulfillment through leisure.
\newline - If you want to shop or spend on services, specify your requirements and budget. You will be informed of available options and costs.
\newline - All activity outcomes depend on your talents and current skills.
\newline - Your solo activity will not be known to others.
\vspace{4pt}\newline
\textbf{Output Format:}
\newline Thinking: [reasoning]
\newline Activity: [description, max 100 words]

\\
\bottomrule
\end{tabular}
\end{table*}

\begin{table*}[h]
\centering
\caption{Agent prompts for weekly review and social evaluation.}
\label{tab:prompts_review_social}
\begin{tabular}{p{0.8in}p{4.3in}}
\toprule
\multicolumn{2}{c}{\textbf{Review \& Social Evaluation Prompts}} \\
\midrule

\textbf{Weekly \newline Review} &

Based on everything that happened this week, please write a weekly summary.
Your summary should include two parts:
\vspace{3pt}\newline
1. \textbf{Summary}: A factual record of what happened this week---key activities and their outcomes, important interactions and conversations, any notable changes or developments.
\vspace{2pt}\newline
2. \textbf{Reflection}: Your personal thoughts and insights---what did you learn or realize? How do you feel about what happened? What would you do differently?
\vspace{4pt}\newline
\textbf{Guidelines:}
\newline - Focus on the most meaningful moments.
\newline - You may update your scratchpads if you have new insights to record.
\vspace{4pt}\newline
\textbf{Output Format:}
\newline Thinking: [reasoning]
\newline Summary: [factual record, $<$300 words]
\newline Reflection: [personal thoughts, $<$300 words]

\\
\midrule

\textbf{Social \newline Evaluation} &

This is a \textbf{COMPLETELY PRIVATE} evaluation that NO OTHER CHARACTER WILL EVER SEE. Be completely honest about your true feelings and judgments.
\vspace{3pt}\newline
Score each person you know on TWO separate dimensions (0--100):
\vspace{3pt}\newline
\textbf{Affection}: How much you personally like them.
\newline Consider: emotional closeness, enjoyment of their company, personal affinity, comfort around them.
\vspace{3pt}\newline
\textbf{Respect}: How much you admire their abilities and character.
\newline Consider: competence, accomplishments, reliability, wisdom and judgment.
\vspace{4pt}\newline
\textbf{Scoring Scale:}
\newline 10 = extreme dislike/contempt
\newline 30 = dislike/low regard
\newline 50 = neutral baseline
\newline 70 = like/respect
\newline 90 = deep affection/great admiration
\vspace{3pt}\newline
Evaluate each person independently against the 50 baseline. Do NOT compare people against each other. Most acquaintances should fall in the 40--60 range. Affection and respect scores can be different for the same person.
\vspace{3pt}\newline
\textbf{Output:} \{``affection'': \{``person'': score, ...\}, ``respect'': \{``person'': score, ...\}\}

\\
\bottomrule
\end{tabular}
\end{table*}

\begin{table*}[h]
\centering
\caption{Environment model prompt for joint activity management: environment modeling, next speaker prediction, and response verification (filtering).}
\label{tab:prompts_god_joint}
\begin{tabular}{p{0.8in}p{4.3in}}
\toprule
\multicolumn{2}{c}{\textbf{Environment Model: Joint Activity World Model}} \\
\midrule

\textbf{Environment \newline Modeling} &

You are the world model for a multi-character role-play. Each time a character acts, you need to perform three tasks.
\vspace{4pt}\newline
\textbf{Task 1: Environment Modeling.} Based on the last character's actions and dialogues, describe the resulting changes in the environment:
\newline - Physical changes in the setting
\newline - Reactions of nameless bystanders/crowds (not the participants)
\newline - Ambient sounds, weather changes, or atmospheric shifts
\vspace{3pt}\newline
Keep descriptions concise (1--3 sentences). Respond to subtle cues in the characters' interactions. Do not invent new named entities or contradict known facts.

\\
\midrule

\textbf{Next Speaker \newline Prediction} &

\textbf{Task 2: Next Speaker Prediction.} Choose exactly one name from \{participants\}. If the activity should conclude now, output ``$\langle$END CHAT$\rangle$''.
\vspace{3pt}\newline
Notes:
\newline - Identify the core participants within this activity.
\newline - Favor a character with an unresolved intent, someone who was addressed, or the one least recently active.

\\
\midrule

\textbf{Response \newline Verification} &

\textbf{Task 3: Response Verification.} Evaluate whether the last speaker's response violates any roleplay principles. Output PASS if no violation. Output REJECT with a brief reason if ANY of the following violations is detected:
\vspace{3pt}\newline
1. \textbf{Scope of control violation}: determined outcome of own actions, or controlled others
\newline 2. \textbf{Physical plausibility violation}: actions violating the world's physical laws
\newline 3. \textbf{Hallucination}: referenced objects or events not present in context
\newline 4. \textbf{Character inconsistency}: behavior deviates from established persona
\newline 5. \textbf{Cognitive boundary violation}: knowledge exceeds the character's background
\newline 6. \textbf{State inconsistency}: abrupt shift in physical/mental state
\newline 7. \textbf{Emotional discontinuity}: abrupt emotional shift
\newline 8. \textbf{Unnatural relationship progression}: relationship develops too abruptly
\newline 9. \textbf{Parroting}: repeated same content 3+ times
\newline 10. \textbf{AI assistant behavior}: spoke like a customer service agent
\newline 11. \textbf{Empty dialogue}: no new information, conversation spinning

\\
\bottomrule
\end{tabular}
\end{table*}

\begin{table*}[h]
\centering
\caption{Environment model prompts for evaluating activity outcomes and determining state changes. Abbreviated; the full version includes detailed delta ranges and fulfillment dimension definitions.}
\label{tab:prompts_god_eval}
\begin{tabular}{p{0.8in}p{4.3in}}
\toprule
\multicolumn{2}{c}{\textbf{Environment Model: Activity Outcome Evaluation}} \\
\midrule

\textbf{Solo \newline Activity \newline Evaluation} &

You are the world model. A character is performing a solo activity. Your task is to evaluate the activity type and determine the outcome.
\vspace{4pt}\newline
\textbf{Step 1: Classify activity type.} Based on the character's intended activity, determine if this is a consumption event (explicitly intends to purchase goods or services). Learning, working, resting are NOT consumption. If uncertain, default to non-consumption.
\vspace{4pt}\newline
\textbf{For consumption events:}
\newline Return only \{``is\_consumption\_event'': true\}. Prices and options are generated in a separate stage.
\vspace{4pt}\newline
\textbf{For non-consumption events:} Evaluate the outcome with state deltas.
\newline - \textbf{Outcome message}: 2--4 sentences describing what happens
\newline - \textbf{Activity-specific rules}: Learning effectiveness depends on current skill level. Work income depends on skills/talents/position (range: 40--200 currency per session).
\newline - \textbf{Delta ranges}: vitality [min, max], mood [min, max], esteem [min, max], skills [min, max], money [0, max] for work only, gain\_items = [] for non-consumption.
\vspace{4pt}\newline
\textbf{Output:} \{``outcome'': ``...'', ``is\_consumption\_event'': false, ``delta\_vitality'': ..., ``delta\_fulfillment'': \{``mood'': ..., ``esteem'': ...\}, ``delta\_skills'': \{...\}, ``delta\_money'': ..., ``gain\_items'': []\}

\\
\midrule

\textbf{Joint \newline Activity \newline Evaluation} &

Multiple characters have just finished a joint activity. Evaluate the outcome and determine state changes for each participant.
\vspace{3pt}\newline
\textbf{Context:} \{activity\_background\}, \{participants\_info\}, \{dialog\_history\}
\vspace{3pt}\newline
Evaluate delta\_vitality, delta\_fulfillment (mood, social, esteem) and delta\_skills for each participant.
\vspace{4pt}\newline
\textbf{Important notes:}
\newline - Joint activities do NOT allow work or consumption. Do not evaluate delta\_money.
\newline - Deltas should reflect each character's actual involvement and personality.
\newline - Social fulfillment reflects bonding and connection.
\newline - Mood depends on activity enjoyment and compatibility with other participants.
\newline - Esteem can increase if the person feels respected or accomplished.
\newline - Each participant must have an outcome, even if they were less active.
\vspace{4pt}\newline
\textbf{Output:} \{``Alice'': \{``delta\_vitality'': ..., ``delta\_fulfillment'': \{``mood'': ..., ``social'': ..., ``esteem'': ...\}, ``delta\_skills'': \{...\}\}, ``Bob'': \{...\}\}

\\
\bottomrule
\end{tabular}
\end{table*}

\begin{table*}[h]
\centering
\caption{Environment model prompts for world management: encounter generation and yearly profile updates. Abbreviated; the full version includes JSON output format specifications.}
\label{tab:prompts_god_world}
\begin{tabular}{p{0.8in}p{4.3in}}
\toprule
\multicolumn{2}{c}{\textbf{Environment Model: World Management}} \\
\midrule

\textbf{Encounter \newline Generation} &

You are the system administrator creating encounter events---random meetings between idle characters.

An encounter is a system-arranged coincidental meeting where two idle characters unexpectedly meet in a natural, believable context.
\vspace{4pt}\newline
\textbf{Pairing Strategy:}
\newline - Prefer pairing characters who have some relationship
\newline - Characters already close can have deepening moments
\newline - Characters who don't know each other can be paired for a ``first meeting''
\newline - Each character can only appear in ONE encounter per day
\vspace{4pt}\newline
\textbf{Scene Description:}
\newline - Describe the circumstance of the meeting objectively
\newline - Create natural conflict, decision point, or interesting situation
\newline - Do NOT describe what the characters think, say, or decide
\newline - Do NOT assume specific character behaviors unless universal
\vspace{4pt}\newline
\textbf{Good examples:}
\newline - ``Both are waiting at the bus stop in the rain, realizing they are the only two people there.''
\newline - ``At the convenience store, both reach for the last bottle of drink on the shelf at the same time.''
\vspace{3pt}\newline
\textbf{Bad examples} (avoid):
\newline - ``Alice sees Bob and feels happy, deciding to go say hello.'' (describes thoughts)
\newline - ``Both are searching for the same rare book.'' (assumes specific behavior)

\\
\midrule

\textbf{Yearly \newline Profile \newline Update} &

A year has passed in the simulation. Based on the character's experiences throughout the year, update their profile for the upcoming year.
\vspace{3pt}\newline
\textbf{Input:} current profile (JSON) + all weekly summaries from the past year.
\vspace{4pt}\newline
\textbf{Fields that MUST NOT change:}
\newline name, birthday, gender, position (handled by position application system), init\_skills, init\_assets, extra\_income.
\vspace{4pt}\newline
\textbf{Fields to update:}
\newline - \textbf{appearance\_and\_impression}: Physical appearance changes very slowly. Only allow subtle aging-related changes or changes clearly indicated by events.
\newline - \textbf{personality\_traits}: Qualitative narrative + quantitative values (each can change by at most $\pm$5/year, range [0, 100]).
\newline - \textbf{talents}: Qualitative + quantitative (at most $\pm$3/year, range [0, 100]).
\newline - \textbf{Other fields}: brief\_introduction, core\_motivation, conflicts, values, preferences, details---update to reflect growth and changes.
\vspace{4pt}\newline
\textbf{Important:} Changes should be gradual and justified by the year's experiences. Do NOT invent events not mentioned in the summaries. Keep the character's core identity intact while allowing natural evolution.

\\
\bottomrule
\end{tabular}
\end{table*}

\end{document}